%% file: main.tex
\documentclass{article} 
\usepackage[dvipsnames]{xcolor}
\usepackage{iclr2025_conference, times}
\iclrfinalcopy

\usepackage{graphicx}%
\usepackage{multirow}%
\usepackage{amsmath,amssymb,amsfonts}%
\usepackage{amsthm}%
\usepackage{mathrsfs}%
\usepackage{mathtools}%
\usepackage{pgffor}
\usepackage[title]{appendix}%
\usepackage{textcomp}%
\usepackage{manyfoot}%
\usepackage{booktabs}%
\usepackage{algorithm}%
\usepackage{algorithmicx}%
\usepackage{algpseudocode}%
\usepackage{fancyhdr}
\usepackage{listings}%
\usepackage[normalem]{ulem}%
\usepackage{CJKutf8}%
\usepackage{setspace}%
\usepackage{xparse}%
\usepackage{multirow}%
\usepackage{bm}%
\usepackage{placeins}%
\usepackage{subcaption}%
\usepackage{lineno}
\usepackage{pgffor}
\usepackage{xstring}
\usepackage[colorlinks]{hyperref}
\usepackage{changes}
\usepackage{rotating}
\usepackage{cleveref}%

\hypersetup{
linkcolor=NavyBlue
,citecolor=NavyBlue
,filecolor=Red
,urlcolor=NavyBlue
,menucolor=Red
,runcolor=Red
,linkbordercolor=NavyBlue
,citebordercolor=NavyBlue
,filebordercolor=Red
,urlbordercolor=NavyBlue
,menubordercolor=Red
,runbordercolor=Red
}

\theoremstyle{thmstyleone}%
\newtheorem{theorem}{Theorem}
\newtheorem{proposition}{Proposition}%
\newtheorem{lemma}{Lemma}%
\newtheorem{corollary}[lemma]{Corollary}%

\theoremstyle{thmstyletwo}%
\newtheorem{remark}{Remark}%
\newtheorem{assumption}{Assumption}%
\newtheorem{LocalAssumption}[assumption]{Assumption}%

\theoremstyle{thmstylethree}%
\newtheorem{definition}{Definition}%

\Crefname{theorem}{Theorem}{Theorems}
\Crefname{lemma}{Lemma}{Lemmas}
\crefname{theorem}{Theorem}{Theorems}
\crefname{lemma}{Lemma}{Lemmas}
\crefname{assumption}{Assumption}{Assumptions}
\Crefname{assumption}{Assumption}{Assumptions}
\crefname{LocalAssumption}{Assumption}{Assumptions}
\Crefname{LocalAssumption}{Assumption}{Assumptions}
\crefname{definition}{Definition}{Definitions}
\Crefname{definition}{Definition}{Definitions}
\crefname{proposition}{Proposition}{Propositions}
\Crefname{proposition}{Proposition}{Propositions}
\crefname{corollary}{Corollary}{Corollaries}
\Crefname{corollary}{Corollary}{Corollaries}
\Crefname{figure}{Figure}{Figures}
\crefname{figure}{Figure}{Figures}
\crefname{table}{Table}{Tables}
\Crefname{table}{Table}{Tables}
\crefname{section}{Section}{Sections}
\Crefname{section}{Section}{Sections}
\crefname{equation}{equation}{equations}
\Crefname{equation}{Equation}{Equations}
\crefname{appendix}{Appendix}{Appendices}
\Crefname{appendix}{Appendix}{Appendices}

\newcommand{\re}[1]{\textcolor{red}{#1}}
\newcommand{\bl}[1]{\textcolor{blue}{#1}}

\renewcommand{\re}[1]{#1}
\renewcommand{\bl}[1]{#1}

\input{symbols.tex}
\input{macro.tex}
\input{setting.tex}

\title{On the Implicit Adversariality of Catastrophic Forgetting in Deep Continual Learning}

\newcommand{\affiliationnju}{$^\lozenge$}
\newcommand{\affiliationseu}{$^\blacklozenge$}
\newcommand{\myauthors}{%
    \renewcommand{\thefootnote}{\fnsymbol{footnote}}%
    Ze Peng\affiliationnju$^\dagger$,
    Jian Zhang\affiliationnju$^\dagger$,
    Jintao Guo\affiliationnju,
    Lei Qi\affiliationseu,
    Yang Gao\affiliationnju$^*$,
    Yinghuan Shi\affiliationnju$^*$\\
    \setlength{\tabcolsep}{-1pt}
    \begin{tabular}{p{1em}l}
        $^\dagger$ & These authors contributed equally to this work. $^*$ Corresponding authors.\\
        \affiliationnju & State Key Laboratory for Novel Software Technology, Nanjing University\\
        \affiliationseu & School of Computer Science and Engineering, Southeast University
    \end{tabular}
    \allaffiliations{}\\
    \texttt{\href{mailto:pengze@smail.nju.edu.cn}{pengze@smail.nju.edu.cn}, \href{mailto:zhang.jian@nju.edu.cn}{zhang.jian@nju.edu.cn}},\\ \texttt{\href{mailto:guojintao@smail.nju.edu.cn}{guojintao@smail.nju.edu.cn}}, 
    \texttt{\href{mailto:qilei@seu.edu.cn}{qilei@seu.edu.cn}},\\ \texttt{\{\href{mailto:gaoy@nju.edu.cn}{gaoy}, \href{mailto:syh@nju.edu.cn}{syh}\}@nju.edu.cn}%
    \renewcommand{\thefootnote}{\arabic{footnote}}%
    \setcounter{footnote}{0}%
}

\begin{document}

\maketitle

\input{abstract.tex}

\input{introduction.tex}

\input{results/results.tex}

\input{conclusion.tex}

\input{methods/methods.tex}

\section*{Acknowledgments}
This work was supported by NSFC Project (62222604, 62506162, 62192783, 62536005, 62206052) and Jiangsu Science and Technology Project (BF2025061, BG2024031).

\bibliography{refs,flatness}
\bibliographystyle{iclr2025_conference}

\newpage

\appendix

\input{results/appendix.tex}

\input{proofs/appendix.tex}



\end{document}

%% file: symbols.tex
\renewcommand{\vec}[1]{\bm{#1}}

\newcommand{\defeq}{\coloneq}
\newcommand{\distribution}{P}
\newcommand{\prob}[2][]{P_{#1}\left(#2\right)}
\newcommand{\reals}{\mathbb{R}}
\newcommand{\ex}[2][]{\mathbb{E}_{#1}\left[#2\right]}
\newcommand{\exx}[1][]{\mathbb{E}_{#1}}
\newcommand{\variance}[2][]{\mathbb{V}_{#1}\left[#2\right]}

\newcommand{\ltwo}{L_2}

\newcommand{\alignment}{\alpha}

\newcommand{\lagrangian}{\mathcal{L}}
\newcommand{\emploss}{\hat{\mathcal{L}}}
\newcommand{\norm}[1]{\left\|#1\right\|}
\newcommand{\Fnorm}[1]{\norm{#1}_F}
\newcommand{\nuclearnorm}[1]{\norm{#1}_*}
\newcommand{\trace}[1]{\operatorname{tr}\left(#1\right)}
\newcommand{\trr}{\operatorname{tr}}

\newcommand{\erank}[1]{\operatorname{erank}\left(#1\right)}

\newcommand{\mat}[1]{\bm{#1}}
\newcommand{\weight}{\mat{W}}
\newcommand{\allweights}{\left(\weight_i\right)_{i=1}^L}

\newcommand{\complexobject}[1]{\bm{#1}}
\newcommand{\task}{\complexobject{T}}

\newcommand{\sampleinput}{\vec{x}}

\newcommand{\EmpiricalInputs}{\mat{X}}
\newcommand{\EmpiricalLabels}{\mat{Y}}

\newcommand{\RandomOrthogonalMatrix}{\mat{U}}

\newcommand{\hessian}{\mat{H}}
\newcommand{\param}{\vec{\theta}}

\newcommand{\set}[1]{\left\{#1\right\}}
\newcommand{\vectorize}[1]{\operatorname{vec}\left(#1\right)}

\NewDocumentCommand{\packed}{m O{i} O{L}  O{1}}{\left(#1\right)_{#2=#4}^{#3}}

\newcommand{\inner}[2]{\left\langle #1, #2 \right\rangle}
\newcommand{\partialfrac}[2]{\frac{\partial #1}{\partial #2}}
\newcommand{\diff}{\operatorname{d}}

\newcommand{\TaskDistribution}[1][]{\distribution_{(\EmpiricalInputs_{#1}, \EmpiricalLabels_{#1})}}
\newcommand{\oldtask}{\task_1}
\newcommand{\newtask}{\task_2}
\newcommand{\oldhessian}{\hessian_1}
\newcommand{\OldInputs}{\EmpiricalInputs_1}
\newcommand{\OldLabels}{\EmpiricalLabels_1}
\newcommand{\NewInputs}{\EmpiricalInputs_2}
\newcommand{\NewLabels}{\EmpiricalLabels_2}

\newcommand{\OldParam}{\param_1}
\newcommand{\NewParam}{\param_2}
\newcommand{\OldLoss}{\emploss_1}
\newcommand{\NewLoss}{\emploss_2}

\newcommand{\trop}{\operatorname{tr}}

\newcommand{\RandRot}{\mat{U}}
\newcommand{\SingularMatrix}[1]{\mat{\Sigma}_{#1}}

\newcommand{\CommonSV}{\mat{\Sigma}}
\newcommand{\mU}{\mat{U}}
\newcommand{\mV}{\mat{V}}

\newcommand{\HessianEigVal}{\lambda}
\newcommand{\HessianEigVec}{\vec{v}}

\newcommand{\perturbation}{\xi}
\newcommand{\vecperturbation}{\vec{\perturbation}}

\newcommand{\depth}{L}
\newcommand{\Input}{\vec{x}}
\newcommand{\Label}{\vec{y}}
\newcommand{\GroundTruth}{\mat{\Phi}}
\newcommand{\OldGroundTruth}{\GroundTruth_1}

\newcommand{\mI}{\mat{I}}

\newcommand{\defto}{\eqqcolon}

\newcommand{\hidden}[1]{\EmpiricalInputs^{(#1)}}

\newcommand{\nullspace}[1]{\left(#1\right)^{\perp}}
\NewDocumentCommand{\approxnull}{m O{}}{\nullspace{#1}_{#2}}

\newcommand{\acc}{\text{ACC}}
\newcommand{\bwt}{\text{BWT}}
\newcommand{\immAcc}{\text{immACC}}

\newcommand{\spectralReg}{\mathcal{L}_{\sigma}}
\newcommand{\mF}{\mat{F}}
\newcommand{\mM}{\mat{M}}

\newcommand{\nearzero}{{\tiny \approx} \mat{0}}

\newcommand{\tightbound}{\underline{\alpha}_{\text{tighter}}}

\newcommand{\mS}{\mat{S}}
\newcommand{\mA}{\mat{A}}
\newcommand{\mB}{\mat{B}}

\newcommand{\diag}{\operatorname{diag}}
\newcommand{\Diag}{\operatorname{Diag}}
\newcommand{\ve}{\vec{e}}

\newcommand{\hadamard}{\odot}
\newcommand{\vK}{\vec{K}}

\newcommand{\newhessian}{\hessian_2}
\newcommand{\kronecker}{\otimes}

\newcommand{\mJ}{\mat{J}}

\newcommand{\pinv}{\dagger}
\newcommand{\snorm}[1]{\norm{#1}_2}
\newcommand{\mIL}[1]{\mI^{\text{left}}_{#1}}
\newcommand{\abs}[1]{\left|#1\right|}

\newcommand{\oldYX}{\OldLabels \OldInputs^\pinv}

%% file: macro.tex
\newcommand{\ie}{i.e.,}
\newcommand{\eg}{e.g.,}
\newcommand{\wrt}{w.r.t.}
\newcommand{\AutoAlignmentProperty}{For $i \in [L-1]$, we have
    \begin{align}
        \weight_{i+1}^\top \weight_{i+1} = \weight_{i} \weight_{i}^\top.
    \end{align}}

\newcommand{\CL}{CL}

\newcommand{\LargeCurvature}{high-curvature}
\newcommand{\AlignmentPhenomenon}{alignment}

\newcommand{\AdversarialAlignment}{adversarial alignment}
\newcommand{\CapAdversarialAlignment}{Adversarial Alignment}

\newcommand{\supplementary}{Supplementary Material}

\newcommand{\backGP}{\text{backGP}}
\newcommand{\GP}{gradient projection}
\newcommand{\backNSCL}{\text{backGP}}
\newcommand{\bGP}{\text{b.GP}}

\newcommand{\optional}{$^\dagger$}

\newcommand{\localBest}[1]{\uwave{#1}}
\newcommand{\best}[1]{\textbf{#1}}
\newcommand{\nocomp}[1]{\textcolor{gray}{#1}}

\newcommand{\spontaneous}{\text{spontaneous}}

\newcommand{\labelsize}{\scriptsize}

\newcommand{\symbollist}{\dagger,\ddagger,\lozenge}

\def\usedsymbols{}

\def\affiliations{}

\newcommand{\affiliation}[2]{%
    \def\affilmark{}%
    \foreach \x in \symbollist {%
        \ifx\affilmark\empty
            \IfSubStr{\usedsymbols}{\x}{}{
                \xdef\affilmark{\x}
                \xdef\usedsymbols{\usedsymbols,\x}
            }%
        \fi%
    }%
    \expandafter\xdef\csname affiliation#1\endcsname{$^{\affilmark}$}%
    \xdef\affiliationItem{$^{\affilmark}$ & #2}%
    \ifx\affiliations\empty
        \edef\affiliations{\affiliationItem}%
    \else
        \edef\affiliations{\affiliations\\\affiliationItem}%
    \fi
}

\newcommand{\allaffiliations}{%
    \setlength{\tabcolsep}{-1pt}
    \begin{tabular}{p{0.75em}lll}
        \affiliations
    \end{tabular}
}

%% file: setting.tex
\newif{\ifMakeReasonsIntriguing}
\MakeReasonsIntriguingtrue

\newif{\ifExplicitHessian}
\ExplicitHessiantrue

\newif{\ifArxiv}
\Arxivtrue

%% file: abstract.tex
\begin{abstract}%
    Continual learning seeks the human-like ability to accumulate new skills in machine intelligence. Its central challenge is catastrophic forgetting, whose underlying cause has not been fully understood for deep networks.
    In this paper, we demystify catastrophic forgetting by revealing that the new-task training is implicitly an adversarial attack against the old-task knowledge.
    Specifically, the new-task gradients automatically and accurately align with the sharp directions of the old-task loss landscape, rapidly increasing the old-task loss.
    This \emph{\AdversarialAlignment{}} is intriguingly counter-intuitive because the sharp directions are too sparsely distributed to align with by chance.
    To understand it, we theoretically show that it arises from training's low-rank bias, which, through forward and backward propagation, confines the two directions into the same low-dimensional subspace, facilitating alignment.
    Gradient projection (GP) methods, a representative family of forgetting-mitigating methods, reduce \AdversarialAlignment{} caused by forward propagation, but cannot address the alignment due to backward propagation.
    We propose \backGP{} to address it, which reduces forgetting by 10.8\% and improves accuracy by 12.7\% on average over GP methods.
\end{abstract}
\ifArxiv
\else
    \keywords{Deep Learning, Continual Learning, Catastrophic Forgetting, Implicit Bias}
\fi

%% file: introduction.tex
Continual learning (\CL{}) aims to equip machine learning systems with the human-like ability to acquire new skills sequentially without sacrificing performance on previously learned tasks.
A central challenge of \CL{} is catastrophic forgetting, where training on new tasks overwrites old-task knowledge and severely degrades old-task performance. 
Successful forgetting mitigation has been made from optimization \citep{nscl,gpm,adaNSCL}, regularization \citep{regularization_CL_1,regularization_CL_2}, parameter expansion \citep{expansion_CL_1,expansion_CL_2,expansion_CL_3}, and experience replay \citep{replay_CL_1,replay_CL_2,replay_CL_3,replay_CL_4} perspectives. However, these methods remain heuristic, offering limited theoretical insight into why forgetting occurs and how to mitigate forgetting by improving existing methods in a principled manner.

Recently, theoretical studies have linked forgetting to data factors such as task similarity, task ordering, or data diversity \citep{interest_in_forgetting_mechanism4,interest_in_forgetting_mechanism5,interest_in_forgetting_mechanism1,interest_in_forgetting_mechanism2,interest_in_forgetting_mechanism3,intreset_in_forgetting_mechanism6}.
However, these analyses only study single-layer networks, resulting in conclusions that cannot be directly applied to deep networks, whose training dynamics and inductive bias differ drastically \citep{implicit_auto_balancedness,dln_bias_1,dln_bias_2,dln_bias_3} and may lead to different forgetting behaviors.
This gap leaves open questions of whether, how, and why forgetting manifests differently in deep networks.

\begin{figure}
    \centering
    \includegraphics[width=0.9\linewidth]{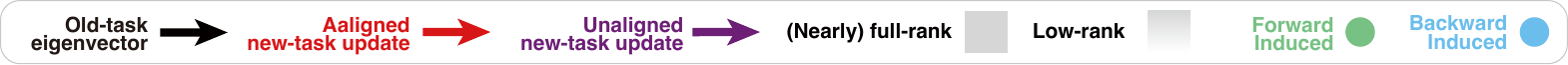}\\
    \begin{subfigure}{0.35\linewidth}
        \centering
        \includegraphics[width=\linewidth]{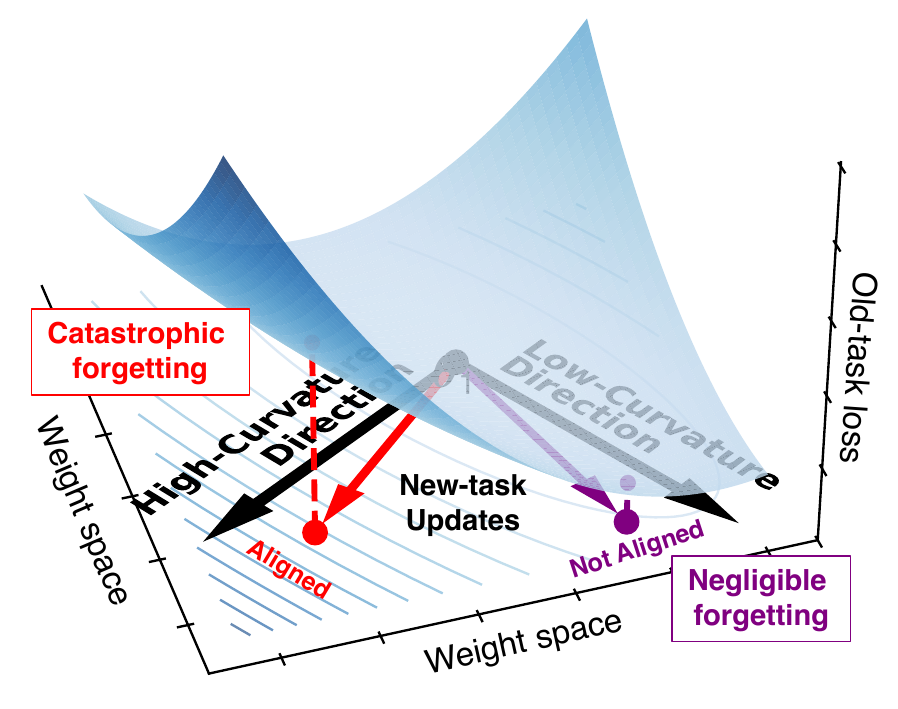}
        \caption{}\label{fig:illustration_alignment}
    \end{subfigure}
    \begin{subfigure}{0.26\linewidth}
        \centering
        \includegraphics[width=\linewidth]{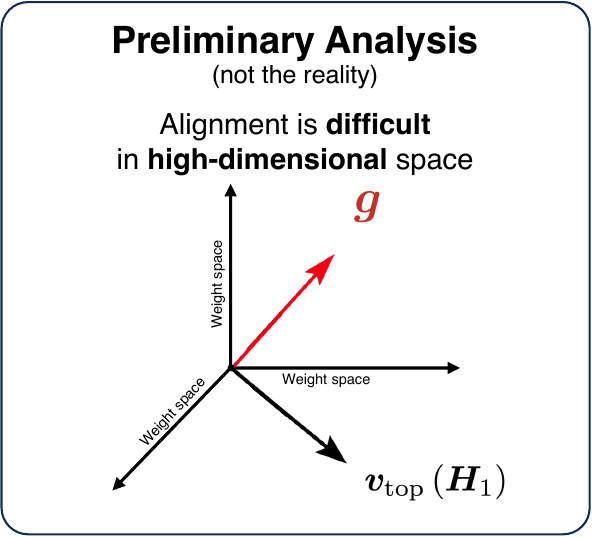}
        \caption{}\label{fig:illustration_adversarial}
    \end{subfigure}
    \begin{subfigure}{0.26\linewidth}
        \centering
        \includegraphics[width=\linewidth]{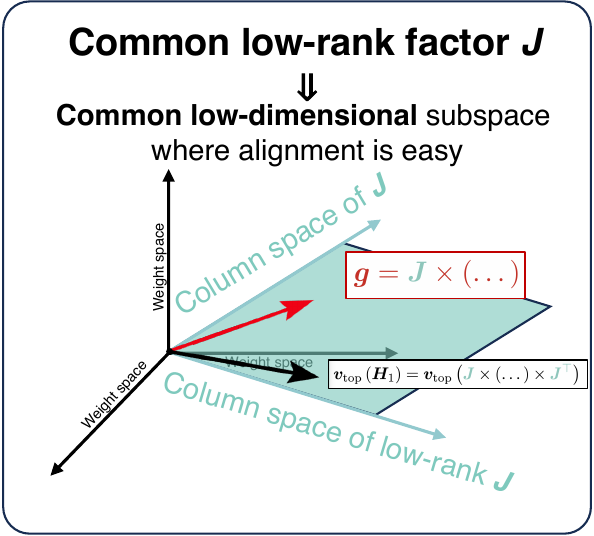}
        \caption{}\label{fig:illustration_causes}
    \end{subfigure}
    \begin{subfigure}{0.95\linewidth}
        \centering
        \includegraphics[width=\linewidth]{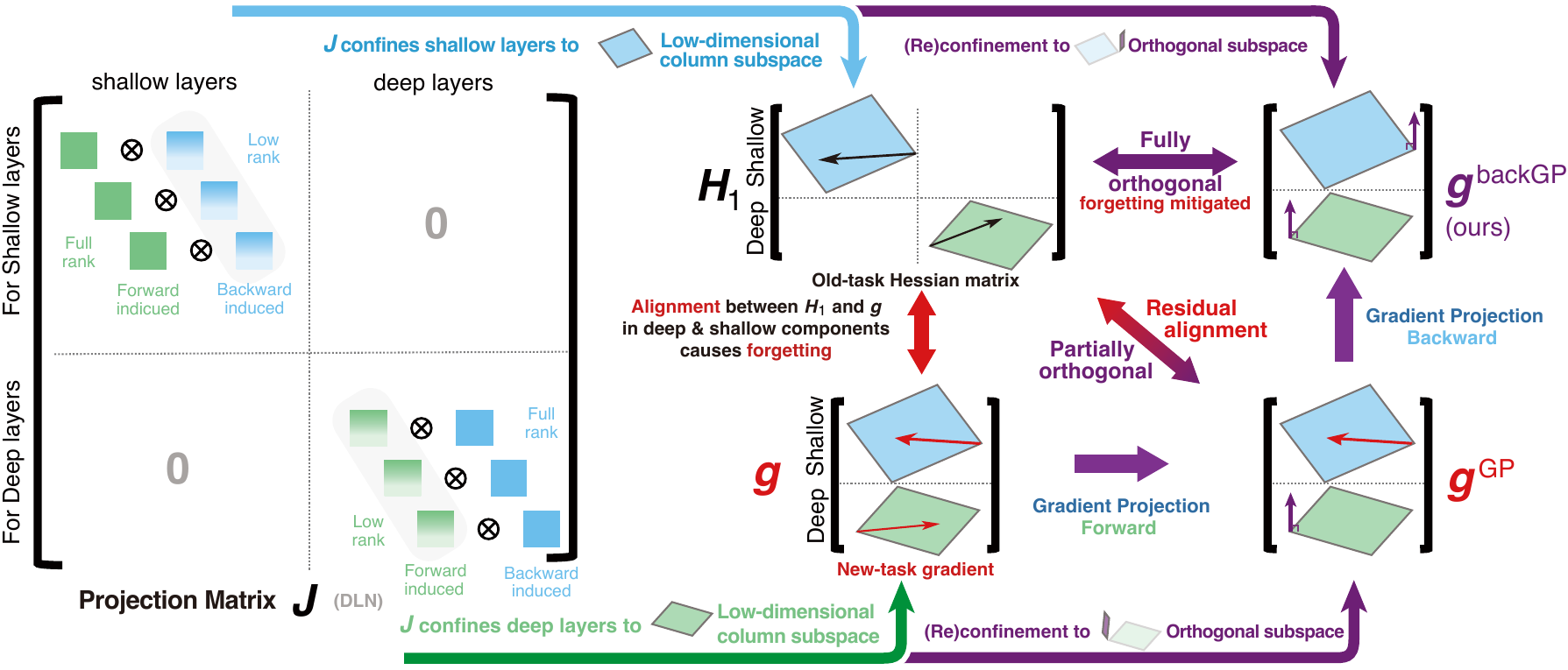}
        \caption{}\label{fig:illustration_mitigation_of_aa}
    \end{subfigure}
    \caption{
        \textbf{Illustration of \AdversarialAlignment{}'s definition, influence, counter-intuitiveness, cause and mitigation.}         
        \Cref{fig:illustration_alignment} illustrates the definition of the alignment using an example of aligned new-task updates, which is contrasted with unaligned updates. \Cref{fig:illustration_alignment} also illustrates that the aligned new-task updates lead to large old-task loss increase, \ie{} catastrophic forgetting, while unaligned new-task updates do not.
        \Cref{fig:illustration_adversarial} shows the expectation of intuitive preliminary analysis, \ie{} the new-task updates and the sparsely distributed old-task \LargeCurvature{} directions should not persistently align in the high-dimensional weight space. The intuitive expectation mismatches the reality, indicating the alignment is counter-intuitive.
        \Cref{fig:illustration_causes} illustrates the cause of \AdversarialAlignment{}, \ie{} both directions have low-rank Jacobian $\mJ$ as a common factor, which confines them to the \textbf{same} \emph{low-dimensional} subspace (column space of $\mJ$), where alignment is much easier.
        \Cref{fig:illustration_mitigation_of_aa} illustrates how existing GP methods and our \backGP{} methods mitigate the \AdversarialAlignment{} (at least for deep linear networks). 
        Details can be found in \cref{sec:analysis_of_existing_methods,sec:resolution_of_limitations}.
    }\label{figure:main}
\end{figure}

A promising tool to analyze deep-network forgetting is the loss landscape, which depicts loss changes \wrt{} model weights.
Catastrophic forgetting has been linked to \emph{the alignment between new-task updates and \LargeCurvature{} directions of local old-task loss landscape} \citep{hessian_perspective_of_regularization_CL,meat_learning_as_regularization,forgetting_and_flatness1,forgetting_and_flatness2}. As illustrated in \cref{fig:illustration_alignment}, these high-curvature directions, \ie{} top eigenvectors of the old-task Hessian, are where old-task loss increases the most rapidly.
Recent theoretical results \citep{hessian_perspective_of_regularization_CL,meat_learning_as_regularization} observe that a wide range of \CL{} algorithms can effectively prevent the alignment, suggesting the critical role of alignment in forgetting. 
However, the existence of alignment has never been directly verified, and the cause of the \spontaneous{} alignment is also unclear, leaving a gap in understanding catastrophic forgetting of deep networks.
Therefore, in this paper, we systematically study the alignment phenomenon with four steps: (1) existence, and \emph{given the existence}, (2) cause, (3) influence (on forgetting), and (4) mitigation of alignment.

We first empirically show deep networks \emph{\spontaneous ly} exhibit strong and persistent alignment between new-task updates and old-task \LargeCurvature{} directions. We also derive theoretical and empirical connections between alignment and forgetting, confirming the existence of alignment and its critical role in catastrophic forgetting. 

When trying to understand the cause of the alignment, we find it highly counter-intuitive and intriguing based on the following preliminary analysis: 
(1) The old and new tasks have distinct data, which weakens the correlation between the \emph{old-task} \LargeCurvature{} directions and the \emph{new-task} gradients, hindering their alignment. Nevertheless, we empirically observe that alignment emerges even when old- and new-task data differ drastically (\eg{} the old task is image classification and the new task is language analysis). 
(2) From the algorithmic implicit bias perspective, stochastic gradient descent for old-task training is biased towards flat minima, where only a few directions have high curvatures \citep{on_large-batch_training,wu_alignment_2022,jastrzebski_relation_2019,sagun_empirical_2018,he_control_2019}, as illustrated in \cref{fig:illustration_adversarial}.
This sparsity of \LargeCurvature{} directions makes it difficult to align with them in the extremely high-dimensional weight space as illustrated in \cref{fig:illustration_adversarial}.
Overall, this \spontaneous{} alignment implies a mysterious implicit adversariality of the new-task training: new-task updates automatically and accurately ``attack'' the most vulnerable but hard-to-locate components of the model's memory of old tasks, which we term as \emph{\AdversarialAlignment{}}.
We emphasize that the adversarial nature and difficulties of the alignment are missing in the previous understanding \citep{meat_learning_as_regularization,hessian_perspective_of_regularization_CL}, which we provide for a more complete picture on the alignment and catastrophic forgetting. 

Intrigued by this difficult-yet-occurring picture, we conduct theoretical analysis and trace the causes of the \AdversarialAlignment{} to the low-rank bias of model weight matrices induced by the old-task training.
These low-rank weight matrices yield low-rank Jacobians in deep networks, which take effect in the computations of the old-task's curvatures and new-task update gradients through forward and backward propagation, respectively.
They act as low-rank projections and pull the \LargeCurvature{} directions and new-task gradients to the same low-dimensional subspace, facilitating alignment as illustrated in \cref{fig:illustration_causes}.
Moreover, depth further intensifies the low-rankness of the projections and the alignment. 
This explains why the behavior of deep networks differs significantly from that of single-layer networks, since single-layer networks have full-rank Jacobians and it is hard for them to achieve \AdversarialAlignment{}, leaving forgetting solely determined by data properties. 
In contrast, adding just one hidden layer immediately introduces low-rankness and results in \AdversarialAlignment{}. Therefore, shallow-network forgetting is mainly governed by data distribution properties \citep{interest_in_forgetting_mechanism1,interest_in_forgetting_mechanism2,interest_in_forgetting_mechanism3,interest_in_forgetting_mechanism4,interest_in_forgetting_mechanism5,intreset_in_forgetting_mechanism6}, whereas deep-network forgetting is also driven by the additional implicit bias caused by low-rankness.

Our above theoretical results provide a principled framework to understand the effectiveness and limitations of existing \CL{} algorithms. 
Focusing on a representative family of forgetting-mitigating methods, Gradient Projection (GP) \citep{nscl,gpm,sgp}, we find them can also effectively mitigate the \AdversarialAlignment{}, but only in the forward propagation, leaving the alignment arising from the backward propagation intact. 
To address this issue, we propose a simple \backGP{} strategy, which further mitigates the alignment due to the backward direction by additionally confining the updates on weight matrices within the nullspace of gradients \wrt{} their outputs. Although conceptually as simple as replicating GP techniques in the backward direction, this modification has never been found in exiting GP methods \citep[Table 1]{forgetting_and_flatness2} without our finer-grained analysis on \AdversarialAlignment{}. This algorithm is plug-and-play and can be easily applied to existing GP methods.
Our extensive experiments show this simple modification effectively improves both forgetting mitigation and final performance by $8.1\%$ and $5.9\%$, respectively. When combined with plasticity-enhancing techniques, the improvement becomes $10.8\%$ less forgetting and $12.7\%$ more final performance.

Beyond the above theoretical analysis and its algorithm application, our results also exhibit broader impacts beyond \CL{}:
(1) It shows forgetting in \CL{} is catastrophic because it involves an adversarial attack, which has never been discovered before. 
This observation reveals the hidden connection between \CL{} and adversarial robustness \citep{implicit_power_iteration}, suggesting that understanding of adversarial samples can be transferred into that of catastrophic forgetting. 
(2) Furthermore, our analysis demonstrates how learning on one task can reshape the learning of subsequent tasks, \ie{} expressivity of deep networks is increased along directions that are important to the pretraining task and is decreased along non-important directions. This insight might inspire future works on studying task interactions in the pretraining-finetuning paradigm of modern foundation models, \eg{} understanding the effectiveness of parameter-efficient finetuning.

%% file: results/results.tex
\section{Results}\label{sec:results}

\input{preliminary.tex}

\input{results/existence.tex}

\input{results/mechanism.tex}

\input{results/application.tex}

%% file: preliminary.tex
\subsection{Preliminary}

We consider a simplified \CL{} scenario where only two tasks are involved, the old and the new tasks, denoted by subscripts $(\cdot)_1$ and $(\cdot)_2$.
The model is trained on the old task first and then trained on the new task. Since we intend to study catastrophic forgetting, the model is trained by vanilla gradient descent without any forgetting mitigation.
For task $t \in \set{1, 2}$, training samples are denoted by \emph{column} vectors $(\Input_{t}, \Label_{t}) \in \reals^{d_x} \times \reals^{d_y}$, or $(\EmpiricalInputs_t, \EmpiricalLabels_t) \in \reals^{d_x \times n_t} \times \reals^{d_y \times n_t}$ when stacked.
We use $\param \in \reals^{d_{\param}}$ for flattened model parameter, and $\param_t$ for parameter after task $t$'s training.
Let $\emploss_t: \reals^{d_{\param}} \to \reals$ denote the empirical loss, and let $\oldhessian$ denote the Hessian of the empirical loss on the old task.

%% file: results/existence.tex
\subsection{Existence of Adversarial Alignment}\label{sec:existence}

We first verify the existence of \AdversarialAlignment{} over a variety of \CL{} tasks and network architectures. To achieve this goal, we first obtain the projection $p_i$ of the new-task update $(\NewParam - \OldParam)$ on each eigenvector $\HessianEigVec_i$ to measure the alignment degrees. 
\begin{align}
    p_i \defeq \cos^2(\HessianEigVec_i, \NewParam - \OldParam) = \frac{\inner{\HessianEigVec_i}{\NewParam - \OldParam}^2}{\norm{\HessianEigVec_i}_2^2 \cdot \norm{\NewParam - \OldParam}_2^2}. 
\end{align}
We compare the new task update to the isotropic Gaussian random perturbation baseline.
We argue that if all the projections on all top eigenvectors of the new task update are large and the sum of them is disproportionately high compared to the number of \LargeCurvature{} directions, we regard \AdversarialAlignment{} as existing.

\begin{figure}
    \centering
    \newcommand{\ypad}{{\tiny \quad \quad \,\,}}%
    \newcommand{\xpad}{{\tiny \quad}}%
    \setstretch{0.05}%
    \begin{subfigure}{\linewidth}
        \centering 
        \includegraphics[width=0.2\linewidth]{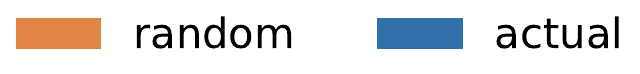}
    \end{subfigure} 
    \begin{subfigure}{0.72\linewidth}
        \begin{subfigure}{\linewidth}
            \labelsize%
            \begin{subfigure}{0.03\linewidth}%
                \rotatebox{90}{\labelsize \ypad Projection CDF}%
            \end{subfigure}%
            \begin{subfigure}{0.3\linewidth}
                \centering
                \,\, ResNet\\
                \includegraphics[width=\linewidth]{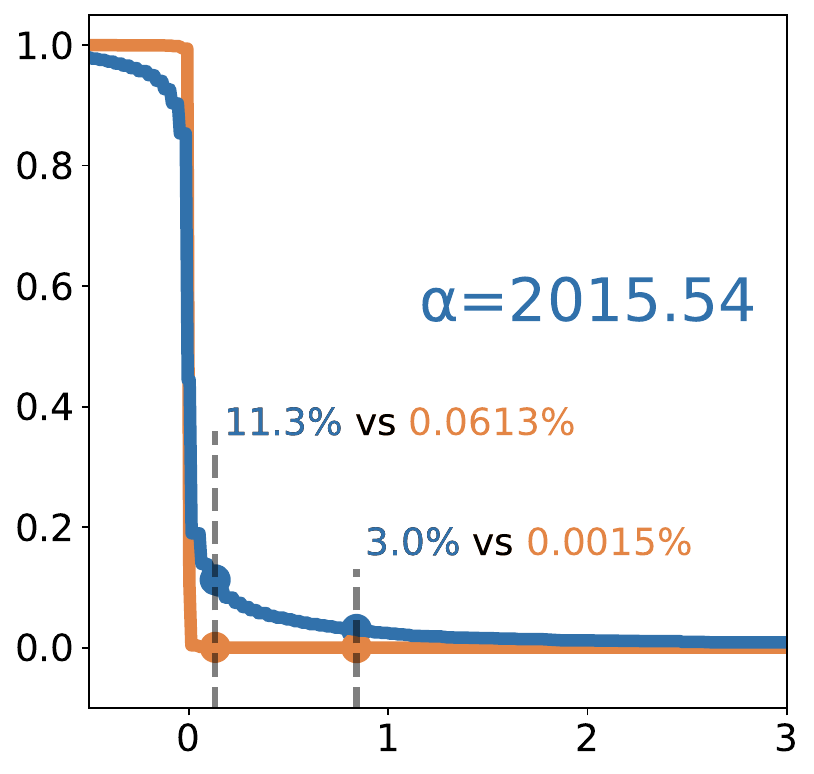}\\
                \vphantom{\labelsize \xpad Eigenvalue}
            \end{subfigure}%
            \begin{subfigure}{0.3\linewidth}
                \centering
                \quad\quad\quad\,\,  VisionTransformer\\
                \includegraphics[width=\linewidth]{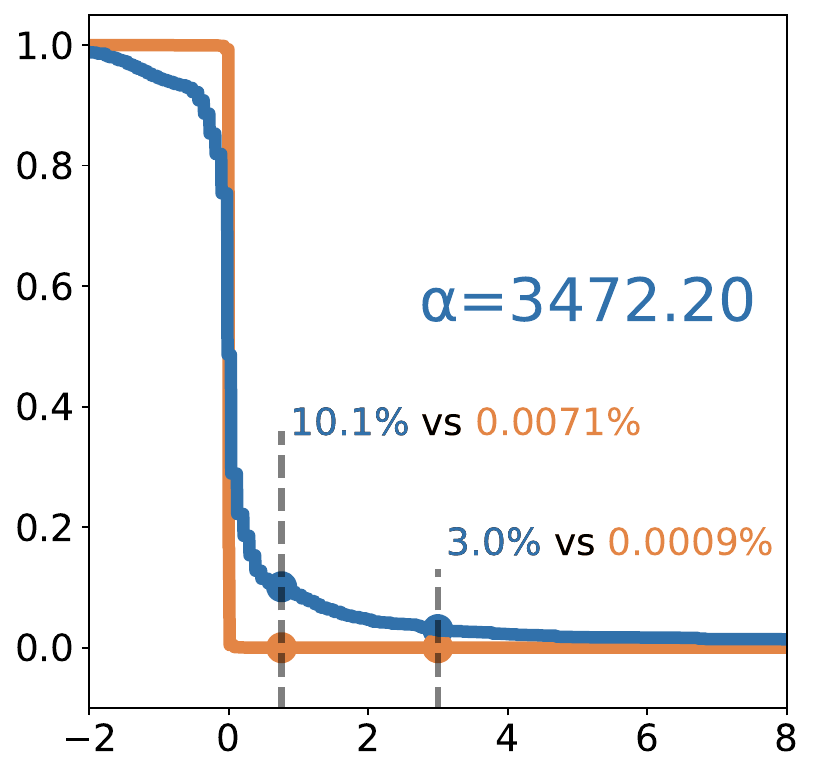}\\
                \labelsize \xpad Eigenvalue
            \end{subfigure}%
            \begin{subfigure}{0.3\linewidth}
                \centering
                \,\, MLP-Mixer\\
                \includegraphics[width=\linewidth]{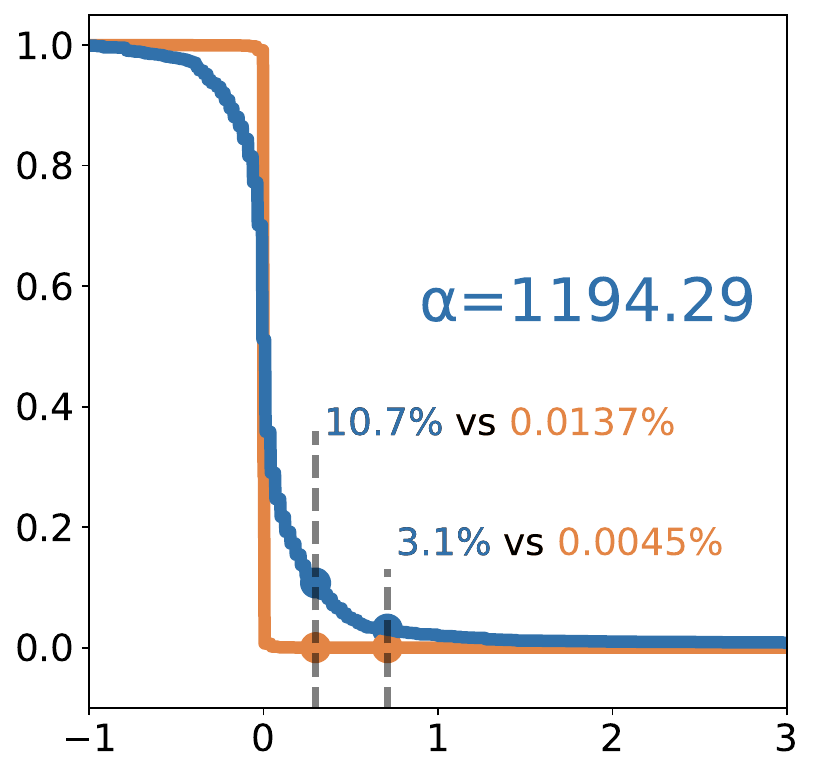}\\
                \vphantom{\labelsize Eigenvalue}
            \end{subfigure}\\
            \quad\\
            \quad\\
            \quad\\
            \quad\\
            \begin{subfigure}{0.03\linewidth}%
                \rotatebox{90}{\labelsize {\tiny \qquad\,\,\,}  $\alignment$}%
            \end{subfigure}%
            \begin{subfigure}{0.3\linewidth}
                \centering
                \includegraphics[width=\linewidth]{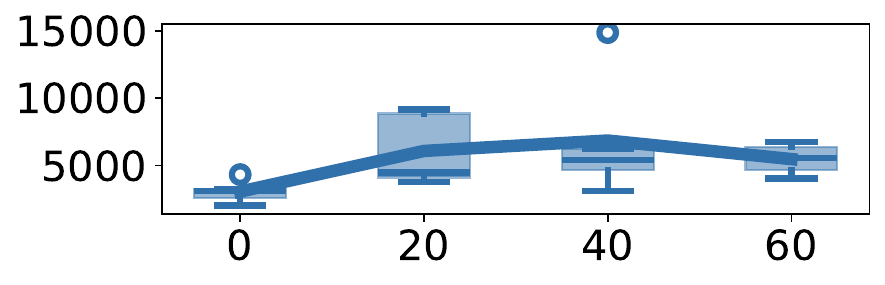}\\
                \vphantom{\labelsize New-task step}
            \end{subfigure}%
            \begin{subfigure}{0.3\linewidth}
                \centering
                \includegraphics[width=\linewidth]{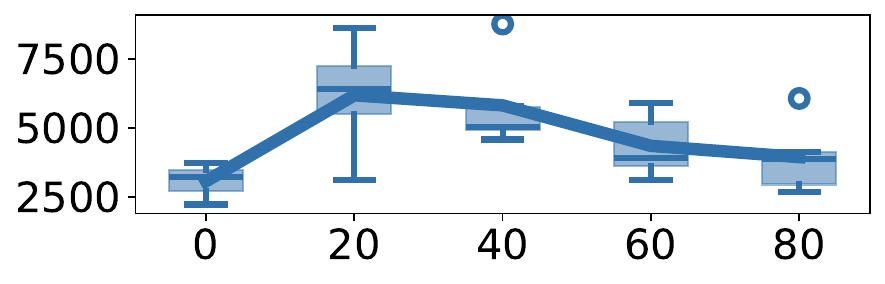}\\
                {\labelsize \quad\, \vphantom{xxxxxxxxxxx} New-task step}\\
            \end{subfigure}%
            \begin{subfigure}{0.3\linewidth}
                \centering
                \includegraphics[width=\linewidth]{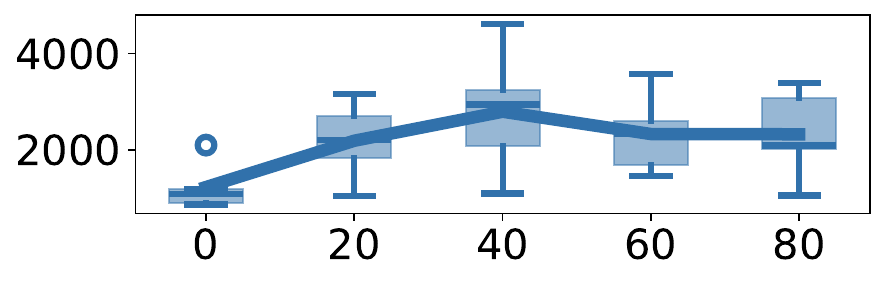}\\
                \vphantom{\labelsize New-task step}
            \end{subfigure}\\
            \caption{10-Split CIFAR100}\label{figure:existence_cifar100}\label{figure:alignment_changes_cifar100}
            \label{xxx}
        \end{subfigure}\\
        \begin{subfigure}{\linewidth}
            \labelsize
            \begin{subfigure}{0.03\linewidth}%
                \rotatebox{90}{\labelsize \ypad Projection CDF}%
            \end{subfigure}%
            \begin{subfigure}{0.30\linewidth}
                \centering
                DLN ($L=4$)
                \includegraphics[width=\linewidth]{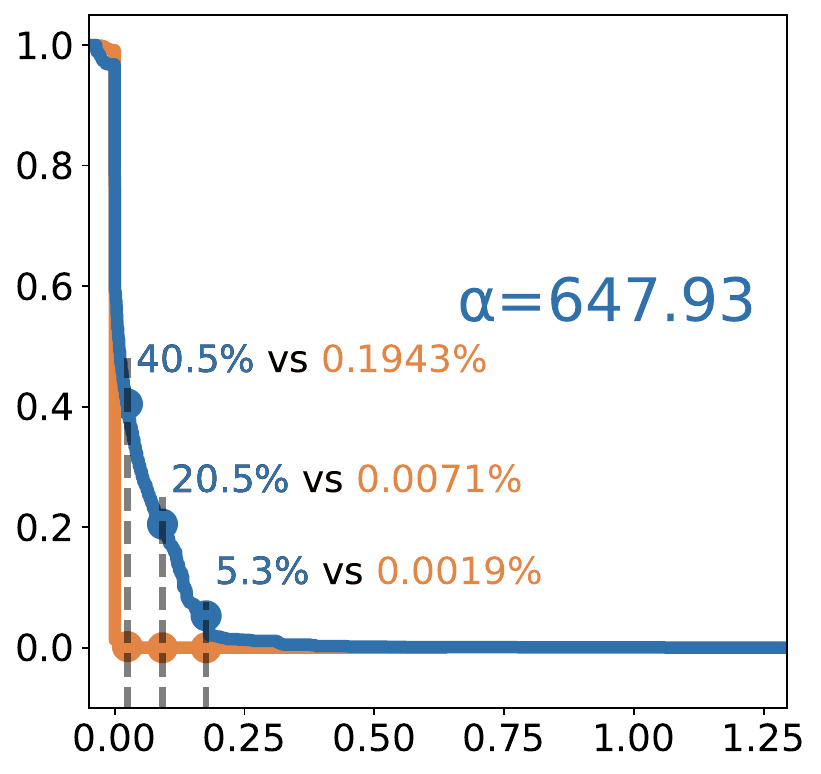}\\
                \vphantom{\labelsize Eigenvalue}
            \end{subfigure}%
            \begin{subfigure}{0.30\linewidth}
                \centering
                DLN ($L=6$)
                \includegraphics[width=\linewidth]{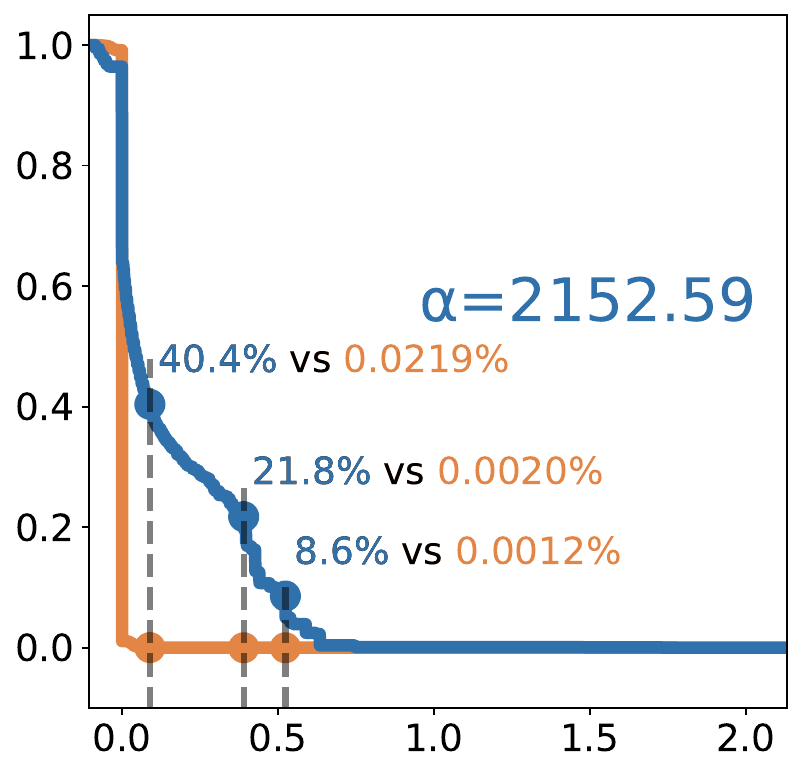}\\
                \labelsize \xpad Eigenvalue
            \end{subfigure}%
            \begin{subfigure}{0.30\linewidth}
                \centering
                DLN ($L=10$)
                \includegraphics[width=\linewidth]{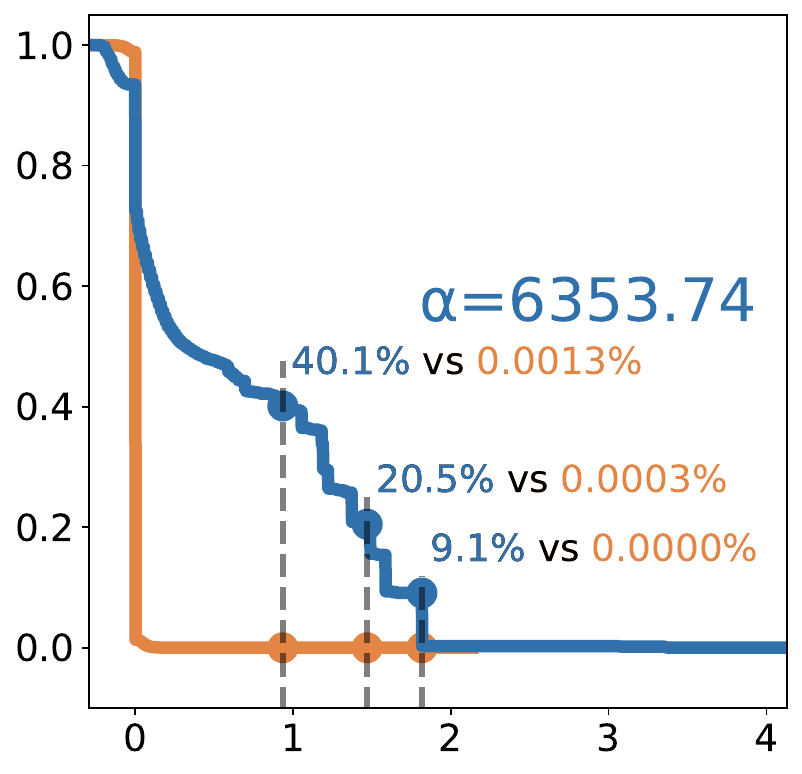}\\
                \vphantom{\labelsize Eigenvalue}
            \end{subfigure}\\
            \quad\\
            \quad\\
            \quad\\
            \quad\\
            \begin{subfigure}{0.03\linewidth}%
                \rotatebox{90}{\labelsize {\tiny \qquad\,\,\,} $\alignment$}%
            \end{subfigure}%
            \begin{subfigure}{0.30\linewidth}
                \centering
                \includegraphics[width=\linewidth]{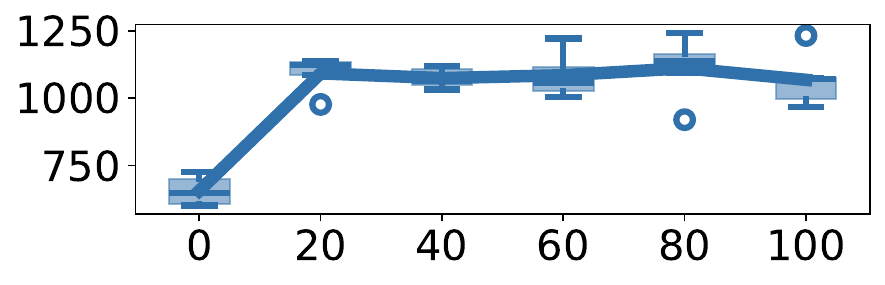}\\
                \vphantom{\labelsize New-task step}
            \end{subfigure}%
            \begin{subfigure}{0.30\linewidth}
                \centering
                \includegraphics[width=\linewidth]{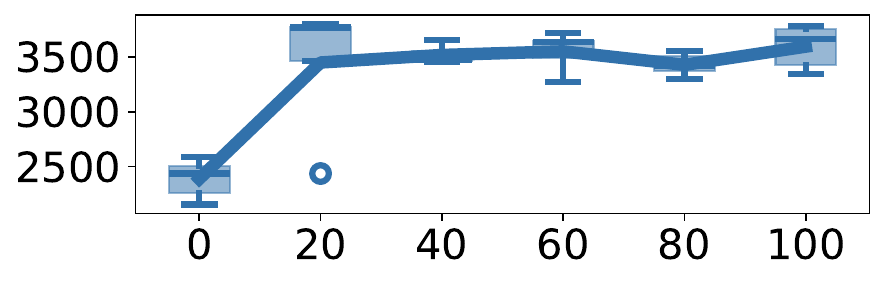}\\
                {\labelsize \quad\, \vphantom{xxxxxxxxxxx} New-task  step}\\
            \end{subfigure}%
            \begin{subfigure}{0.30\linewidth}
                \centering
                \includegraphics[width=\linewidth]{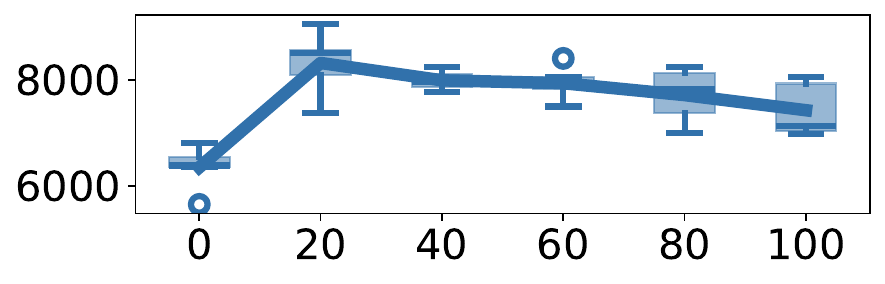}\\
                \vphantom{\labelsize New-task step}
            \end{subfigure}
            \caption{Synthetic Randomly Rotated Whitened MNIST}\label{figure:existence_randrot_mnist}\label{figure:alignment_changes_randrot_mnist}
        \end{subfigure}
    \end{subfigure}
    \renewcommand{\ypad}{{\tiny \quad \,\,}}
    \begin{subfigure}{0.25\linewidth}
        \centering
        \labelsize
        \quad \, VisionTransformer\\
        \rotatebox{90}{\labelsize \ypad Projection CDF}\includegraphics[width=0.864\linewidth]{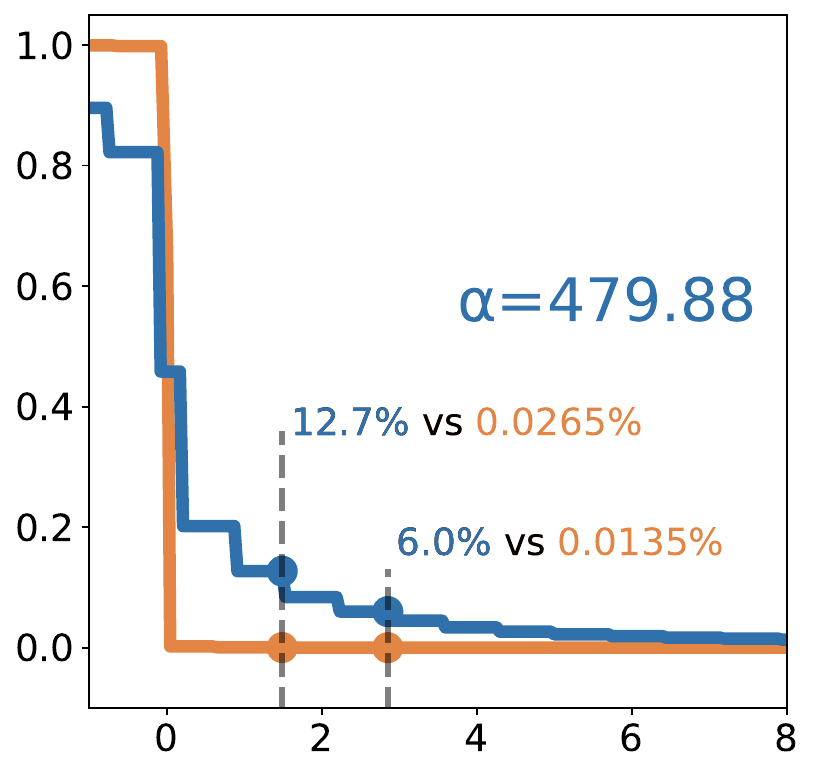}
        \labelsize Eigenvalue\\
        \quad\\
        \quad\\
        \quad\\
        \quad\\
        \rotatebox{90}{\labelsize \quad\,  $\alignment$}\includegraphics[width=0.864\linewidth]{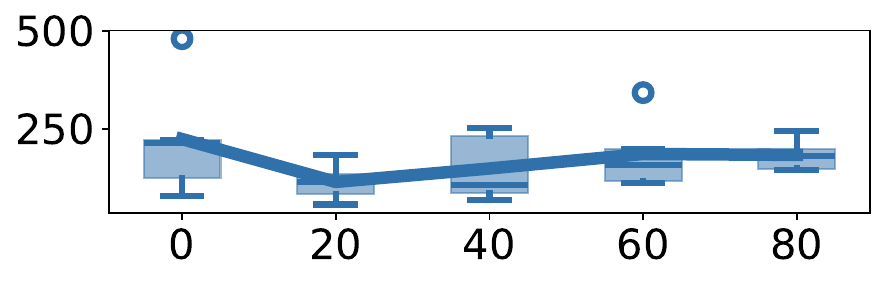}
        {\labelsize \quad\, \vphantom{xxxxxxxxxxx} New-task step}\\
        \quad\\
        \vphantom{\tiny Cross-Modal}\\
        \vphantom{\tiny Cross-Modal}\\
        \vphantom{\tiny Cross-Modal}\\
        \vphantom{\tiny Cross-Modal}\\
        \vphantom{\tiny Cross-Modal}\\
        \vphantom{\tiny Cross-Modal}\\
        \quad\\
        \quad\\
        \quad\\
        \quad \,\, MLP-Mixer\\
        \rotatebox{90}{\labelsize \ypad Projection CDF}\includegraphics[width=0.864\linewidth]{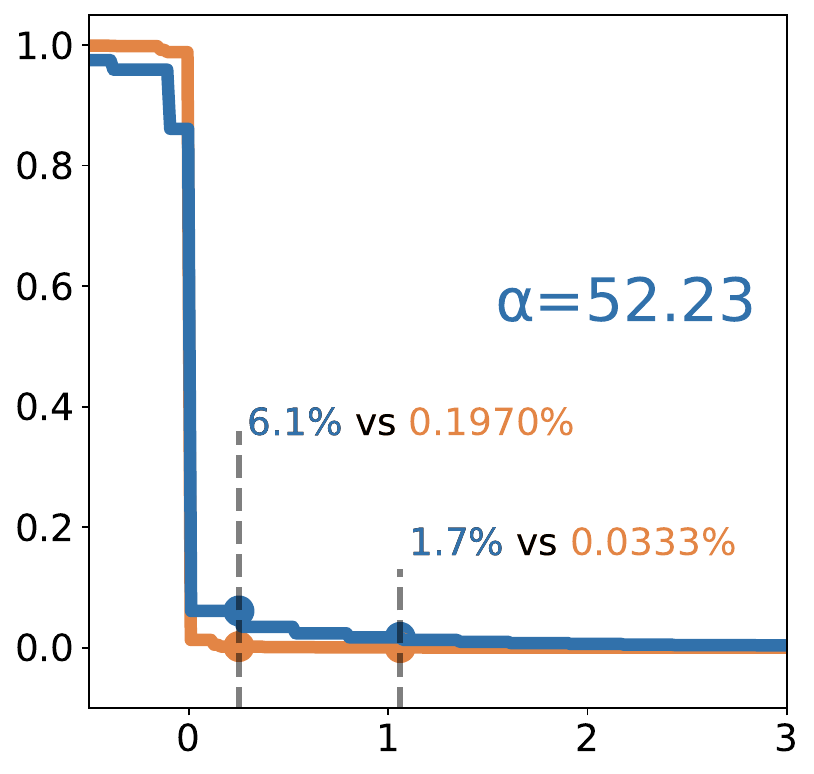}
        \labelsize \xpad Eigenvalue\\
        \quad\\
        \quad\\
        \quad\\
        \quad\\
        \rotatebox{90}{\quad \,  $\alignment$} \includegraphics[width=0.864\linewidth]{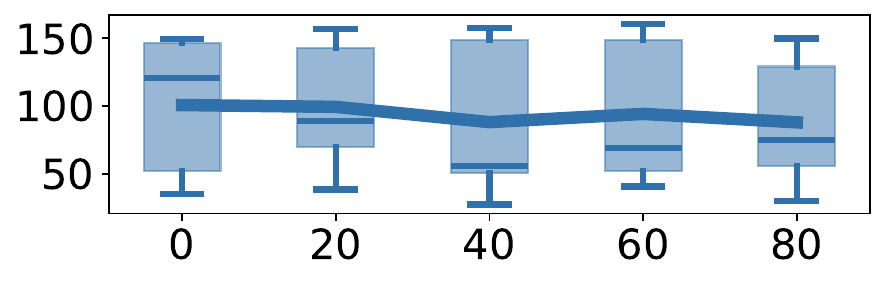}
        {\labelsize \qquad\qquad New-task step}\\
        \caption{Cross-Modal}\label{figure:existence_cross_modal}\label{figure:alignment_changes_cross_modal}
    \end{subfigure}
    \caption{
        \textbf{The empirical evidence of \AdversarialAlignment{}.}
        Cumulative distribution functions (CDFs, top) show that the projection of new-task updates is   disproportionately high onto high-curvature directions of old tasks across datasets and architectures, while random perturbations do not. Box plots (bottom) track the persistence of this alignment during the early steps of new-task training. Results are shown for (a) CIFAR-100 (10-split), (b) randomly rotated whitened MNIST (synthetic), and (c) cross-modal \CL{} (old task: first split of 10-split CIFAR100 for image classification, new task: SST2 for sentimental analysis). See \cref{sec:verification} for full details.
        We observe the new-task update has a large projection onto the eigenvectors of large curvatures $\sim 10^{0}$ compared to the baseline, even though such directions are sparse (see the baseline's flat CDF) and the tasks have different data (\eg{} cross-modal). 
        }\label{figure:existence}
\end{figure}
 
\cref{figure:existence} presents the empirical results. We plot the cumulative distribution function (CDF) of $p_i$, which measures how much the new-task updates project onto the old-task eigenvectors.
Since the CDF of random perturbations is flat, they never align with the high-curvature directions. Instead, according to repeated experiments across different \CL{} settings and architectures, we find that nearly 10\% of the updates align with the top 0.06\% of high-curvature directions, which is extremely sparse. It further confirms that new-task updates strongly align with the most sensitive directions of the old task, showing the adversarial nature of the \AlignmentPhenomenon{}.

To better understand the evolution of \AdversarialAlignment{} at each step, we further quantify the degree of \AlignmentPhenomenon{} by
\begin{align}
    \alignment(\mat{A}, \vec{r}) 
    \defeq \dim \vec{r} \cdot \frac{\ex{\vec{r}^\top \mat{A} \vec{r}}}{\trace{\mat{A}} \cdot \mathbb{E} \norm{\vec{r}}_2^2}, \label{eq:alignment_definition}
\end{align}
where $\mat{A}$ is a symmetric matrix and $\vec{r}$ is a random or deterministic vector.
The larger $\alignment$ is, the more \AdversarialAlignment{} is. 
See \cref{sec:definition} for its derivation.
The box diagrams in \cref{figure:existence} show the evolution of \AdversarialAlignment{} during the first $80$ steps of new-task training.
We observe \AdversarialAlignment{} maintains a large magnitude, and in non-cross-modal tasks, it even increases in the early stage, indicating that \AdversarialAlignment{} is a persistent phenomenon.
 
\begin{figure}
    \newcommand{\ypad}{{\tiny \quad \quad}}
    \newcommand{\xpad}{{\tiny \hphantom{xx}}}
    \setstretch{0.2}%
    \begin{center}%
        \includegraphics[width=0.5\linewidth]{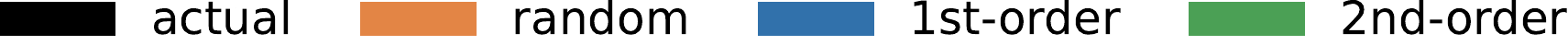}%
    \end{center}%
    \begin{subfigure}{0.72\linewidth}
        \begin{subfigure}{\linewidth}
            \labelsize
            \begin{subfigure}{0.03\linewidth}%
                \rotatebox{90}{\labelsize \ypad Forgetting}%
            \end{subfigure}%
            \begin{subfigure}{0.3\linewidth}
                \centering
                \quad ResNet\\
                \includegraphics[width=\linewidth]{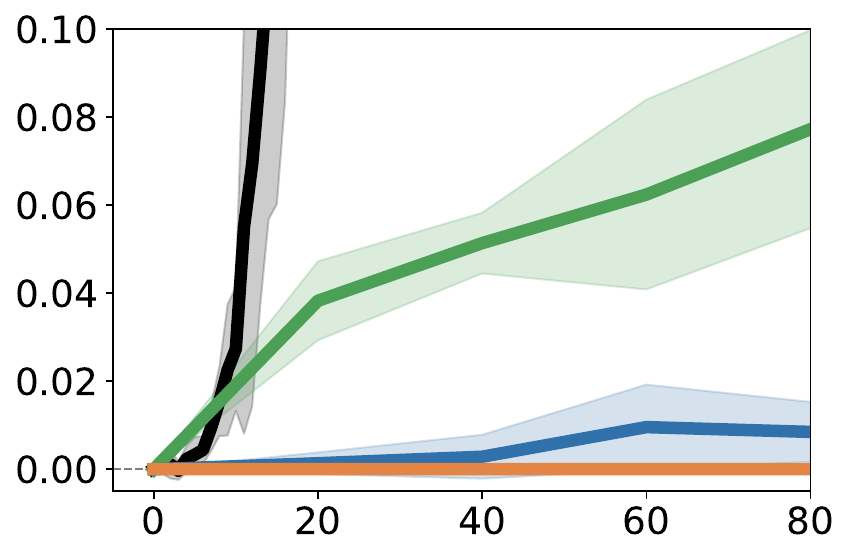}\\
                \labelsize \vphantom{New-task step}
            \end{subfigure}
            \begin{subfigure}{0.3\linewidth}%
                \centering
                \quad \quad \,\, VisionTransformer
                \includegraphics[width=\linewidth]{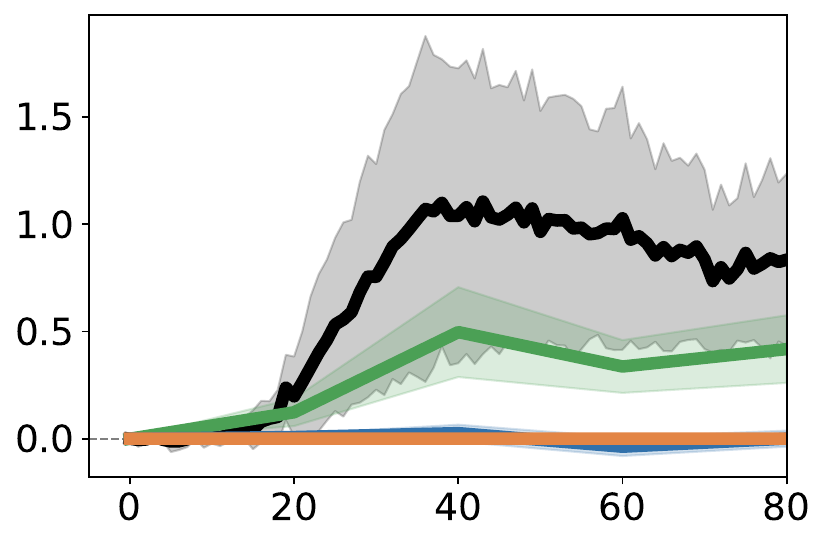}
                \labelsize \xpad New-task step
            \end{subfigure}
            \begin{subfigure}{0.3\linewidth}
                \centering
                \quad MLP-Mixer
                \includegraphics[width=\linewidth]{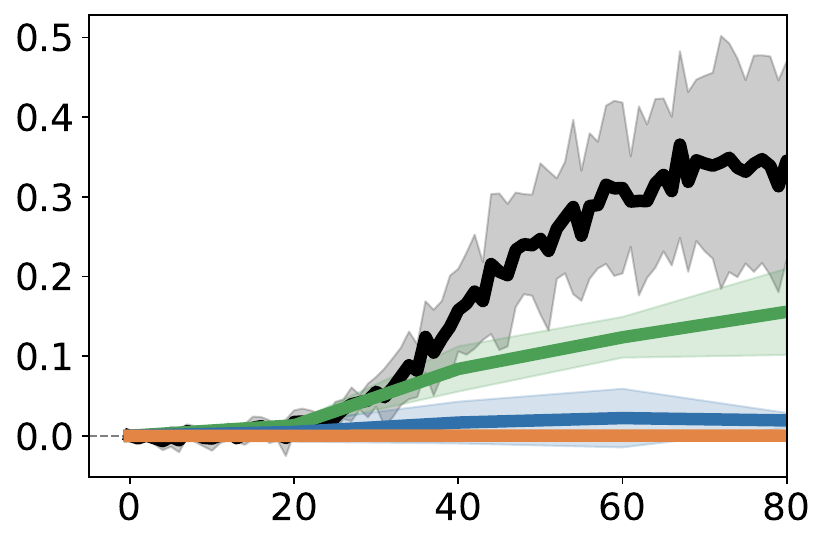}
                \labelsize \vphantom{New-task step}
            \end{subfigure}
            \caption{10-Split CIFAR100}\label{figure:all_forgettings_cifar100}
        \end{subfigure}
        \begin{subfigure}{\linewidth}
            \labelsize
            \begin{subfigure}{0.03\linewidth}%
                \rotatebox{90}{\labelsize \ypad Forgetting}%
            \end{subfigure}%
            \begin{subfigure}{0.316\linewidth}
                \centering
                DLN $(L=4)$
                \includegraphics[width=\linewidth]{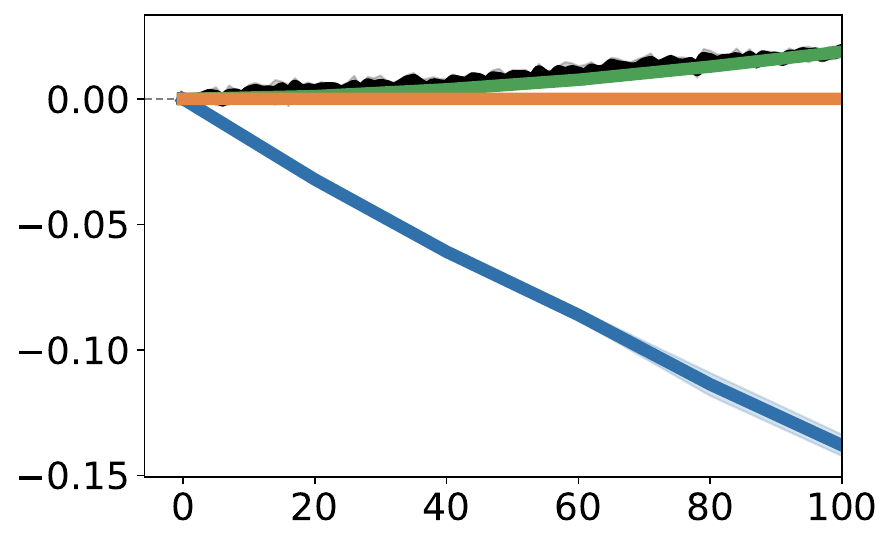}\\
                \labelsize \vphantom{New-task step}
            \end{subfigure}%
            \begin{subfigure}{0.3\linewidth}%
                \centering
                DLN $(L=6)$
                \includegraphics[width=\linewidth]{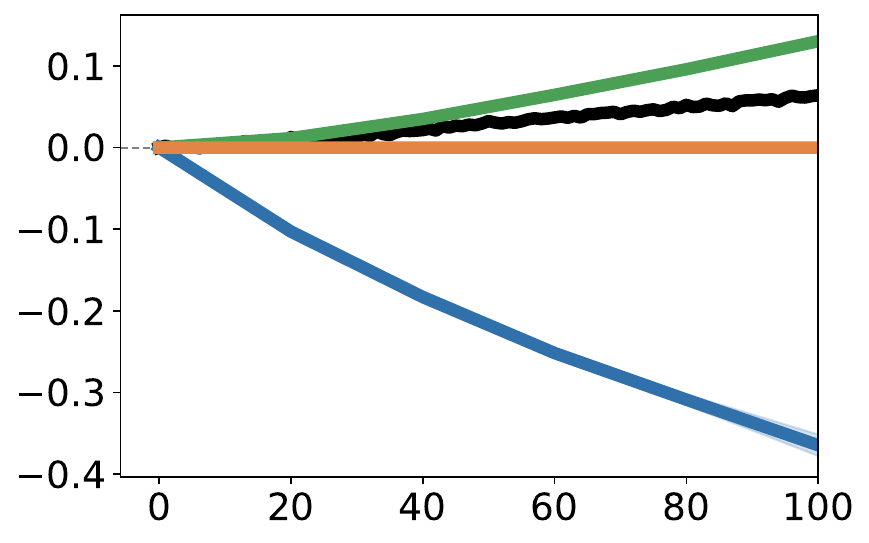}\\
                \labelsize \xpad New-task step%
            \end{subfigure}%
            \begin{subfigure}{0.3\linewidth}%
                \centering
                DLN $(L=10)$\\
                \includegraphics[width=\linewidth]{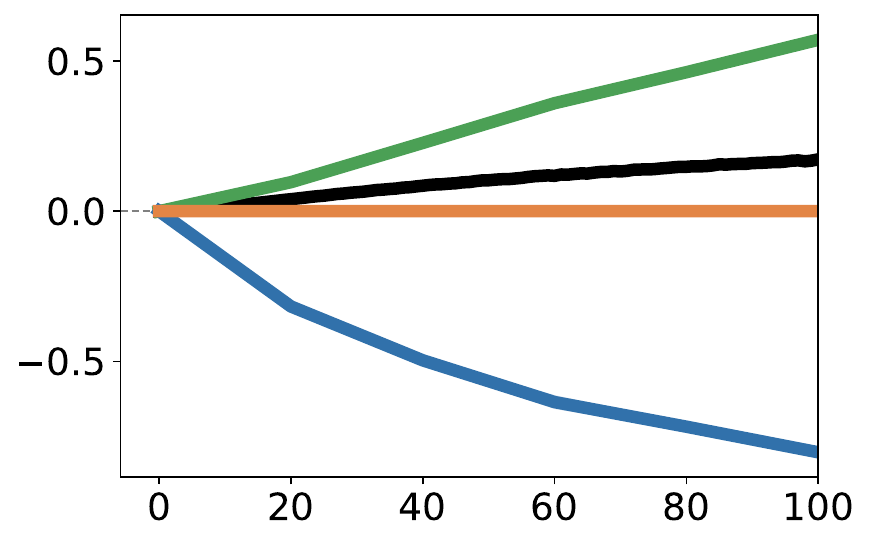}\\
                \labelsize \vphantom{New-task step}
            \end{subfigure}
            \caption{Synthetic Randomly Rotated MNIST.}\label{figure:all_forgettings_randrot_mnist}
        \end{subfigure}
    \end{subfigure}
    \renewcommand{\ypad}{{\tiny \quad \,}}
    \begin{subfigure}{0.25\linewidth}
        \centering
        \labelsize
        \hphantom{xxxxxxxxx} VisionTransformer
        \rotatebox{90}{\labelsize \ypad Forgetting}\includegraphics[width=0.92\linewidth]{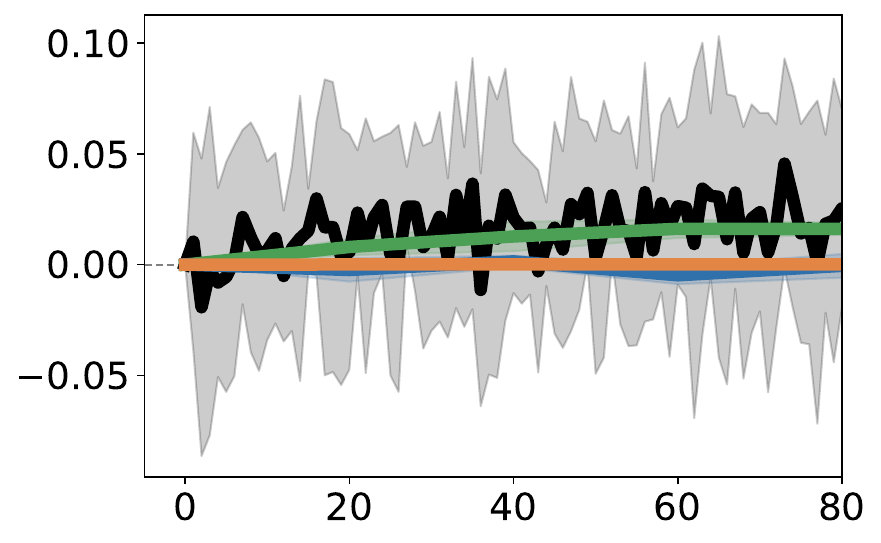}\\
        \labelsize \xpad New-task step\\
        \vphantom{\small Cross-Modal}\\
        \vphantom{\small Cross-Modal}\\
        \vphantom{\small Cross-Modal}\\
        \hphantom{xxx} MLP-Mixer
        \rotatebox{90}{\labelsize \ypad Forgetting}\includegraphics[width=0.92\linewidth]{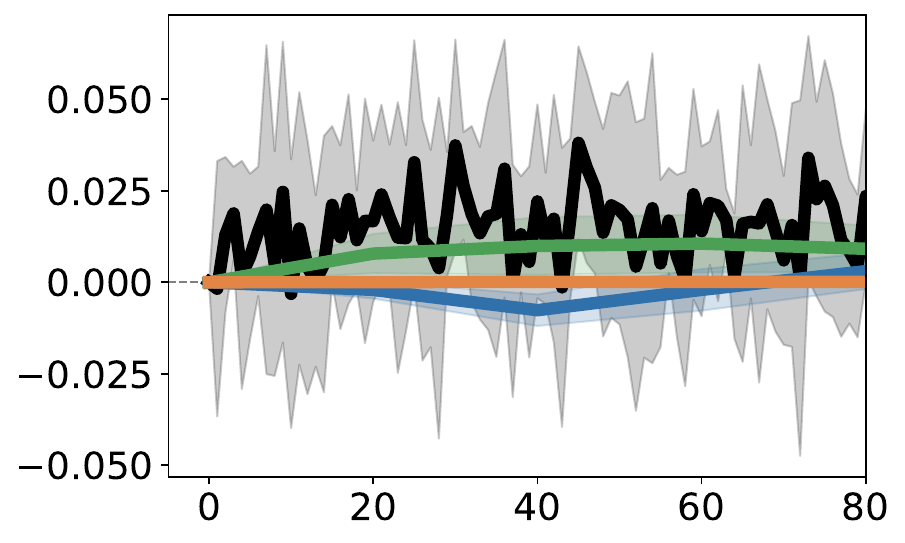}
        \labelsize \xpad New-task step
        \caption{Cross-Modal}\label{figure:all_forgettings_cross_modal}
    \end{subfigure}
    \caption{\textbf{Connection between \AdversarialAlignment{} and forgetting.}
        We present the various forgetting (old-task loss increase) recorded during the experiments in \cref{figure:existence}.
        Actual forgetting (black) rises sharply with new-task training. Its second-order approximation (green) can capture this rise especially at initial new-task training, while random perturbations (orange) induce negligible forgetting. First-order approximations (blue) capture little of the effect or even predict negative forgetting. Average results over 5 runs are reported. The experimental settings are the same as \cref{figure:existence}, and the results are also recorded in the experiment for \cref{figure:existence}. Full details can be found in \cref{sec:verification}.
    }\label{figure:all_forgettings}
\end{figure}

\subsection{Influence of Adversarial Alignment on Forgetting}\label{sec:alignment_and_forgetting}

Adversarial alignment is directly connected to forgetting: if old-task weight $\OldParam$ is sufficiently trained to be a local minimum, forgetting can be decomposed into
\begin{align}
    \OldLoss(\NewParam) - \OldLoss(\OldParam)
    \approx&  \frac{1}{2} \cdot \underbrace{\alignment(\oldhessian, \Delta \param)}_{\substack{\text{adversarial}\\\text{alignment}}} \cdot \underbrace{\norm{\Delta\param}_2^2}_{\substack{\text{update} \\ \text{magnitude}}} \cdot \underbrace{\ex[\vecperturbation \sim \mathcal{N}(\vec{0}, \vec{I} / \dim \param)]{\vecperturbation^\top \oldhessian \vecperturbation}}_{\substack{\text{robustness against} \\ \text{random perturbation}}}, \label{eq:alignment_and_forgetting}
\end{align} 
where $\Delta \param \defeq \NewParam - \OldParam$ is the new-task update.
See \cref{proposition:forgetting_decomposition} in \supplementary{} for the formal result. 
This decomposition first indicates that random perturbations can lead to small but non-zero forgetting, and \AdversarialAlignment{} amplifies it to catastrophic forgetting (\eg{}, as shown from the box diagrams in \cref{figure:existence}, the \AlignmentPhenomenon{} amplifies the catastrophic forgetting by an order of $10^{3}$).
The amplification is achieved by accurately biasing the new-task updates to the most sensitive directions, \ie{} the \LargeCurvature{} directions.
Empirically, by further comparing the (original, first-order, and second-order approximated) forgetting induced by model updates and random perturbations in \cref{figure:all_forgettings}, we find that without \AdversarialAlignment{} (\ie{} if the new-task updates become random), forgetting would be negligible.  
Therefore, \AdversarialAlignment{} is crucial to forgetting, and removing the adversariality may drastically reduce forgetting.
However, the cause underlying its emergence is poorly understood. The rest of this paper aims for this understanding.

%% file: results/mechanism.tex
\subsection{Cause of Adversarial Alignment}\label{sec:mechanism}

\subsubsection{Ruling out Trivial Explanations}

To find the cause of adversarial alignment, we first systematically examine important components in deep learning and \CL{}: data, training algorithm, architecture, and model weight. 
Adversarial alignment can be seen as a correlation between the old- and new-task properties. It requires the information on the old task training to be transferred to the new task so that the training of new task can accurately ``attack'' the old-task knowledge. Since architecture or training algorithms are essentially memoryless across tasks, they cannot support such information channel to achieve alignment.

The data similarity between the old and new tasks may be responsible for the correlation. To test whether the phenomenon is fully driven by data, we either decrease data similarity by cross-modal \CL{} tasks in \cref{figure:existence_cross_modal}, or synthesize \CL{} tasks where the old task is whitened MNIST and the new task is generated by randomly rotating the old-task input vectors, which can eliminates hidden data similarity at least for linear regression \citep{interest_in_forgetting_mechanism4,interest_in_forgetting_mechanism5}. However, despite the difficulties, \AdversarialAlignment{} still exists.
We also vary the depth of deep linear networks and find that deeper models have larger alignment, even though data similarity remains the same. 
Therefore, data similarity cannot fully explain \AdversarialAlignment{} and there are non-data mechanisms. 

Another intuitive explanation is the accidental alignment, \ie{} there are a moderate number of \LargeCurvature{} directions, so that new-task updates can align with them accidentally. \Cref{figure:existence} shows \LargeCurvature{} directions' distribution is not moderate but sparse, and special biases or correlations are required to align with them.

Overall, the only remaining component is the model weight, which can be passed from the old task to the new task and is the only hidden information channel supporting the correlation between the new- and old-task training. 

\subsubsection{Adversarial Alignment is Caused by Implicit Bias of Low-Rankness}

\begin{figure}
    \centering
    \includegraphics[width=0.99\linewidth]{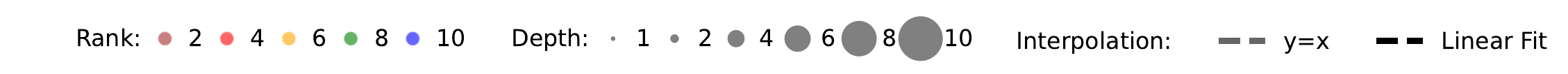}
    \begin{subfigure}{0.23\linewidth}
        \centering\setstretch{0.2}
        \begin{subfigure}{\linewidth}
            \rotatebox{90}{\setstretch{0.2}\labelsize \qquad \quad \quad $\tightbound$}\includegraphics[width=0.9\linewidth]{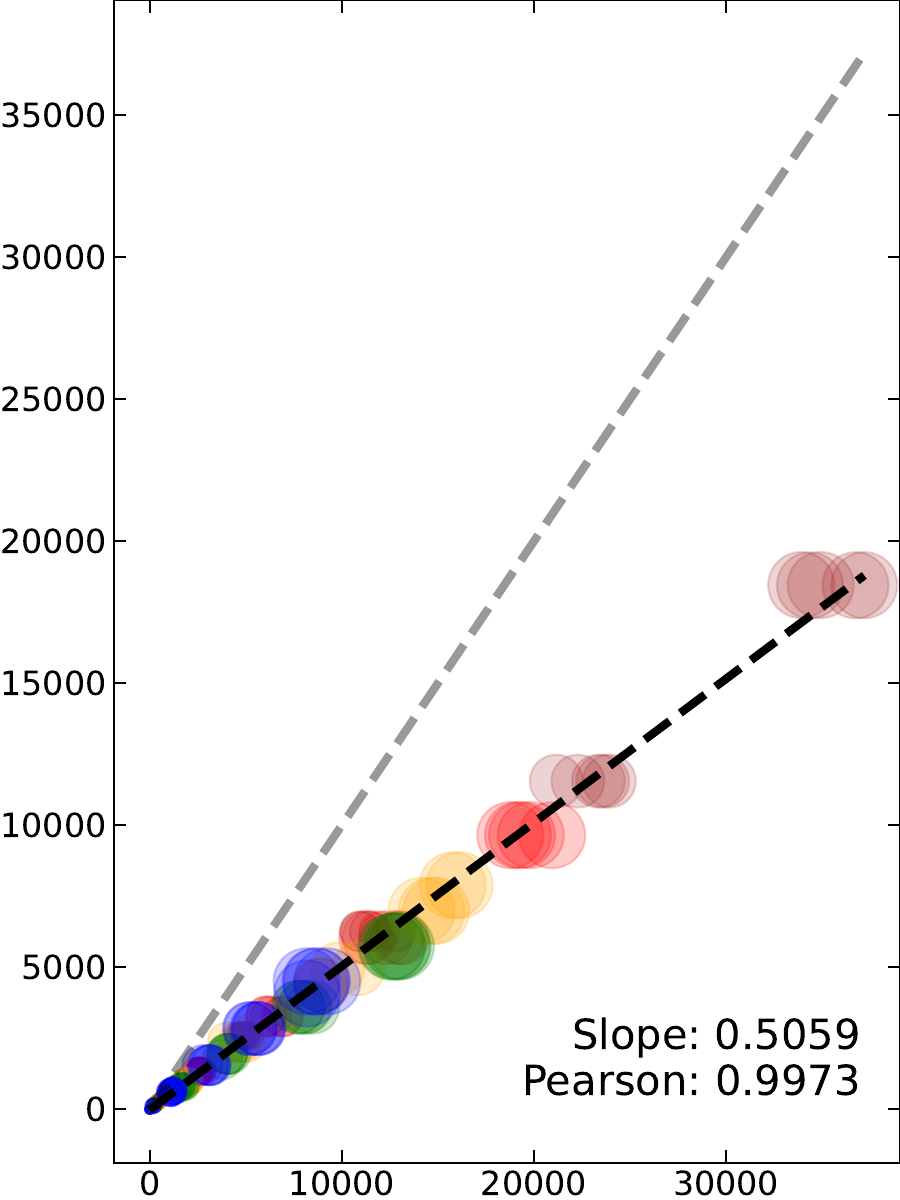}
        \end{subfigure}\\
        {\labelsize \qquad \qquad \qquad \qquad \qquad \qquad \hphantom{xxx} Estimated $\alpha$}
        \caption{Tighness.}\label{figure:verification_tightness}
    \end{subfigure} 
    \NewDocumentCommand{\TransitionSubfigure}{m m O{}}{
        \begin{subfigure}{0.23\linewidth}
            \centering\setstretch{0.2}
            \begin{subfigure}{\linewidth}
                \begin{flushright}
                \rotatebox{90}{\labelsize  \qquad \,\, $\alignment$} \includegraphics[width=0.91\linewidth]{pic/main/phase_transition/actual_transition_#1.pdf}
                \end{flushright}
            \end{subfigure}
            \begin{subfigure}{\linewidth}
                \begin{flushright}
                \rotatebox{90}{\setstretch{0.0} \labelsize \quad $\tightbound$} \includegraphics[width=0.88\linewidth]{pic/main/phase_transition/theoretical_transition_#1.pdf}
                \end{flushright}
            \end{subfigure}\\
            {\labelsize \hphantom{xx} Rank}
            \caption{$L=#2$}\label{#3}
        \end{subfigure}
    }
    \TransitionSubfigure{1}{1}[figure:phase_transition_1]
    \TransitionSubfigure{2}{2}[figure:phase_transition_2]
    \TransitionSubfigure{46810}{4,6,8,10}[figure:phase_transition_46810]
    \caption{
        \textbf{Verification of \AdversarialAlignment{} lower-bounds.}
        \cref{figure:verification_tightness} shows the correlation between the lower-bound and the estimated $\alpha$ in each experiment. It shows the lower-bound (1) is lower than the estimated $\alignment$, (2) is well correlated with the actual $\alignment$, and (3) is tight up to constant factors within the scope of the experiments.
        \cref{figure:phase_transition_1,figure:phase_transition_2,figure:phase_transition_46810} verify the phase transition predicted by the lower-bound. The experiments are conducted on the whitened MNIST dataset with random rotation of the old task as the new task. The rank is controlled by taking labels modulo rank $r$.
        When $L=1$, the alignment is not related to the rank of $\OldGroundTruth$.
        When $L\ge 2$, the alignment is inversely proportional to the rank of $\OldGroundTruth$. For each depth-rank configuration, we run experiments 5 times. The 10-rank results are recorded in experiment for \cref{figure:existence_randrot_mnist}.
    }
\end{figure}

In this section, we seek the theoretical understanding of the hidden channel provided by the model weight. 
To this end, we focus on technically feasible tasks, \ie{} regression using deep linear networks (DLN) of depth $L$, which is defined by $f_{\param}(\Input) \defeq \weight_{L} \weight_{L-1} \cdots \weight_{1} \Input$, where $\weight_{i} \in \reals^{\dim \Input \times \dim \Input}$ and $\param \defeq \begin{bmatrix}
    \vectorize{\weight_i}
\end{bmatrix}_i$. 
We also assume the standard $\ltwo$ regularization and the old-task parameter $\OldParam$ is well-trained under the regularized old-task, so that it is a local minimum of the regularized empirical loss. 
To eliminate the data similarity factor, we employ the same data generation process as \cref{figure:existence_randrot_mnist}, \ie{} whitened old-task data and new-task data generated by random rotation of old-task data.
We derive the expression of \AdversarialAlignment{} at the first step of the new-task training with several simplifications and arrive at the following lower-bound:
\begin{align}
    \alignment(\oldhessian, \nabla_{\param} \emploss_2(\OldParam)) \gtrapprox  \frac{\dim \param}{2 \dim \Input \cdot \erank{\SingularMatrix{\OldLabels \OldInputs^\top}^{2 (1 - 1/L)}}}.\label{eq:lowerbound_interpretable}
\end{align}
Here, $\mat{\Sigma}_{\OldLabels \OldInputs^\top}$ is the singular value matrix of the old input-output correlation $\OldLabels \OldInputs^\top \defto \OldGroundTruth$. Effective rank $\erank{\cdot}$ is a soft rank defined in \cref{sec:definition_erank} that reflects the concentration of the spectrum. 
See \cref{theorem:alignment_lowerbound_interpretable} in \supplementary{} for the formal statement.
We also derive a tighter but more complicated bound $\alignment \gtrapprox \tightbound$ and verify its tightness in \cref{figure:verification_tightness}.
See \cref{eq:lowerbound_tighter} in \cref{sec:assumptions} for $\tightbound$ and see \cref{theorem:alignment_lowerbound_tighter} in \supplementary{} for the formal statement. Results with relaxed assumptions can be found in \cref{appendix:relexation}.

From \cref{eq:lowerbound_interpretable}, we conclude that it is low-rankness and depth that induce \AdversarialAlignment{} in DLNs. Specifically, the low-rankness encourages \AdversarialAlignment{} since it is inversely proportional to the rank of the powered $\OldGroundTruth$. 
In addition, when the network becomes deeper, $\SingularMatrix{\OldGroundTruth}^{2(1-1/L)}$ has a larger exponent, resulting in exponentially faster increases of top singular values than small singular values.
Therefore, the spectrum of $\SingularMatrix{\OldGroundTruth}^{2(1-1/L)}$ concentrates more at the large singular values, making it lower-rank and intensifying \AdversarialAlignment{}. 

Interestingly,  we observe that a  phase transition of \AdversarialAlignment{} happens when depth increases from $L=1$ to $L=2$, as shown in \cref{figure:phase_transition_1,figure:phase_transition_2,figure:phase_transition_46810}.
When $L=1$, we have $\SingularMatrix{\OldGroundTruth}^{2(1 - 1/L)} = \mat{I}$, whose rank is $\dim \Input$ and the $\alignment$ is minimal and unrelated to the rank of $\OldGroundTruth$.
When $L \ge 2$, $\SingularMatrix{\OldGroundTruth}^{2(1 - 1/L)}$ has an exponent $\ge 1$, making $\alignment = \Omega\left(\frac{1}{\erank{\OldLabels \OldInputs^\top}}\right)$.
It indicates that depth is a key factor in the \AdversarialAlignment{}, and the catastrophic forgetting in deep networks is totally different from single-layer networks.

\subsubsection{How Low-Rankness Leads to Adversarial Alignment}\label{sec:interpretation}

\newcommand{\vg}{\vec{g}}
\newcommand{\mG}{\mat{G}}
\newcommand{\Signal}{\mat{\Sigma}_{\text{signal}}}
\NewDocumentCommand{\JacobianSV}{m m O{#1 - #2} O{\Signal^{#3}}}{
    \begin{bmatrix}
        #4 & \hphantom{\approx}\mat{0} \\
        \mat{0} & \nearzero 
    \end{bmatrix}
}
\NewDocumentCommand{\SimplifiedConsecutiveProduct}{m m O{#1 - #2} O{\Signal^{#3}}}{\mU_{#1} \JacobianSV{#1}{#2}[#3][#4] \mV_{#2}^\top}
\NewDocumentCommand{\TransposedSimplifiedConsecutiveProduct}{m m O{#1 - #2} O{\Signal^{#3}}}{\mV_{#2} \JacobianSV{#1}{#2}[#3][#4] \mU_{#1}^\top}

\newcommand{\transparentUnderbrace}[2]{\smash{\underbrace{#1}_{\scalebox{0.6}{\text{#2}}}}\vphantom{#1}}
\newcommand{\transparentOverbrace}[2]{\smash{\overbrace{#1}^{\centering\scalebox{0.6}{\text{#2}}}}\vphantom{#1}}
\newcommand{\nonTransparentUnderbrace}[2]{\underbrace{#1}_{\scalebox{0.6}{\text{#2}}}}

\ifExplicitHessian
    Although we have found that low-rankness and depth lead to the \AdversarialAlignment{}, the detail of this process is still unclear.
    To explicitly understand it, we revisit the definition (\ref{eq:alignment_definition}) of alignment and analyze the key steps when proving \cref{eq:lowerbound_interpretable}, \ie{} computing the new-task gradient $\vg$ and its quadratic form with the old-task Hessian $\oldhessian$:
    \begin{align}
        \vg 
        =& \Diag\left(\nonTransparentUnderbrace{\weight_{i-1:1}}{new forward} \kronecker \nonTransparentUnderbrace{\weight_{L:i+1}^\top}{new backward}\right) \times \left(\vec{1} \kronecker \vectorize{(\weight_{L:1} \NewInputs - \NewLabels) \NewInputs^\top}\right)\\
        \defto& \mJ \times \left(\vec{1}_L \kronecker \vectorize{\partialfrac{\NewLoss}{f_{\OldParam}} \NewInputs^\top}\right),\\ 
        \ex{\vg^\top \oldhessian \vg}
        =&\Fnorm{ \sum_{i=1}^L \weight_{L:i+1} \mG_i \weight_{i-1:1} \OldInputs}^2
        = \exx \norm{\Diag \left((\OldInputs^\top \nonTransparentUnderbrace{\weight_{i-1:1}^\top}{\text{old forward}}) \kronecker \nonTransparentUnderbrace{\weight_{L:i+1}}{\text{old backward}}\right) \times \vg}^2\\
        \defto& \exx \norm{ (\mI_L \kronecker (\OldInputs \kronecker \mI_{\dim \Input}))\times  \mJ^\top \times \vg}^2,
    \end{align}
    where $\kronecker$ denotes Kronecker product, $\Diag(\cdot)$ constructs block-diagonal matrices,
    \begin{align}
        \mJ \defeq \Diag\left(\transparentUnderbrace{\weight_{i-1:1}}{old/new forward} \kronecker \transparentUnderbrace{\weight_{L:i+1}^\top}{old/new backward}\right).
    \end{align}
    is the Jacobian, $\mG_i \defeq \partialfrac{\NewLoss}{\weight_i}$ is the matrix-shaped new-task gradient \wrt{} the $i$-th layer's weight, $\vg \defeq \begin{bmatrix}
        \vectorize{\mG_i}
    \end{bmatrix}_i$ is the flattened new-task gradient vector, $\weight_i$ is the $i$-th weight immediately after the training of old task, $\vec{1}_L \in \reals^{L}$ is the all-one vector. 
    $\weight_{b:a} \defeq \weight_{b} \weight_{b-1} \cdots \weight_{a}$ denotes a consecutive product of weight matrices, which \textit{come from both the forward and backward propagations}. 
    The above equations reveal the possible distributions of new-task gradients as well as the old-task \LargeCurvature{} directions are confined to the low-dimensional principal subspace of $\mJ$'s column space.

    To see how small the subspace is and how strict the confinement is, we need finer-grained properties of $\mJ$, which is controlled by old-task weights. We find that under the commonly applied $\ltwo$ regularization, the old-task weights at all layers become low-rank with the same low-rank singular values $\mat{\Sigma} = \begin{bmatrix}
            \Signal & \mat{0} \\
            \mat{0} & \mat{\Sigma}_{\text{nuisance}}
    \end{bmatrix}$ with $\min{\Signal} \gg \max{\mat{\Sigma}_{\text{nuisance}}}$ and the same ``adjacent'' singular vectors (see \cref{lemma:auto_alignment_implication_singular_values}). 
    As a result, when the network is deep, for middle layer $i$, the two consecutive weight products become low-rank with the form $\weight_{L:i+1} = \SimplifiedConsecutiveProduct{L}{i-1}[L-i], 
    \weight_{i-1:1} = \SimplifiedConsecutiveProduct{i-1}{1}[i-1]$, where ``$\nearzero$'' denotes matrices close to zero.
    Therefore, $\mJ$ comprises of a lot of low-rank matrices. According to \cref{prop:kronecker_erank}, the effective rank of $\mJ$ is at most 
    \begin{align}
        \erank{\mJ} \le \sum_{i=1}^L \transparentUnderbrace{\erank{\weight_{i-1:1}}}{forward} \cdot \transparentUnderbrace{\erank{\weight_{L:i+1}}}{backward},
    \end{align}
    which is much smaller than $\dim \mJ = \dim \vg = L \cdot \dim^2 \Input$.
    As a result, $\mJ$ is low-rank, making both the new-task gradient $\vg$ and the \LargeCurvature{} directions of the old Hessian $\oldhessian$ lie in a low-dimensional subspace, and facilitating their alignment. 

    Note that since the sub-matrix $\weight_{i-1:1} \kronecker \weight_{L:i+1}^\top$ of $\mJ$, which is responsible for the alignment involving $\mG_i$ or $\weight_i$, does not involve $\weight_i$ itself but \emph{other layers} $\weight_{L:i+1}$ and $\weight_{i-1:1}$. As a result, the alignment requires at least 2 layers, otherwise $\mJ$ would be full rank and \AdversarialAlignment{} would not happen.
    It explains the phase transition between \cref{figure:phase_transition_1} and \cref{figure:phase_transition_2}.
    It also explains why depth intensifies the \AdversarialAlignment{}: by our assumption that the old task is well interpolated, one must have $\weight_{L:1} = \OldGroundTruth$ that is low-rank and by $\ltwo$ regularization's implicit bias, we observe the low-rankness of $\OldGroundTruth$ is evenly distributed among the weights at all layers in the sense of $\SingularMatrix{\weight_i} = \SingularMatrix{\OldGroundTruth}^{1/L}$. As a result, when depth $L$ increases, the current layer will be attributed with less low-rankness, leaving more low-rankness for \emph{other layers} as a whole. As a result, $\weight_{i-1:1} \kronecker \weight_{L:i+1}$ and $\mJ$ will be lower-rank when depth $L$ increases, leading to more \AdversarialAlignment{}.
\else
    Although we find that low-rankness and depth lead to the \AdversarialAlignment{}, but why it happens is still not clear.
    To explicitly understand it, we revisit the definition (\ref{eq:alignment_definition}) of alignment and analyze a key step of computing its numerator, which is the product of the new-task gradients $\vg$ and the old-task Hessian $\oldhessian$:
    \begin{align}
        &\ex{\vg^\top \oldhessian \vg}
        =\exx \Fnorm{ \sum_{i=1}^L \transparentUnderbrace{\weight_{L:i+1}}{old task backward} \mG_i \transparentUnderbrace{\weight_{i-1:1} \OldInputs}{old task forward}}^2,
        \label{eq:numerator}
    \end{align}
    where $\mG_i \defeq \partialfrac{\NewLoss}{\weight_i}$ is the (matrix-shaped) new-task gradient \wrt{} the $i$-th layer's weight, $\vg \defeq \begin{bmatrix}
        \vectorize{\mG_i}
    \end{bmatrix}_i$ is the flattened new-task gradient vector, $\weight_i$ is the $i$-th weight immediately after the training of old task. 
    $\weight_{b:a} \defeq \weight_{b} \weight_{b-1} \cdots \weight_{a}$ denotes a consecutive product of weight matrices, which come from the forward and backward propagation of the old-task training, respectively.
    We argue that the two weight products (\ie{} $\weight_{L:i+1}$ and $\weight_{i-1:1}$) are the key to the alignment since they can help reveal how the old-task weights control the \LargeCurvature{} directions of the old-task Hessian as well as the condition to align with them.

    To dig out the condition in \cref{eq:numerator}, we first transform the \cref{eq:numerator} with our finding that under the commonly applied $\ltwo$ regularization, the DLN weights at all layers become low-rank with the same low-rank singular values $\mat{\Sigma} = \begin{bmatrix}
            \Signal & \mat{0} \\
            \mat{0} & \mat{\Sigma}_{\text{nuisance}}
    \end{bmatrix}$ with $\min{\Signal} \gg \max{\mat{\Sigma}_{\text{nuisance}}}$ and same ``adjacent'' singular vectors (see \cref{lemma:auto_alignment_implication_singular_values}). 
    As a result, when the network is deep, for middle layer $i$, the two consecutive weight products become low-rank in the form of $\weight_{L:i+1} = \SimplifiedConsecutiveProduct{L}{i-1}[L-i], \weight_{i-1:1} = \SimplifiedConsecutiveProduct{i-1}{1}[i-1]$, where ``$\nearzero$'' denotes matrices close to zero.
    In this way, these low-rank products only preserve the old-task critical components of features (in the forward propagation) or gradients (in the backward propagation) corresponding to $\Signal^{L-i}$ and $\Signal^{i-1}$ and suppress other components to near zero. Then, we can transform \cref{eq:numerator} into: 
    \begin{align}
        &\ex{\vg^\top \oldhessian \vg}
        = \exx \Fnorm{\sum_{i=1}^L \transparentUnderbrace{\SimplifiedConsecutiveProduct{L}{i-1}[L-i]}{old backward} \mG_i \transparentUnderbrace{\SimplifiedConsecutiveProduct{i-1}{1}[i-1] \OldInputs}{old forward}}^2.
    \end{align}
    Therefore, the condition of achieving \AdversarialAlignment{}, \ie{} to have large $\ex{\vg^\top \oldhessian \vg}$ without having too large $\ex{\norm{\vg}_2^2}$, is that $\mG_i$ must keep large spectrum in the row or column subspaces corresponding to $\Signal^{L-i}$ and $\Signal^{i-1}$ and have other components suppressed to near zero.

    The new-task gradient $\mG_i$ at the first step exactly satisfies this condition because it uses the same group of weights as the weight at the end of old-task training in the new-task forward and backward propagation:
    \begin{align}
        &\mG_i 
        = \nonTransparentUnderbrace{\weight_{L:i+1}^\top}{{new backward}} (\weight_{L:1} \NewInputs - \NewLabels) \nonTransparentUnderbrace{\NewInputs^\top \weight_{i-1:1}^\top}{new forward}\label{eq:new_update}\\
        =& \nonTransparentUnderbrace{\TransposedSimplifiedConsecutiveProduct{L}{i-1}[L-i]}{new backward} (\weight_{L:1} \NewInputs - \NewLabels) \nonTransparentUnderbrace{\NewInputs^\top \TransposedSimplifiedConsecutiveProduct{i-1}{1}[i-1] }{new forward}.
    \end{align}
    Since $\mG_i$ has low-rank $\weight_{L:i+1}$ and $\weight_{i-1:1}$ factors, it is projected onto the row and column subspaces determined by $\Signal^{L-i}$ and $\Signal^{i-1}$, while other spectrum outside them are suppressed to zero. Satisfying the condition discussed above, the new-task gradient implicitly aligns with the \LargeCurvature{} directions of the old Hessian.

    We summarize the emergence of \AdversarialAlignment{} as follows: The old-task training has low-rank bias, where old-task weights are low rank and these ranks encode the critical information (e.g., \LargeCurvature{} directions) of the old task. 
    Such information is decoded implicitly because the low-rank weights act as low-rank projections in both forward and backward propagation. 
    Such that it shapes the old-task Hessians and new-task gradients by projecting them into the same low-dimensional subspace with the same low-rank projections, overcoming the sparsity of \LargeCurvature{} directions caused by the curse of dimensionality and facilitating alignment.

    Note that since $\mG_i$ in alignment does not involve $\weight_i$ itself but \emph{other layers} $\weight_{L:i+1}$ and $\weight_{i-1:1}$ as shown in \cref{eq:new_update}, the alignment requires at least 2 layers.
    It explains the phase transition between \cref{figure:phase_transition_1} and \cref{figure:phase_transition_2}.
    It also explains why depth intensifies the \AdversarialAlignment{}: by our assumption that the old task is well interpolated, one must have $\weight_{L:1} = \OldGroundTruth$ that is low-rank and by $\ltwo$ regularization's implicit bias, we observe the low-rankness of $\OldGroundTruth$ is evenly distributed among the weights at all layers: $\SingularMatrix{\weight_i} = \SingularMatrix{\OldGroundTruth}^{1/L}$. As a result, when depth $L$ increases, the current layer will be attributed with less low-rankness, leaving more low-rankness for \emph{other layers} as a whole. As a result, the low-rank projections $\weight_{L:i+1}, \weight_{i-1:1}$ formed by \emph{other layers} will be lower-rank when depth $L$ increases, leading to more \AdversarialAlignment{}.
\fi

%% file: results/application.tex
\subsection{Mitigation of Adversarial Alignment}\label{sec:application}

\begin{figure}
    \centering
    \setstretch{0.2}
    \includegraphics[width=0.95\linewidth]{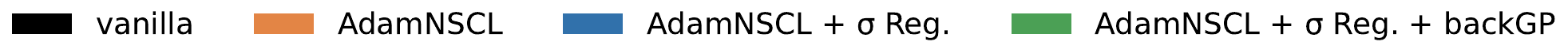}
    \begin{subfigure}{\linewidth}
        \centering
        \begin{subfigure}[t]{0.32\linewidth}
            \centering
            \labelsize
            $\alignment(\hessian_{t-1}, \param_t - \param_{t-1})$
            \includegraphics[width=\linewidth]{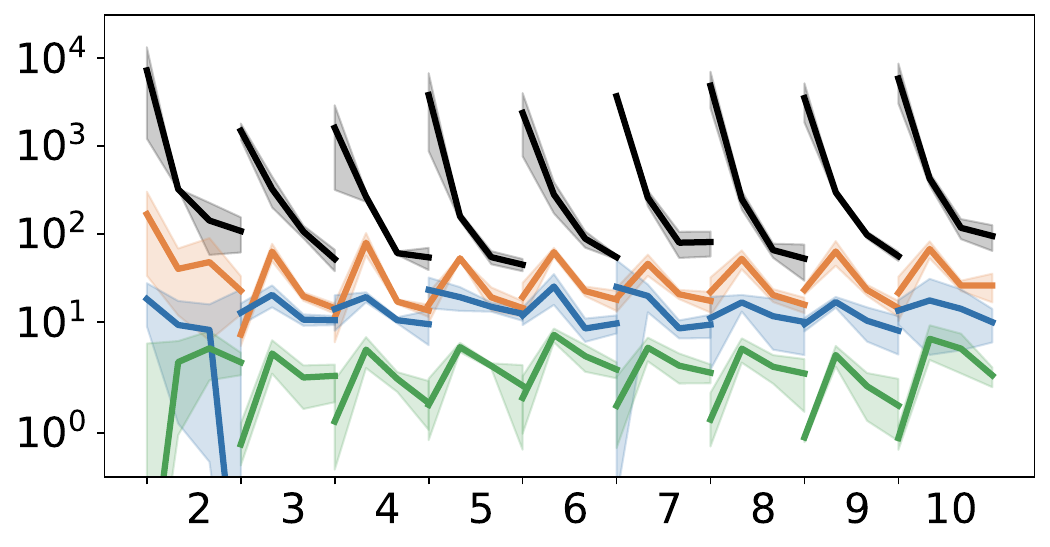}\\
            \quad Task $t$
        \end{subfigure}
        \begin{subfigure}[t]{0.32\linewidth}
            \centering
            \labelsize
            $\norm{\param_t  - \param_{t-1}}_2^2$
            \includegraphics[width=\linewidth]{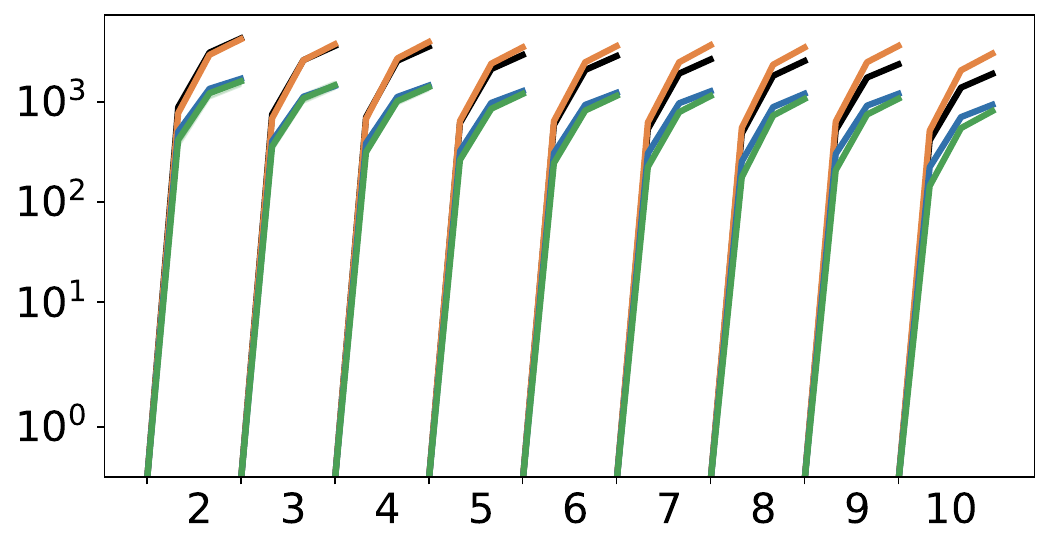}\\
            \quad Task $t$
        \end{subfigure}
        \begin{subfigure}[t]{0.32\linewidth}
            \centering
            \labelsize
            $\trace{\hessian_{t-1}}$
            \includegraphics[width=\linewidth]{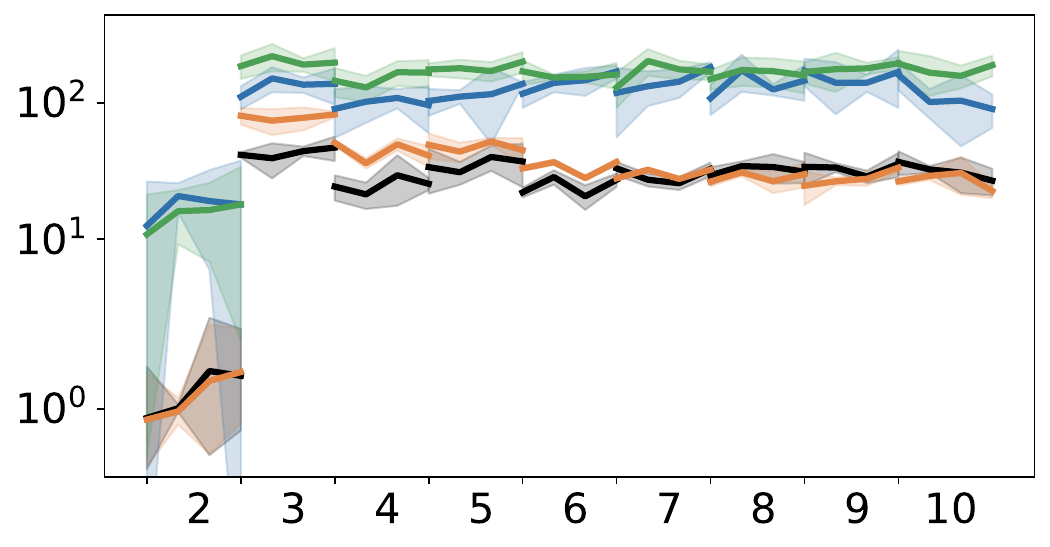}\\
            \quad Task $t$
        \end{subfigure}
        \caption{Alignments, update norms and Hessian traces of/between Task $t$ and the last task $t-1$.}
    \end{subfigure}
    \begin{subfigure}{\linewidth}
        \centering
        \begin{subfigure}[t]{0.32\linewidth}
            \centering
            \labelsize
            $\alignment(\hessian_{t-1}, \param_t - \param_{2})$
            \includegraphics[width=\linewidth]{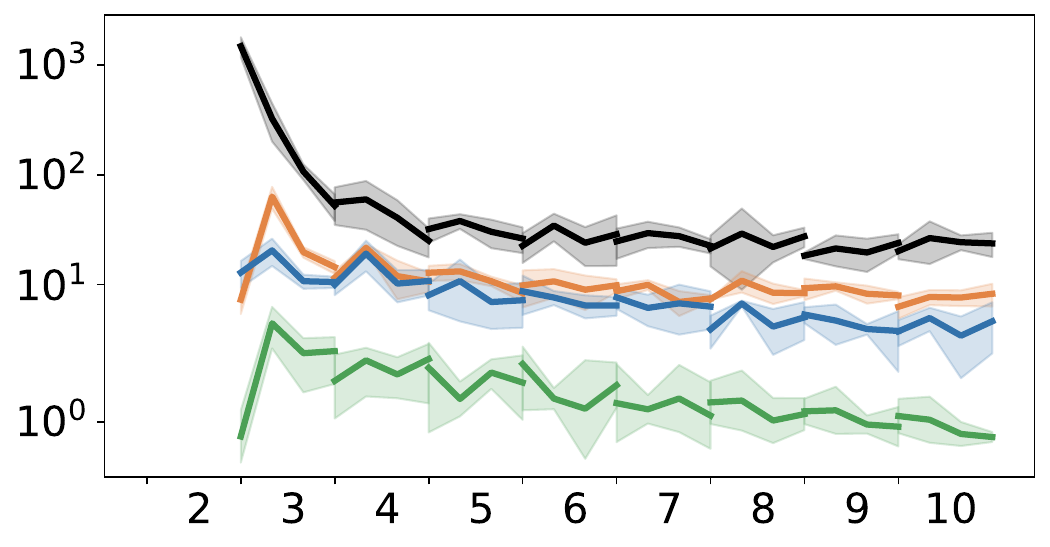}\\
            \quad Task $t$
        \end{subfigure}
        \begin{subfigure}[t]{0.32\linewidth}
            \centering
            \labelsize
            $\norm{\param_t  - \param_{2}}_2^2$
            \includegraphics[width=\linewidth]{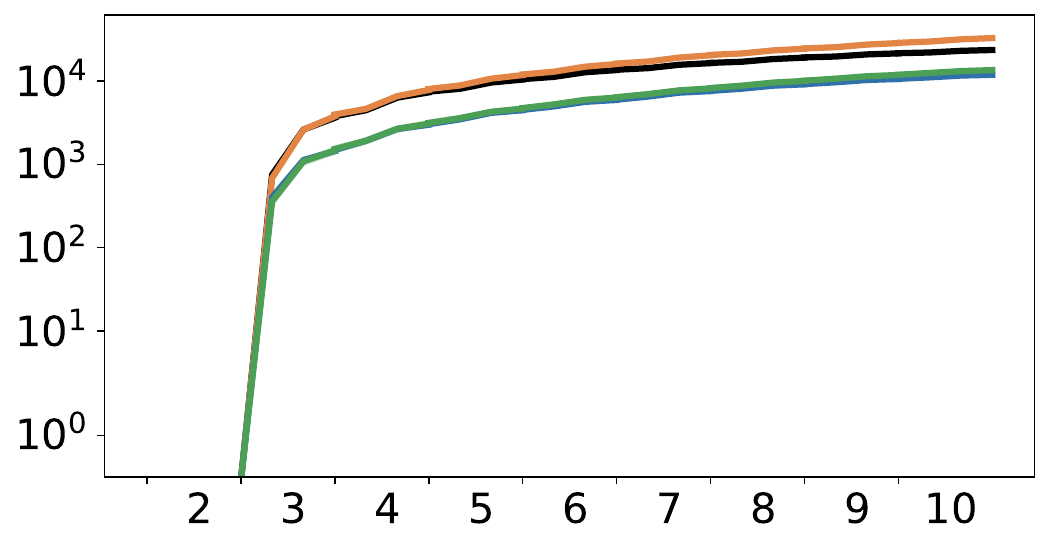}\\
            \quad Task $t$
        \end{subfigure}
        \begin{subfigure}[t]{0.32\linewidth}
            \centering
            \labelsize
            $\trace{\hessian_{2}}$
            \includegraphics[width=\linewidth]{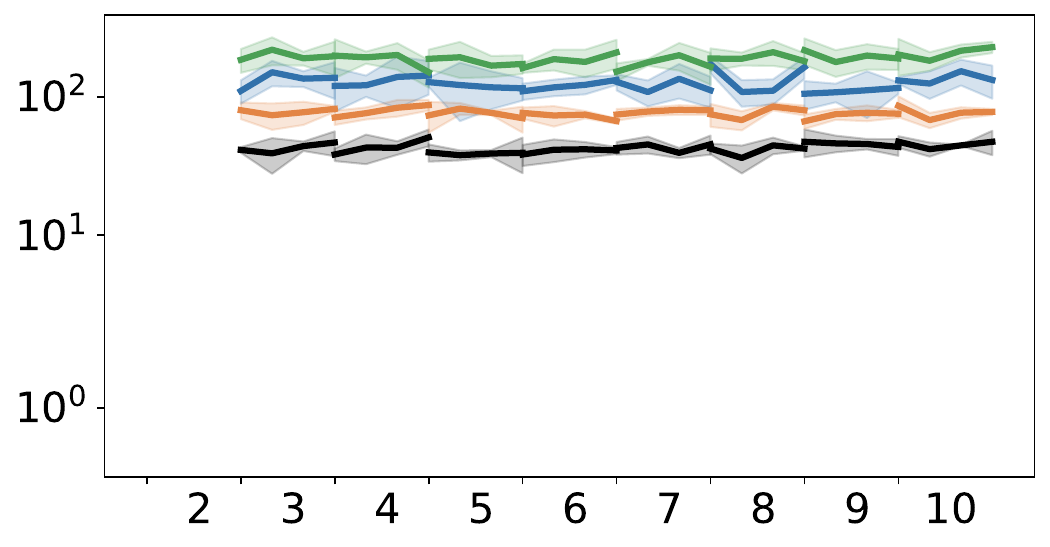}\\
            \quad Task $t$
        \end{subfigure}
        \caption{Alignments, update norms and Hessian traces of/between Task $t$ and the most severely forgotten Task $2$.}
    \end{subfigure}
    \caption{\textbf{Effectiveness of algorithms through the lens of \cref{eq:alignment_and_forgetting}.} 
        We run vanilla training (without any forgetting mitigation), AdamNSCL with various regularizers or with \backNSCL{} on 10-split CIFAR100. At each task, we compute the \AdversarialAlignment{}, weight difference, and Hessian with/of old tasks. Two kinds of old tasks are considered, \ie{} the one before the current task, and the most forgotten (with the most loss increase) task $2$. 
        Data are recorded every 4000 steps.
        Regarding \AdversarialAlignment{}, we observe (1) AdamNSCL reduces \AdversarialAlignment{} compared to the vanilla method; (2) even if AdamNSCL is used, \AdversarialAlignment{} still exists, \ie{} residual adversariality; (3) spectral regularization also reduces \AdversarialAlignment{} but leaves residual adversariality; (4) \backNSCL{} further reduces the residual \AdversarialAlignment{}.
        Regarding the other two factors, we observe (1) spectral regularization reduces all update norms as while as Hessian traces of tasks $6 \sim 10$, while forward or backward GPs do not change them drastically; (2) all tested methods affect early tasks' Hessian traces in the inverse way as the alignment, leading to a tendency to increase forgetting. 
        Through the lens of \cref{eq:alignment_and_forgetting}, we conclude that forward and our backward GPs mitigate forgetting exactly by reducing the \AdversarialAlignment{}, instead of affecting the two other factors. Results in these figures are recorded during the experiment of \cref{table:comparison}. 
        }\label{figure:ablations}
\end{figure}

We note \AdversarialAlignment{} is induced by both forward and backward propagation.
Current representative \CL{} method of gradient projection (GP) families \citep{nscl,gpm,sgp,forgetting_and_flatness2} can alleviate \AdversarialAlignment{} induced by the forward propagation, but leave residual adversariality induced by the backward propagation, as summarized in \cref{figure:main} and elaborated in \cref{sec:analysis_of_existing_methods}.
See \cref{figure:ablations} for empirical evidence, where GP methods reduce \AdversarialAlignment{} from $\alignment \sim 10^{3}$ to $\alignment \sim 10^{2}$.
To alleviate the residual adversariality, we apply GP techniques to the backward direction and propose the backward gradient projection (\backGP{}) method, as elaborated in \cref{sec:resolution_of_limitations}.

We evaluate our methods on standard \CL{} benchmarks, \ie{} CIFAR100 split into 10 or 20 tasks and TinyImageNet split into 25 tasks.
Let $T$ be the number of tasks and let $a_{t, i}$ denote the accuracy of task $i$ immediately after the training of task $t$. We evaluate the methods using (final) accuracy $\acc \defeq \frac{1}{T}\sum_{i=1}^T a_{T, i}$ for the overall performance, backward transfer $\bwt \defeq \frac{1}{T-1} \sum_{i=1}^{T-1} (a_{T, i} - a_{i, i})$ for forgetting and immediate accuracy $\immAcc \defeq \frac{1}{T} \sum_{i=1}^T a_{i, i}$ for plasticity.
We use modern backbone ConvNeXt \citep{convnext} and spectral regularization \citep{orthogonality_regularization}
\begin{align}
    \spectralReg(\weight_i) \defeq \Fnorm{\weight_i \weight_i^\top - \mI}^2
\end{align}
to boost plasticity. See \cref{sec:details_of_experiments} for the detailed discussion.

\cref{table:comparison} shows the experiment results.
Spectral regularization and modern architecture improve the plasticity by at least $10\%$ and further improve final performances.
However, in this high-plasticity regime, GP methods forget more with $\bwt \approx -10\%$, making forgetting the major problem again.
After adding our \backNSCL{}, forgetting is reduced to minimal ($\bwt \approx -1\%$). Although plasticity is partially sacrificed ($\approx -2\%$), the final accuracy is further improved by approximately $ 5\%$.
The improvement is the most drastic in the 20-split CIFAR100 setting, where the final accuracy surpasses $91\%$.
Therefore, adding \backNSCL{} is effective in alleviating the residual forgetting of GP methods and boosting their performance in high-plasticity \CL{}.

We further examine if \backNSCL{} alleviates forgetting in the same manner as it is designed. As shown in \cref{figure:ablations}, \backNSCL{} further reduces residual \AdversarialAlignment{}. At the same time, new-task update norms and old-task Hessian traces remain the same or increase, confirming that forgetting is alleviated exactly through reducing adversariality. From both \cref{table:comparison,figure:ablations}, we note spectral regularization also helps alleviate forgetting, possibly by pushing weights toward the identity, reducing low-rankness and making Jacobians less adversarial.

\begin{sidewaystable}
    \newcommand{\bypercent}{(\%)}
    \newcommand{\nostd}{\hphantom{$\pm 0.00$}}
    \centering
    \scalebox{0.9}{
        \centering
        \begin{tabular}{lccccccccc}
            \toprule
            & \multicolumn{3}{c}{10-Split CIFAR100} & \multicolumn{3}{c}{20-Split CIFAR100} & \multicolumn{3}{c}{25-Split TinyImageNet}\\
            \cmidrule(lr){2-4}
            \cmidrule(lr){5-7}
            \cmidrule(lr){8-10}
            & $\acc$ \bypercent & $\bwt$ \bypercent  & $\immAcc$\bypercent 
            & $\acc$ \bypercent & $\bwt$ \bypercent  & $\immAcc$\bypercent 
            & $\acc$ \bypercent & $\bwt$ \bypercent  & $\immAcc$\bypercent \\
            \midrule
            EWC$^\ddagger$ & $70.77$ \hphantom{$\pm 0.00$}  & $-2.83$ \hphantom{$\pm 0.00$}  & $73.32$ \hphantom{$\pm 0.00$}  & $71.66$ \hphantom{$\pm 0.00$}  & $-3.72$ \hphantom{$\pm 0.00$}  & $75.19$ \hphantom{$\pm 0.00$}  & $52.33$ \hphantom{$\pm 0.00$}  & $-6.71$ \hphantom{$\pm 0.00$}  & $58.77$ \hphantom{$\pm 0.00$}  \\
            MAS$^\ddagger$ & $66.93$ \hphantom{$\pm 0.00$}  & $-4.03$ \hphantom{$\pm 0.00$}  & $70.56$ \hphantom{$\pm 0.00$}  & $63.84$ \hphantom{$\pm 0.00$}  & $-6.29$ \hphantom{$\pm 0.00$}  & $69.82$ \hphantom{$\pm 0.00$}  & $47.96$ \hphantom{$\pm 0.00$}  & $-7.04$ \hphantom{$\pm 0.00$}  & $54.72$ \hphantom{$\pm 0.00$}  \\
            SI$^\ddagger$ & $60.57$ \hphantom{$\pm 0.00$}  & $-5.17$ \hphantom{$\pm 0.00$}  & $65.22$ \hphantom{$\pm 0.00$}  & $59.76$ \hphantom{$\pm 0.00$}  & $-8.62$ \hphantom{$\pm 0.00$}  & $67.95$ \hphantom{$\pm 0.00$}  & $45.27$ \hphantom{$\pm 0.00$}  & $-4.45$ \hphantom{$\pm 0.00$}  & $49.54$ \hphantom{$\pm 0.00$}  \\
            LwF$^\ddagger$ & $70.70$ \hphantom{$\pm 0.00$}  & $-6.27$ \hphantom{$\pm 0.00$}  & $76.34$ \hphantom{$\pm 0.00$}  & $74.38$ \hphantom{$\pm 0.00$}  & $-9.11$ \hphantom{$\pm 0.00$}  & $83.03$ \hphantom{$\pm 0.00$}  & $56.57$ \hphantom{$\pm 0.00$}  & $-11.19$ \hphantom{$\pm 0.00$}  & $67.31$ \hphantom{$\pm 0.00$}  \\
            MEGA$^\ddagger$ & $54.17$ \hphantom{$\pm 0.00$}  & $-2.19$ \hphantom{$\pm 0.00$}  & $56.14$ \hphantom{$\pm 0.00$}  & $64.98$ \hphantom{$\pm 0.00$}  & $-5.13$ \hphantom{$\pm 0.00$}  & $69.85$ \hphantom{$\pm 0.00$}  & $57.12$ \hphantom{$\pm 0.00$}  & $-5.90$ \hphantom{$\pm 0.00$}  & $62.78$ \hphantom{$\pm 0.00$}  \\
            A-GEM$^\ddagger$ & $49.57$ \hphantom{$\pm 0.00$}  & $-1.13$ \hphantom{$\pm 0.00$}  & $50.59$ \hphantom{$\pm 0.00$}  & $61.91$ \hphantom{$\pm 0.00$}  & $-6.88$ \hphantom{$\pm 0.00$}  & $68.45$ \hphantom{$\pm 0.00$}  & $53.32$ \hphantom{$\pm 0.00$}  & $-7.68$ \hphantom{$\pm 0.00$}  & $60.69$ \hphantom{$\pm 0.00$}  \\
            \midrule
            GAGP* \citep{gagp}                      &  $74.77$ \nostd                   & $-1.00$ \nostd                & $75.79$ \nostd             &  $80.82$ \nostd                   & $-2.19$ \nostd               & $82.90$ \nostd                     &  \nocomp{$64.17$ \nostd}                & \nocomp{$-0.41$ \nostd}                & \nocomp{$64.56$ \nostd} \\
            AdaBOP* \citep{non_GP_method_results}   &  $76.56$ \nostd                   & $-2.56$ \nostd                & $78.86$ \nostd             &  $79.65$ \nostd                   & $-3.95$ \nostd               & $83.40$ \nostd                     &  $63.01$ \nostd                         & $-6.84$ \nostd                         & $69.58$ \nostd \\ 
            TRGP* \citep{trgp}                      &  $74.46$ \nostd                   & $-0.90$ \nostd                & $75.27$ \nostd             &                                   &                              &                                    &  \nocomp{$61.78$ \nostd}                & \nocomp{$-0.5$  \nostd}                & \nocomp{$62.26$ \nostd} \\
            ROGO* \citep{rogo}                      &  $74.04 \pm 0.35$                 &                               &                            &                                   &                              &                                    &  \nocomp{$63.66 \pm 1.24$}              &                               &                \\
            \multicolumn{2}{l}{DF* \citep{forgetting_and_flatness2}}\\
            \,\, + TRGP*                            &  $76.15 \pm 0.22$                 & $\best{-0.00} \pm 0.00$       & $76.15$ \nostd             &                                   &                              &                                    &  \nocomp{$70.76 \pm 1.09$}              & \nocomp{$-0.00 \pm 0.00$}              & \nocomp{$70.76$ \nostd} \\
            \,\, + ROGO*                            &  $74.68 \pm 0.34$                 & $-0.01 \pm 0.00$              & $74.69$ \nostd             &                                   &                              &                                    &  \nocomp{$71.07 \pm 1.69$}              & \nocomp{$-0.01 \pm 0.00$}              & \nocomp{$71.08$ \nostd} \\
            \,\, + SGP*                             &  $76.50 \pm 0.61$                 & $\best{-0.00} \pm 0.00$       & $76.50$ \nostd             &                                   &                              &                                    &  \nocomp{$71.22 \pm 1.40$}              & \nocomp{$-0.00 \pm 0.00$}              & \nocomp{$71.22$ \nostd} \\ 
            SD* \citep{sd}\\
            \,\, + NSCL*                                      &  $75.97 \pm 0.66$                 & $-2.88 \pm 0.89$              & $78.56$ \nostd             & $76.50 \pm 1.02$                  & $-3.99 \pm 0.96$             & $80.29$ \nostd                     &  $60.38 \pm 0.75$                       & $-4.81 \pm 1.00$                       & $65.00$ \nostd \\
            \,\, + TRGP*                                      &  $75.50 \pm 0.35$                 & $-0.96 \pm 0.09$              & $76.36$ \nostd             & $83.84 \pm 0.12$                  & $-0.72 \pm 0.20$             & $84.52$ \nostd                     &  $65.80 \pm 0.16$                       & $-0.49 \pm 0.08$                       & $66.27$ \nostd \\
            GPCNS* \citep{gpcns}                    &  $74.40 \pm 0.42$                 & $-2.16 \pm 0.92$              & $76.34$ \nostd             &                                   &                              &                                    &  \nocomp{$63.78 \pm 0.62$}              & \nocomp{$-2.84 \pm 1.15$}              & \nocomp{$66.48$ \nostd} \\
            \,\, + TRGP*                            &  $75.58 \pm 0.36$                 & $-0.06 \pm 0.33$              & $75.63$ \nostd             &                                   &                              &                                    &  \nocomp{$66.07 \pm 0.47$}              & \nocomp{$+0.03 \pm 0.29$}              & \nocomp{$66.04$ \nostd} \\
            \,\, + SGP*                             &  $76.25 \pm 0.38$                 & $-0.13 \pm 0.05$              & $76.37$ \nostd             &                                   &                              &                                    &  \nocomp{$63.98 \pm 0.53$}              & \nocomp{$-0.81 \pm 0.31$}              & \nocomp{$64.75$ \nostd} \\
            \midrule
            NSCL* \citep{nscl}                      &  $73.77$ \hphantom{$\pm 0.00$}    & $-1.60$ \hphantom{$\pm 0.00$} & $75.21$ \hphantom{$\pm 0.00$} &  $75.95$ \hphantom{$\pm 0.00$}    & $-3.66$ \hphantom{$\pm 0.00$} & $79.43$ \hphantom{$\pm 0.00$}     &  $58.28$ \hphantom{$\pm 0.00$}    & $-6.05$ \hphantom{$\pm 0.00$} & $64.09$ \hphantom{$\pm 0.00$}         \\
            NSCL                                    &   $77.95 \pm 0.62$                & $-9.16 \pm 0.56$              & $86.19 \pm 0.28$              & $76.03 \pm 0.70$                  & $-13.95 \pm 0.68$             & $89.28 \pm 0.78$                  & $52.67 \pm 1.63$                  & $-16.73 \pm 1.10$             & $68.73 \pm 0.72$                      \\
            \,\, $+\spectralReg$                   &   $81.96 \pm 0.79$                & $-8.44 \pm 0.84$              & $\localBest{\best{89.55}} \pm 0.31$       & $82.93 \pm 0.69$                  & $-10.71 \pm 0.87$             & $\localBest{\best{93.10}} \pm 0.25$           & $68.75 \pm 0.72$                  & $-9.10 \pm 1.00$              & $\localBest{77.49} \pm 0.42$          \\
            \,\, $+\spectralReg+\bGP{}$     &   $\localBest{86.61} \pm 0.40$    & $\localBest{-1.43} \pm 0.45$  & $87.89 \pm 0.12$              & $\localBest{90.30} \pm 0.50$      & $\localBest{-0.93} \pm 0.31$  & $91.18 \pm 0.36$                  & $\localBest{73.23} \pm 0.34$      & $\localBest{-0.94} \pm 0.34$  & $74.13 \pm 0.43$                      \\
            \midrule
            GPM* \citep{gpm}               &   \hphantom{$\pm 0.00$}           &  \hphantom{$\pm 0.00$}        &  \hphantom{$\pm 0.00$}        &  $72.48$ \hphantom{$\pm 0.00$}    & $\best{-0.00}$ \hphantom{$\pm 0.00$} & $72.48$ \hphantom{$\pm 0.00$}     &  \nocomp{$60.41$ \hphantom{$\pm 0.00$}}    & \nocomp{$-0.00$ \hphantom{$\pm 0.00$}} & \nocomp{$60.41$ \hphantom{$\pm 0.00$}}         \\
            GPM                                     &   $76.79 \pm 0.65$                & $-9.98 \pm 0.75$              & $86.77 \pm 0.31$              & $69.95 \pm 0.93$                  & $-18.14 \pm 1.18$             & $88.08 \pm 0.52$                  & $59.74 \pm 1.11$                  & $-13.73 \pm 1.06$             & $73.46 \pm 0.38$                      \\
            \,\, $+\spectralReg$                   &   $80.19 \pm 0.89$                & $-8.97 \pm 1.07$              & $\localBest{89.16} \pm 0.39$  & $77.56 \pm 2.45$                  & $-14.89 \pm 2.67$             & $\localBest{92.45} \pm 0.29$      & $63.49 \pm 0.86$                  & $-13.55 \pm 0.83$             & $\localBest{77.04} \pm 0.32$          \\
            \,\, $+\spectralReg+\bGP{}$     &   $\localBest{86.32} \pm 0.30$    & $\localBest{-0.85} \pm 0.24$       & $87.17 \pm 0.16$              & $\localBest{89.75} \pm 0.24$      & $-0.35 \pm 0.13$       & $90.10 \pm 0.28$                  & $\localBest{72.70} \pm 0.23$      & $\best{-0.06} \pm 0.18$       & $72.76 \pm 0.27$                      \\
            \midrule
            SGP* \citep{sgp}               &  $76.05$ \hphantom{$\pm 0.00$} & $\localBest{}{-0.01}$ \hphantom{$\pm 0.00$} & $76.06$ \hphantom{$\pm 0.00$} &   \hphantom{$\pm 0.00$}           &  \hphantom{$\pm 0.00$}        &  \hphantom{$\pm 0.00$}            &  \nocomp{$62.83$ \hphantom{$\pm 0.00$}}    & \nocomp{$-0.01$ \hphantom{$\pm 0.00$}} & \nocomp{$62.84$ \hphantom{$\pm 0.00$}}         \\
            SGP                                     &   $81.28 \pm 0.57$                & $-4.81 \pm 0.34$              & $86.09 \pm 0.29$              & $77.88 \pm 1.24$                  & $-10.50 \pm 1.25$             & $88.39 \pm 0.72$                  & $66.55 \pm 0.59$                  & $-8.16 \pm 0.60$              & $74.71 \pm 0.26$                      \\
            \,\, $+\spectralReg$                   &   $84.43 \pm 0.33$                & $-5.07 \pm 0.43$              & $\localBest{89.50} \pm 0.20$  & $86.64 \pm 0.70$                  & $-5.86 \pm 0.72$              & $\localBest{92.50} \pm 0.37$      & $73.87 \pm 0.56$                  & $-4.47 \pm 0.48$              & $\localBest{\best{78.33}} \pm 0.10$               \\
            \,\, $+\spectralReg+\bGP{}$     &   $\localBest{\best{87.37}} \pm 0.21$         & $-1.06 \pm 0.32$  & $88.43 \pm 0.24$              & $\localBest{\best{91.35}} \pm 0.28$           & $\localBest{-1.22} \pm 0.24$  & $92.57 \pm 0.29$                  & $\localBest{\best{75.72}} \pm 0.42$           & $\localBest{-1.31} \pm 0.30$       & $77.02 \pm 0.39$                      \\
            \bottomrule
        \end{tabular} 
    }
    \caption{\textbf{Evaluation and Comparison.} 
        We report results averaged over 5 runs.
        For each metric, we mark the best among all methods in bold and the locally best result under the same forward GP method in wave underline.
        The asterisk (*) indicates the performance is copied from original papers, where AlexNets or ResNets are used as the backbone.
        The double dagger ($\ddagger$) indicates the result is copied from \citet{non_GP_method_results}, where ResNet18 is used as the backbone. 
        Experiments without the above marks are conducted by ourselves using ConvNeXt-Tiny as the backbone.
        The $\immAcc$ metric is not reported in some papers, which we compute ourselves by $\immAcc = \acc - \frac{T-1}{T} \bwt$.
        Gray results use MiniImageNet (100 classes, 20 Splits) instead of TinyImageNet (200 classes, 25 Splits) in the original paper. When marking the best results, we exclude these MiniImageNet results.
    }\label{table:comparison}
\end{sidewaystable}

%% file: conclusion.tex
\section{Discussion}

Catastrophic forgetting is a long-standing challenge in continual learning, whose theoretical understanding is still limited or restricted to shallow networks.
We identify the adversarial nature of catastrophic forgetting of deep networks. We first confirm the existence of \AdversarialAlignment{} phenomenon in deep continual learning, \ie{} the new task updates have large projections onto the \LargeCurvature{} directions of the old task, even when the tasks have different loss landscapes and the old-task \LargeCurvature{} directions are sparse.
The \AdversarialAlignment{} amplifies the forgetting thousands of times by accurately attacking the most fragile part of the model's memory on the old task.
We identify non-data but algorithmic inductive bias as a key factor in the emergence of the \AdversarialAlignment{}. 
Particularly, the low-rank structure of old-task weights encodes the information about the old-task \LargeCurvature{} directions and passes it to the new task. During forward and backward propagation, these weights form low-rank Jacobians and act as low-rank projections pulling the new-task gradient and the old-task \LargeCurvature{} directions to the same low-dimensional subspace, producing \AdversarialAlignment{}.
Depth intensifies the low-rankness in the projections and increases the adversariality, leading to a phase transition of alignment in deep networks that is not covered by previous studies on shallow networks.
We connect gradient projection methods to \AdversarialAlignment{} alleviation, identify and mitigate their residual adversariality induced by the backward direction. The resulting \backNSCL{} alleviates forgetting and boosts continual learning performance by a large margin.

We list limitations of our work: 
(1) Our theoretical analysis assumes new-task data is randomly generated from the old task.  
(2) We only theoretically study deep linear networks for technical tractability. How adversarial \AlignmentPhenomenon{} arises in non-linear networks remains an open question. We conjecture there are at least two differences: the sparsity of non-linear neuron activation \citep{activation_sparsity_1,activation_sparsity_2} may create more low-rankness, but the non-linear activation also makes the Jacobians input-dependent and may hinder the low-rankness from being passed to new tasks.
(3) Our theoretical result only addresses the first step of new-task training for technical tractability. We conjecture the later dynamics involve at least two trends: (a) the new-task training that learns new features, increases the ranks of weights and Jacobians, and finally reduces the adversariality (\cref{figure:alignment_changes_cross_modal,figure:ablations}), and (b) an implicit power iteration of the old-task Hessian happens, akin to the generation of adversarial samples \citep{implicit_power_iteration}, which strengthens the alignment in initial steps (\cref{figure:alignment_changes_cifar100,figure:alignment_changes_randrot_mnist,figure:ablations}). We elaborate on the implicit power iteration in \cref{appendix:power_iteration} in the \supplementary{}. 
(4) We only study \AdversarialAlignment{} in the second order, whereas higher-order terms also contribute to forgetting (compare the second-order and the actual forgettings in \cref{figure:all_forgettings}). We conjecture that the low-rank bias will similarly induce low-rank Jacobians and pull the sharp directions of old-task higher-order derivatives and the new-task gradient to the same subspace.

For future works, we suggest that recognizing the adversarial nature of catastrophic forgetting opens opportunities to transfer insights on adversarial attack/robustness to continual learning, \eg{} transferring the implicit power iteration underlying adversarial sample generation to \CL{}. Beyond the scope of \CL{}, our result is an example of how training on one dataset shapes the learning on other datasets. This conjectures a preliminary model on how pretraining helps downstream-task finetuning, \ie{} increasing expressiveness (gradient norm) along the few directions important (\LargeCurvature{}) to the pretraining task and decreasing expressiveness along other directions. Since they may help us understand why finetuning can be done by only updating a small portion of parameters and why beyond-pretraining finetuning is inefficient, we believe it is worth future investigation.

%% file: methods/methods.tex
\section{Methods} \label{sec:methods}

\input{methods/verification.tex}

\input{methods/theory.tex}

\input{methods/algorithm.tex}

%% file: methods/verification.tex
\subsection{Verifying \CapAdversarialAlignment}\label{sec:verification}

We first argue in more detail the necessity of directly verifying the existence of \AdversarialAlignment{}, as a complement to the discussion in the Introduction.
That is, we discuss how much and how sufficient the existing evidence is regarding its existence, and whether we need more evidence. 
Prior works \citep{meat_learning_as_regularization,hessian_perspective_of_regularization_CL} have shown that a wide range of \CL{} algorithms, which are effective in alleviating forgetting, explicitly or implicitly prevent the alignment (when the alignment exists).
We acknowledge that these works have, at least, suggested that \AdversarialAlignment{} should exist, so that \CL{} algorithms can effectively alleviate forgetting, which is consistent with empirical observations.
However, the argument is indirect and is not conclusive because, strictly, the fact that \CL{} algorithms can suppress the alignment and alleviate forgetting does not imply that \CL{} algorithms indeed suppress the alignment or that the alignment exists. Particularly, it is possible for the following three conditions to hold simultaneously:   (1) the alignment does not exist, (2) \CL{} algorithms suppress alignment, and (3) \CL{} algorithms alleviate forgetting. For example, the alignment is (near) ``zero'' and \CL{} algorithms suppress it from (near) zero to (near) zero, and \CL{} algorithms alleviate forgetting through some unknown mechanisms (\eg{} reducing higher-order forgetting, or rearranging parameters beyond the Taylor expansion's convergence radius). 
In this case, forgetting is not related to alignment and the alleviation of forgetting cannot be explained by the suppression of alignment, making both the previously discovered influence and mitigation of alignment meaningless.
Furthermore, our preliminary analysis in the Introduction also suggests that the alignment should not exist. 
Confronted with evidence that is indirect and preliminary analysis that suggests the opposite, the existence of the alignment is suspectable and we must directly verify the existence of \AdversarialAlignment{}.

In later subsections, we list the details of our verification experiments and define the quantitative measure of alignment.

\subsubsection{Experimental Settings}

To verify the existence of \AdversarialAlignment{}, we conduct \CL{} experiments over a variety of tasks and architectures and test whether the new task update indeed has a large projection onto the \LargeCurvature{} directions of the old task. Here we list important aspects of the experiments.  See \cref{table:verification_hyperparameters} for more details, like hyperparameters.
The first \CL{} experiment (\cref{figure:existence_cifar100}) is 10-split CIFAR100 that is standard in \CL{} literature, where a ResNet18 \citep{resnet}, a VisionTransformer-Small \citep{vit}, and a MLP-Mixer-Small \citep{mixer} are trained. The models are only trained on the first 2 tasks, referred to as the old and the new tasks, respectively. The old-task Hessian is computed immediately after the old task training, and the new-task update is computed at the first step of the new-task training in the CDF diagrams and at the first 80 steps of the new-task training in the box diagrams of \cref{figure:existence}.
The second \CL{} experiment is a visual-lingual multi-modal one (\cref{figure:existence_cross_modal}), where the old task is the first split of 10-split CIFAR100 and the new task is the entire SST2 dataset of language sentiment analysis. Since it is harder to adapt a pixel-level convolution network to text, we only use patch/token-based models like VisionTransformer and MLP-Mixer. When trained on the old visual task, the image is cut into patches and embedded by a trainable linear projection. When trained on the new language task, we tokenize and embed the sentences by the pretrained (frozen) tokenizer and embedding of \texttt{LLAMA-2.1} \citep{llama}. The old Hessian and the new update are computed in the same way as the previous experiment. 
The third experiment (\cref{figure:existence_randrot_mnist}) is a synthetic one, where the old task is the entire MNIST dataset after whitening and the new task is constructed by (1) randomly sample a $784 \times 784$ orthogonal matrix $\mat{U}$ that is uniformly distributed in the Haar sense, (2) flatten every $1 \times 28 \times 28$ whitened MNIST image into a $784$-dimensional vector and stacking them as a matrix $\NewInputs^{(0)}$, and (3) compute $\NewInputs \defeq \mat{U} \NewInputs^{(0)}$, while keeping the labels unchanged. 
Deep linear networks of different depths are trained in this experiment.
The whitening is done by the following steps: (1) flattening all $1 \times 28 \times 28$ MNIST images into $784$-dimensional vectors and stacking them as a matrix $\OldInputs^{(0)} \in \reals^{784 \times n_{1}}$, (2) adding element-wise Gaussian noise $\mat{\Xi}$ of standard deviation $0.01$ to make the noised sample $\tilde{\OldInputs}^{(0)} \defeq \OldInputs^{(0)} + \mat{\Xi}$ full-rank, and (3) computing $\OldInputs \defeq \left(\left(\tilde{\OldInputs}^{(0)} \left(\tilde{\OldInputs}^{(0)}\right)^\top\right)^{-1}\right)^{1/2} \tilde{\OldInputs}^{(0)}$ so that $\OldInputs \OldInputs^\top = \mI$. 
No pretraining is used. We replace a randomly initialized classifier before the training of each task.

\newcommand{\normalizedUpdate}{\bar{\vec{g}}}

The experiments involve Hessian eigenvalue and eigenvectors, whose computation is expensive for deep networks and requires approximated numerical methods.
Our goal is \emph{plotting} $p_i \defeq \frac{\inner{\HessianEigVec_i}{\vec{g}}^2}{\norm{\vec{g}}_2^2}$.
Lanczos algorithm has been used to directly compute the \emph{plot} of spectral densities in \texttt{PyHessian} \citep{pyhessian}, and we intend to develop a variation of it for our use.
Specifically, we want to compute the CDF of the density
\begin{align}
    \psi(t) \defeq \sum_{i} p_i \cdot \delta(t - \lambda_i) = \sum_{i} \inner{\HessianEigVec_i}{\normalizedUpdate}^2 \cdot \delta(t-\lambda_i),
\end{align}
where $\normalizedUpdate \defeq \frac{\vec{g}}{\norm{\vec{g}}_2}$ is the normalized new task update.
The density is composed of Dirac delta functions, which are relaxed to small-variance Gaussians:
\begin{align}
    \psi_\sigma(t) \defeq \sum_{i} \inner{\HessianEigVec_i}{\normalizedUpdate}^2 f(\lambda_i; t, \sigma) \defeq \sum_{i} \inner{\HessianEigVec_i}{\normalizedUpdate}^2 \cdot \frac{1}{\sigma \sqrt{2\pi}} e^{-(t - \lambda_i)^2 / (2 \sigma^2)}
\end{align}
Then our goal becomes the subgoal $\phi_\sigma^{\vec{v}}$ of \citet{pyhessian} with the Rademacher random vector $\vec{v}$ replaced by $\normalizedUpdate$. Since the use of the Lanczos algorithm to compute $\phi_\sigma^{\vec{v}}$ does not rely on specific properties of $\vec{v}$, we reuse the subsequent steps from \citet{pyhessian}.
The implementation is based on \texttt{PyHessian}'s spectral density function, where the Rademacher random vector is replaced with the new update.

\subsubsection{Definition of \CapAdversarialAlignment{}}\label{sec:definition}

Here, we derive the definition of \AdversarialAlignment{}.
Intuitively, \AdversarialAlignment{} means that the new task update has large projections onto the eigenvectors whose eigenvalues are also large.
Assume PSD matrix $\mat{A}$ and random vector $\vec{r}$ are involved in the \AlignmentPhenomenon{}.
The extreme case is that the new task has all its projection onto the directions with the largest eigenvalues, the extent to which can be captured by the inner product between the eigenvalues and the projection distribution
\begin{align}
    \inner{\vec{\lambda}}{\vec{p}} = \frac{\ex{\vec{r}^\top \mat{A} \vec{r}}}{\mathbb{E} \norm{\vec{r}}_2^2},
\end{align}
where $\vec{\lambda} \defeq \begin{bmatrix}
    \lambda_i
\end{bmatrix}_i$ is the vector of eigenvalues, and $\vec{p} \defeq \begin{bmatrix}
    \frac{\ex{\inner{\HessianEigVec_i}{\vec{r}}^2}}{\ex{\norm{\vec{r}}_2^2}}
\end{bmatrix}_i$ is the projection distribution.
However, this value is not normalized and is not invariant to the scale of $\mat{A}$.
Moreover, we have no intuition on what scale of this value means large alignment.
Therefore, we compare it with a baseline formed by a Gaussian random vector $\vecperturbation$ with the same expected squared norm, \ie{}
\begin{align}
    \frac{\ex[\vecperturbation \sim \mathcal{N}(\vec{0}, \mathbb{E} \norm{\vec{r}}_2^2 \cdot \mat{I} / \dim \vec{r})]{\vecperturbation^\top \mat{A} \vecperturbation}}{\ex{\norm{\vecperturbation}_2^2}} = \frac{1}{\dim \vec{r}} \cdot \trace{\mat{A}}.
\end{align}
Their fraction indicates how much the actual updates align with \LargeCurvature s better than a random perturbation does, leading to the definition of \AdversarialAlignment{}:
\begin{definition}
    Given a PSD matrix $\mat{A}$ and a random vector $\vec{r}$, their \AdversarialAlignment{} is defined as
    \begin{align}
        \alpha(\mat{A}, \vec{r}) \defeq \frac{
            \ex{\vec{r}^\top \mat{A} \vec{r}}
        }{
            \ex[\vecperturbation \sim \mathcal{N}(\vec{0}, \mathbb{E}\norm{\vecperturbation}_2^2 \cdot \mat{I} / \dim \vec{r})]{\vecperturbation^\top \mat{A} \vecperturbation}
        } 
        = \dim \vec{r} \cdot \frac{\ex{\vec{r}^\top \mat{A} \vec{r}}}{\trace{\mat{A}} \cdot \mathbb{E} \norm{\vec{r}}_2^2}.
    \end{align}
\end{definition}

As a result, $\alignment = 1$ means no alignment, while $\alignment \gg 1$ means strong alignment.
\Cref{eq:alignment_and_forgetting} also anchors the scale and meaning of $\alignment$ in a consistent manner, where $\alignment = 1$ means no amplification of forgetting, and $\alignment \gg 1$ means strong amplification of forgetting.

%% file: methods/theory.tex
\subsection{Definition of Effective Rank}\label{sec:definition_erank}

Since we intend to build quantitative connections between \AdversarialAlignment{} and low-rankness, we need to quantify the rank of weights and data.
The standard rank is not suitable because practical data often contain small but non-zero singular values, and the standard hard rank considers all of them full-rank.
Instead, we want to quantify such matrices as low-rank because the small singular values do not contribute much compared to the large ones and can be ignored.
Existing alternatives consider (normalized) singular values as a distribution and consider rank as the spread of the distribution, which smoothly ignores small singular values. As a result, they use the exponential of Shannon entropy on the spectral distribution as a soft-rank \citep{effective_rank_1,effective_rank_2}.
However, Shannon entropy involves a logarithm within expectation, which is unfriendly to matrix multiplication. We instead select Rényi entropy with $\alpha=2$, which is defined as
\begin{align}
    H_{\alpha} = \frac{1}{1-\alpha} \log \left(\sum_i p_i^\alpha\right) = -\log \left(\sum_i \left(\frac{\sigma_i}{\sum_{j} \sigma_j}\right)^2\right).
\end{align}
Throwing away the out-of-summation logarithm, we define the soft-rank:
\begin{definition}
    Given a symmetric PSD matrix $\mat{A}$, its soft-rank is defined as
    \begin{align}
        \erank{\mat{A}} \defeq \frac{1}{\sum_i \left(\frac{\sigma_i(\mat{A})}{\sum_{j} \sigma_j(\mat{A})}\right)^2} = \frac{\trace{\mat{A}}^2}{\trace{\mat{A}^2}}.
    \end{align}
\end{definition}
It is easy to verify that $\erank{\mat{A}} \le \text{rank}(\mat{A})$. Moreover, when $\mat{A}$ has $\text{rank}(\mat{A})$ non-zero singular values and all these singular values are equal (say $\sigma$), we have $\erank{\mat{A}} \defeq \frac{(\text{rank}(\mat{A}) \cdot \sigma)^2}{\text{rank}(\mat{A}) \cdot \sigma^2} = \text{rank}(\mat{A})$, justifying it is soft version of standard rank. 
Moreover, we have the following proposition that helps us understand the low-rank structure of $\mJ$.
\begin{proposition}\label{prop:kronecker_erank}
    Let $\mA$ and $\mB$ be two symmetric PSD matrices. Then
    \begin{align}
        \erank{\mA \kronecker \mB} 
        =& \erank{\mA} \cdot \erank{\mB}.
    \end{align}
    Let $\set{\mA_i}$ be a sequence of symmetric PSD matrices. Then
    \begin{align}
        \erank{\Diag(\mA_i)} \le \sum_i \erank{\mA_i}.
    \end{align}
\end{proposition}
Its proof can be found in \cref{appendix:technical_lemmas} in \supplementary{}.

\subsection{Assumptions of Theoretical Results}\label{sec:assumptions}

To understand the source of \AdversarialAlignment{}, we derive expressions and lower-bounds of its first-step value under the assumptions of deep linear networks (DLN), whitened data, $\ltwo$ regularization, and sufficient training. Here, we discuss the motivations and necessities of these assumptions.
The formal results and proofs can be found in \cref{appendix:proofs} in the \supplementary{}.

Forgetting can be measured by the increase in the empirical loss or in the testing loss.
We acknowledge that the testing loss is more important in practice. However, it involves an analysis of generalization.
On the other hand, a low empirical loss is the basis of a low testing loss. Moreover, forgetting and alignment are already evident under empirical loss, where generalization is not involved. Therefore, forgetting and the alignment have unignorable causes in training, and we consider one must study forgetting in empirical loss before considering testing loss.
As a result, we mainly focus on the forgetting in the \emph{empirical} loss, and all losses and samples are empirical losses and training samples.

We only consider the first-step \AdversarialAlignment{} because (1) according to \cref{figure:existence,figure:ablations}, the initial steps of new-task training have strong \AdversarialAlignment{} and (2) the only-first-step analysis is more technically feasible. Analysis of multi-step training dynamics of DLN is a separate active research topic, and devoting too much effort to it is out of the scope of this paper.

We also assume deep linear networks (DLN) for technical feasibility. Although highly simplified, this model shares non-convexity and multiple local minima with deep neural networks. Importantly, given the training data, the multiple local minima have different local curvatures. As a result, local curvatures are also affected by the implicit bias in the training, unlike shallow linear regression, where local curvatures of minima are solely determined by the data.
This allows old-task weights to shape the old-task Hessian.

Now we turn to assumptions on the data, especially the new-task data. Since we care about \AdversarialAlignment{} under dissimilar tasks, we must create some dissimilarity between the old and new tasks.
On the other hand, if the new task is somehow adversarially anti-dissimilar, the \AlignmentPhenomenon{} may not be strong enough to exist. Inspired by existing works where the new task is generated by random permutation of pixels, we generate the new task by random rotation of the old task data by a random orthogonal matrix. When the input has a high dimension, neither the very similar nor the very anti-similar new task will be generated.
We assume whitened input data, \ie{} $\OldInputs \OldInputs^\top = \mat{I}$, which is a common assumption in \CL{} or other theoretical works.

We also assume standard $\ltwo$ regularization. It induces the low-rankness of weights. Moreover, it implies auto-balancedness (see \cref{lemma:auto-alignment,lemma:auto_alignment_implication_singular_values}) between adjacent weights, making it easier to simplify their products and prove them also low-rank. 
We remark that these auto-balanced and low-rank properties can be achieved by implicit bias of (S)GD alone, which is an active topic of optimization and generalization of DLNs \citep{dln_bias_1,dln_bias_2,implicit_auto_balancedness}. 
Therefore, the $\ltwo$ regularization assumption is potentially dispensable in theoretical analysis. 
However, such analysis requires an every-direction auto-balancedness, while existing works only bound the imbalance along the worst direction, \ie{} bounding $\snorm{\weight_2 - \weight_1^\top}$ \citep{implicit_auto_balancedness}.
We believe such potential technical improvement is more an issue of optimization research, and putting too much effort on it may deviates from our focus on \CL{}.

Lastly, we assume sufficient training on the old task under $\ltwo$ regularization so that (1) the empirical samples are interpolated, and (2) a local minimum of the regularized empirical loss is reached.
The assumption is motivated by the fact that old tasks are usually sufficiently trained in \CL{}, and helps us obtain auto-balancedness of each-layer weights and how each-layer weights connect to the old-task training data.
Similar assumptions have been adopted by \citet{meat_learning_as_regularization}, where they are used to argue that in the second-order approximation of old-task loss changes, the first-order term is negligible and the forgetting is dominated by the second-order terms.
Many shallow-layer theories \citep{interest_in_forgetting_mechanism4,interest_in_forgetting_mechanism5} also adopt similar assumptions for obtaining explicit expression of the weights after the old- and new-task training.

Based on these assumptions, we prove the lower-bound \cref{eq:lowerbound_interpretable}
.
We also derive a tighter but more complicated lower-bound
\newcommand{\GtSV}{\SingularMatrix{\OldLabels \OldInputs^\top}}
\begin{align}
        \alignment
        \gtrapprox& \tightbound \\
        \defeq& 
            \frac{\substack{
                \left(
                    1
                    +
                    \frac{1}{L^2 \cdot \dim \sampleinput} \sum_{i=1}^L \sum_{j=1}^L \erank{\GtSV^{2\max(i+j-2, 3L-(i+j)) / L}}      
            \right)}}{\substack{
                \frac{\sum_{i=1}^L\erank{\GtSV^{2\min(i-1, L-i)/L}}}{L} 
                 \left(
                    1
                    + 
                    \frac{\sum_{i=1}^L \erank{\GtSV^{2\min(i-1, 2L-i) / L}}}{L \cdot \dim \sampleinput} 
                \right)
            }}
        \cdot \frac{\dim \param}{\erank{\GtSV^{2(1-1/L)}}}.\label{eq:lowerbound_tighter}
\end{align}
for empirical verification.
The full formal statements are in \cref{theorem:alignment_lowerbound_interpretable,theorem:alignment_lowerbound_tighter} in \supplementary{}, which explicitly reflect the dependence on the degree of interpolating the old task by $\tau$ and $\rho$.
In the main text, we ignore such dependence by assuming the interpolation is nearly perfect (\eg{} when the $\ltwo$ regularization is infinitesimally weak). This assumption lets us take $\tau \to 0, \rho \to 0, \weight_{L:1} \to \oldYX$ and $\erank{\weight_{L:1}^{2(1 - 1/L)}} \to \erank{\GtSV^{2(1 - 1/L)}}$, which are the source of the approximation in the ``$\gtrapprox$'' inequalities. To approximate this assumption, experiments are configured to include sufficient training so that the old-task empirical loss is very low. We remark that the bounds is not very sensitive to the interpolation errors $\tau, \rho$ with linear dependence. Therefore, it is reasonable to ignore it under the suitably configured experiments in the main text for clarity and simplicity.

%% file: methods/algorithm.tex
\subsection{Effectiveness of Gradient Projection Methods}\label{sec:analysis_of_existing_methods}

\newcommand{\mGP}{\mG^{\text{GP}}}
\newcommand{\mJGP}{\mJ^{\text{GP}}}
\newcommand{\vgp}{\vg^{\text{GP}}}

We apply our theoretical findings to understand the effectiveness and limitations of existing \CL{} algorithms. We focus on \GP{} (GP) methods since they have not been related to \AlignmentPhenomenon{} with \LargeCurvature{} directions by \citet{hessian_perspective_of_regularization_CL} and \citet{meat_learning_as_regularization}.

GP methods alleviate forgetting by projecting the new-task gradients onto the subspace that is orthogonal to the old-task gradients.
This is typically done by projecting the new-task gradients \wrt{} linear layers onto the (approximated) null space of old-task input covariances:
\begin{align}
    \mGP_i 
    \defeq& \mG_i \approxnull{\hidden{i-1}_1 \left(\hidden{i-1}_1\right)^\top}
    = \mG_i \approxnull{\weight_{i-1:1} \OldInputs \OldInputs^\top \weight_{i-1:1}^\top}\\
    \approx& \mG_i 
    \re{
    \underbrace{
    \mU_{i-1} \begin{bmatrix}
        \mat{0} & \mat{0} \\
        \mat{0} & \mat{\Sigma}_{\text{nuisance}}^{i-1} 
    \end{bmatrix} \mU_{i-1}^\top
    }_{\text{gradient projection}}}
    = \mG_i 
    \re{
    \mU_{i-1} \begin{bmatrix}
        \mat{0} & \hphantom{\approx}\mat{0} \\
        \mat{0} & \nearzero
    \end{bmatrix} \mU_{i-1}^\top
    }
\end{align}
where $\mGP_i$ is the projected gradient, $\hidden{i}_1$ stacks the (hidden) inputs to the $i$-th linear layer of the old task, and $\approxnull{\mat{A}} = \mat{U} \mat{U}^\top$ with $\mat{U}$ being the unit orthogonal bases of the null space of $\mat{A}$ or its rank-$r$ approximation. As a result, the projected gradients will not or only minimally change the hidden features of old tasks, \ie{}
\begin{align}
    \Delta \hidden{i}_1 
    =& \left(\weight_i + \eta \cdot \mGP_i\right) \hidden{i-1}_1 - \weight_i \hidden{i-1}_1 \\
    =& \eta \cdot \mG_i \underbrace{\approxnull{\hidden{i-1}_1 \left(\hidden{i-1}_1\right)^\top} \hidden{i-1}_1}_{\approx 0}
    \approx 0, 
\end{align}
and induce less forgetting.

The effectiveness of these methods can also be understood from mitigating \AdversarialAlignment{} with \LargeCurvature{} directions.
Although they are designed following the intuition of avoiding alignment with old-task gradients, it is not the major contribution because the non-mitigated forgetting caused by old-task gradient alignment is ignorable according to \cref{figure:all_forgettings_cifar100,figure:all_forgettings_cross_modal,figure:all_forgettings_randrot_mnist}.
As a result, avoiding gradient alignment can at most decrease the forgetting by a small amount.
On the other hand, considerable forgetting is found in the second-order approximation. We find GP methods are also effective in avoiding \AdversarialAlignment{} and mitigating this second-order forgetting. 
\ifExplicitHessian
    To see this, we compute which subspace the new-task \emph{projected} gradient reside. After some calculations, we obtain
    \begin{align}
        \vgp =& \mJGP \times \left(\vec{1}_L \kronecker \vectorize{\partialfrac{\NewLoss}{f_{\OldParam}} \NewInputs^\top}\right) 
    \end{align}
    where projected Jacobian $\mJGP \defeq \Diag \left(\left(\approxnull{\weight_{i-1:1} \OldInputs \OldInputs^\top \weight_{i-1:1}^\top} \nonTransparentUnderbrace{\weight_{i-1:1}}{forward}\right) \kronecker \nonTransparentUnderbrace{\weight_{L:i+1}^\top}{backward} \right)$. Note that $\mJGP$ differs from $\mJ$ only by the nullspace projector introduced by GP methods.
    \emph{When} $\weight_{i-1:1}$ is low-rank, so would be $\weight_{i-1:1} \OldInputs \OldInputs^\top \weight_{i-1:1}^\top$, whose nullspace projector will accurately remove the principal component $\Signal^{i-1}$ of $\weight_{i-1:1}$. Recall that this component determines the column space of $\mJ$, which further determines the subspace where the new-task gradients reside. As a result, removing such component will push the new-task gradients toward to a subspace that is orthogonal to the subspace where the old-task \LargeCurvature{} directions lie.
    We verify this by computing the product between $\mJ^\top$ and $\mJGP$, which is $0$ when the column spaces of two matrices are orthogonal:
    \begin{align}
        &\mJ^\top \times \mJGP\\
        =&\Diag\left(\transparentUnderbrace{\weight_{i-1:1}^\top}{forward} \kronecker \transparentUnderbrace{\weight_{L:i+1}}{backward}\right)
        \times 
            \Diag \left(\left(\approxnull{\weight_{i-1:1} \OldInputs \OldInputs^\top \weight_{i-1:1}^\top} \transparentUnderbrace{\weight_{i-1:1}}{forward}\right) \kronecker \transparentUnderbrace{\weight_{L:i+1}^\top}{backward} \right)\\
        =&
            \Diag \left(\left(\transparentUnderbrace{\weight_{i-1:1}^\top \approxnull{\weight_{i-1:1} \OldInputs \OldInputs^\top \weight_{i-1:1}^\top} \weight_{i-1:1}}{forward; $\approx \mat{0}$ \emph{for deep layers with low-rank $\weight_{i-1:1}$}}\right) \kronecker \left(\transparentUnderbrace{\weight_{L:i+1} \weight_{L:i+1}^\top}{backward}\right) \right). \label{eq:j_jgp_product}
    \end{align}
    Therefore, GP methods push new-task updates to the subspace that is orthogonal to the old-task \LargeCurvature{} directions, thereby mitigating \AdversarialAlignment{} and forgetting.
    This is empirically verified in \cref{figure:ablations}, where \AdversarialAlignment{} is drastically decreased by GP methods.
\else
    To see this, we compute the alignment between the old-task Hessian and the new-task projected gradient in DLN:
    \begin{align}
        &\ex{\left(\vgp\right)^\top \oldhessian \vgp} 
        =\Fnorm{ \sum_{i=1}^L \weight_{L:i+1} \mGP_i \weight_{i-1:1} \OldInputs}^2\\
        =&\Fnorm{ \sum_{i=1}^L \weight_{L:i+1} \left(\weight_{L:i+1}^\top (\weight_{L:1} \NewInputs - \NewLabels) \NewInputs^\top \weight_{i-1:1}^\top\right) \approxnull{\weight_{i-1:1} \OldInputs \OldInputs^\top \weight_{i-1:1}^\top} \weight_{i-1:1} \OldInputs}^2\\
        \approx&
            \Fnorm{ \sum_{i=1}^L (\dots) \times \NewInputs^\top \TransposedSimplifiedConsecutiveProduct{i-1}{1}[i-1]\mU_{i-1} \begin{bmatrix}
            \mat{0} & \hphantom{\approx}\mat{0} \\
            \mat{0} & \nearzero
        \end{bmatrix} \mU_{i-1}^\top \SimplifiedConsecutiveProduct{i-1}{1}[i-1]\OldInputs}^2.\\
        =&
            \Fnorm{ \sum_{i=1}^L (\dots) \times \NewInputs^\top \TransposedSimplifiedConsecutiveProduct{i-1}{1}[i-1][\mat{0}]\mU_{i-1} \begin{bmatrix}
            \mat{0} & \hphantom{\approx}\mat{0} \\
            \mat{0} & \nearzero
        \end{bmatrix} \mU_{i-1}^\top \SimplifiedConsecutiveProduct{i-1}{1}[i-1][\mat{0}]\OldInputs}^2.\\
    \end{align}
    When $\weight_{i-1:1}$ is low-rank enough, the projection eliminates the principal components of $\NewInputs^\top \weight_{i-1:1}^\top \times \weight_{i-1:1} \OldInputs$ and breaks the \AdversarialAlignment{} induced by the forward propagation.
    As a result, \AdversarialAlignment{} and forgetting are mitigated.
    This is empirically verified in \cref{figure:ablations}, where \AdversarialAlignment{} is drastically decreased by GP methods.
\fi

\subsection{Limitation of Gradient Projection Methods and Resolution}\label{sec:resolution_of_limitations}

Analysis in \cref{sec:interpretation} indicates \AdversarialAlignment{} is induced by both forward and backward propagation of the new-task training. Among them, \cref{eq:j_jgp_product} in \cref{sec:analysis_of_existing_methods} suggests GP methods have handled the alignment induced by forward propagation, but leave the backward-related part $\weight_{L:i+1} \weight_{L:i+1}^\top$ intact.
Since at shallow layers, the forward-related part $\weight_{i-1:1}$ is not low-rank (\eg{} $\weight_{1-1:1}=\mI$) but the backward-related part $\weight_{L:i+1}$ has low-rankness and contributes to the most of the alignment, the existing GP methods miss the main drive and may leave residual \AdversarialAlignment{} at shallow layers.
For a concrete failure, we focus on layer $i=1$, the alignment between the layer-$1$ components of the new-task updates and old-task \LargeCurvature{} directions, and the layer-$1$ component $\mJ_1 \defeq \weight_{1-1:1} \kronecker \weight_{L:1+1}^\top$ of $\mJ$. We compute the top eigenvectors of the old-task Hessian \emph{\wrt{} the shallow layer $\weight_1$} as $\set{(\vec{e}_j \kronecker \HessianEigVec_k, \HessianEigVal_k) \mid j \in \set{1, \dots, \dim \Input}, (\HessianEigVec_k, \HessianEigVal_k) \in \text{TopEig}(\weight_{L:i+1}^\top \weight_{L:i+1})}$, where $\vec{e}_j$ is the $j$-th standard basis vector and $\text{TopEig}(\cdot)$ denotes the set of top eigenvector-eigenvalue pairs.
This suggests the old-task \LargeCurvature{} directions span the entire the principal subspace of $\mat{A} \kronecker \weight_{L:i+1}^\top$'s column space 
\begin{align}
    \text{span}(\set{\vec{u}_j \kronecker \HessianEigVec_k \mid (\vec{u}_j, \mu_j) \in \text{TopEig}(\mat{A} \mat{A}^\top), (\HessianEigVec_k, \HessianEigVal_k) \in \text{TopEig}(\weight_{L:1+1}^\top \weight_{L:1+1})})
\end{align}
whatever $\mat{A}$ is. Thus manipulating only the forward-related part in $\mJ_1 \defeq \weight_{1-1:1} \kronecker \weight_{L:1+1}^\top$ will always project the new-task updates \wrt{} $\weight_1$ to the subspace spanned by the \LargeCurvature{} directions. 
As a result, the shallow-layer components of the new-task gradient and old-task \LargeCurvature{} directions still align adversarially, which also implies global \AdversarialAlignment{}.
This \emph{residual adversariality} is confirmed empirically in \cref{figure:ablations}, where considerable \AdversarialAlignment{} of $\alignment \sim 10^2$ still exists after applying an existing GP method.

\newcommand{\mbGP}{\mG^{\text{\backGP}}}
We alleviate the limitation by inserting nullspace projections in the backward direction. Such projections will come into effect the same way as in \cref{sec:analysis_of_existing_methods}.
We need to identify and eliminate the principal components in the backward multiplication. 
Ideally, we can left-multiply $\approxnull{\weight_{L:i+1} \weight_{L:i+1}^\top} \approx     \mV_{i+1} \begin{bmatrix}
    \mat{0} & \mat{0} \\
    \mat{0} & \mat{\Sigma}_{\text{nuisance}}^{L-i} 
\end{bmatrix} \mV_{i+1}^\top$ to remove the principal components of the adversariality-inducing low-rank projection $\weight_{L:i+1} \weight_{L:i+1}^\top$.
However, such construction is only available in DLNs and is hard to extend to non-linear networks.
Noting that multiplication with effectively low-rank projections will approximately inherit their principal components, we use the null space of $\weight_{L:i+1}$ multiplied with gradients \wrt{} model output, $\approxnull{\weight_{L:i+1} \partialfrac{\OldLoss}{(\weight_{L:1} \OldInputs)} \left(\partialfrac{\OldLoss}{(\weight_{L:1} \OldInputs)}\right)^\top \weight_{L:i+1}^\top} = \approxnull{\partialfrac{\OldLoss}{\hidden{i}_1} \left(\partialfrac{\OldLoss}{\hidden{i}_1}\right)^\top}$, which is the gradients \wrt{} the hidden outputs and can be extended to non-linear networks.
This intuition leads to our backward GP (\backGP{}) method, which is a mirrored version of existing (forward) GP methods:
\begin{align}
    \mbGP_i 
    \defeq& \bl{\approxnull{\partialfrac{\OldLoss}{\hidden{i}_1} \left(\partialfrac{\OldLoss}{\hidden{i}_1}\right)^\top}}  \mG_i \re{\approxnull{\hidden{i-1}_1 \left(\hidden{i-1}_1\right)^\top}}\\
    \approx&     \bl{
    \underbrace{
    \mV_{L-i} \begin{bmatrix}
        \mat{0} & \mat{0} \\
        \mat{0} & \mat{\Sigma}_{\text{nuisance}}^{L-i} 
    \end{bmatrix} \mV_{L-i}^\top
    }_{\text{backward gradient projection}}}  \mG_i \re{
    \underbrace{
    \mU_{i-1} \begin{bmatrix}
        \mat{0} & \mat{0} \\
        \mat{0} & \mat{\Sigma}_{\text{nuisance}}^{i-1} 
    \end{bmatrix} \mU_{i-1}^\top
    }_{\text{gradient projection}}}.
\end{align}

Now we describe the details in implementing \backGP{}.
We make \backNSCL{} a plugin for existing GP methods.
For simplicity, we let \backNSCL{} resemble the most basic GP method, AdamNSCL \citep{nscl}, but in the backward direction.
From now on, we assume there are $T$ tasks, and use an additional subscript $(\cdot)_{t, \dots}$ to denote the task that the variable belongs to.
For task $t$, let column vector $\vec{z}_{t, i, k}$ denote the $k$-th (hidden) output of the $i$-th linear layer, where $k$ iterates over samples and patches/tokens in the training dataset.

The first task is trained using standard gradient descent or its variants.
When training task $t > 1$, we collect the gradients of previous tasks $\tau < t$ \wrt{} the linear layers' hidden outputs, \ie{} $\begin{bmatrix}
    \partialfrac{\emploss_\tau}{\vec{z}_{\tau,i, k}}
\end{bmatrix}_{\cdot, k} = \partialfrac{\emploss_\tau}{\mat{Z}_{\tau, i}}$.
Then we compute the gradient covariance of all past tasks $\mM_{t, i} \defeq \sum_{\tau < t} \partialfrac{\emploss_\tau}{\mat{Z}_{\tau, i}} \left(\partialfrac{\emploss_\tau}{\mat{Z}_{\tau,i}}\right)^\top = \mM_{t-1, i} + \partialfrac{\emploss_{t-1}}{\mat{Z}_{t-1,i}} \left(\partialfrac{\emploss_{t-1}}{\mat{Z}_{t-1,i}}\right)^\top$.
We then compute the eigenvalue decomposition $\mM_{t, i} = \mV_{\mM_{t, i}} \mat{\Lambda}_{\mat{M}_{t, i}} \mV_{\mM_{t, i}}^\top \defto \sum_{j} \lambda_j \vec{v}_j \vec{v}_j^\top$ and remove the principal components to obtain the (approximated) nullspace. Specifically, let $\epsilon_{\backNSCL} \in (0, 1)$ be the hyperparameter meaning the larger it is, the more spectrum is retained in the approximate nullspace and the more ``parameter'' will be updated. Then let $\approxnull{\mM_{t, i}}[\epsilon_{\backNSCL}] \defeq \sum_{j \ge \dim \mM_{t, i} - r} \vec{v}_j \vec{v}_j^\top$ be the approximated nullspace's projector, where $r$ is the largest integer such that $\frac{\sum_{j \ge \dim \mM_{t, i} - r} \sigma_j}{\sum_{j} \sigma_j} \leq \epsilon_{\backNSCL}$.
The projection of the backward-side direction is the projector, \ie{} $\mat{B}_{t,i} \defeq \approxnull{\mM_{t, i}}[\epsilon_{\backNSCL}]$.

Our \backNSCL{} is intended to be combined with existing GP methods. Therefore, let $\mF_{t, i}$ be the forward-side projection given by the to-be-combined GP method, such that it would be used as $\mG^{\text{GP}}_{t, i} \defeq \mG_{t, i} \mF_{t, i}$ by the existing GP method. 
With projections of both sides, the update during task-$t$-training is given by 
\begin{align}
    \mG^{\backNSCL{}}_{t, i} \defeq \mat{B}_{t, i} \mG_{t, i} \mF_{t, i}.
\end{align}

\subsection{Details of Experiments}\label{sec:details_of_experiments}

Before running experiments, we acknowledge that GPs are already highly efficient in reducing forgetting, making forgetting no longer a major issue at least for standard \CL{} benchmarks.
Its constraints on new-task gradients have been considered too strict, and major efforts have turned to relaxing the constraints and increasing plasticity \citep{sgp,forgetting_and_flatness2,gpcns,adaNSCL}.  
However, another line of studies find vanilla deep neural networks lose plasticity during continual learning even when no constraints are put on the new-task gradients \citep{loss_of_plasticity1,loss_of_plasticity2,loss_of_plasticity3}. It suggests plasticity loss is partially caused by \emph{non-GP reasons} \citep{other_loss_of_plasticity} and there may exist plasticity-boosting methods other than relaxing GP constraints.
These works lead to spectral regularizers to improve plasticity \citep{spectral_regularization1,spectral_regularization2}, although they do not consider forgetting. 
Empirically, we find adding a simple spectral or orthogonality regularization \citep{orthogonality_regularization}
\begin{align}
    \spectralReg(\weight_i) \defeq \Fnorm{\weight_i \weight_i^\top - \mI}^2
\end{align}
and using modern architectures (\eg{} ConvNeXts instead of ResNets) can boost GP's plasticity and performance. According to \cref{table:comparison}, the drastic plasticity improvements of $\ge 10\%$ further improve the final performance.
However, such plasticity improvements lead to drastically more forgetting ($\sim 10\%$), re-making forgetting the major issue in high-plasticity \CL{}.
Therefore, we apply \backNSCL{} to alleviate such residual forgetting.

We use 10- and 20-split CIFAR100 and 25-split TinyImageNet benchmarks. 
We use recent GP methods as well as their spectral-regularized versions as baselines. Our \backNSCL{} is combined with the spectrally regularized GP methods.
Since BatchNorm layers' parameters are not protected by GP methods and require special treatment, we do not use BatchNorm-involving ResNets as in many \CL{} literature. Instead, we use ConvNeXt \citep{convnext} with affine-transform-free LayerNorm layers as the backbone, whose block configuration is summarized in \cref{table:convnext_blocks}. 
No pretraining is used.

We use different classification heads for each task as \citet{nscl,gpm,sgp}. 
We train all model parameters during the first task. In later tasks, we freeze parameters that are not protected by GP, including all LayerNorm parameters, all biases, and all \texttt{layer\_scale} parameters.
We also freeze linear or convolution layers whose input dimension (=$\langle\text{channel}\rangle \times \langle\text{kernel size}\rangle^2$ for convolution layers) or output dimension is smaller than $64$, since the corresponding features or gradients often have approximately uniform input spectrum and every subspace has too much spectrum to be ignored. 
Lastly, for each convolution-linear-activation-linear structure in ConvNeXt, we freeze the first linear layer. Otherwise, baseline methods still exhibit catastrophic forgetting. 
Detailed hyperparameters can be found in \cref{table:hyperparameters}.
Specifically, the gradient of spectral regularization is applied outside of Adam, \ie{} in an AdamW manner.
We also use a large learning rate for the classifier and a small learning rate for the backbone, following \citet{nscl}.


%% file: results/appendix.tex
\section{Experiment Hyperparameters}\label{appendix:hyperparameters}

\subsection{Hyperparameters of Experiments in \cref{sec:existence}}
\FloatBarrier
 
The hyperparameters of experiments in \cref{sec:existence} are summarized in \cref{table:verification_hyperparameters}.

\begin{table}
    \centering
    \caption{Hyperparameters of experiments that verify the existence of \AdversarialAlignment{}.}\label{table:verification_hyperparameters}
    \begin{tabular}{lccccccc}
        \toprule 
        & \multicolumn{3}{c}{CIFAR100} & \multicolumn{2}{c}{Cross-Modal} & Synthetic MNIST\\ 
        \cmidrule(lr){2-4}
        \cmidrule(lr){5-6}
        \cmidrule(lr){7-7}
        Architecture & ResNet & ViT & MLP-Mixer
        & ViT & MLP-Mixer
        & DLN\\
        \midrule
        Model size          & ResNet18  & Small     & Small             &   Small   & Small         & $L\in \set{1,2,4,6,8,10}$\\
        \midrule
        Epoch               & 100       & 200       & 100               &  200      & 100           & 200\\
        Batch size          & 16        & 32        & 32                & 16        & 16            & 512\\
        Optimizer           & SGD       & SGD       & SGD               & SGD       & SGD           & SGD\\
        Learning rate       & 0.001     & 0.01      & 0.9               & 0.001     & 0.001         & 0.5\\
        Momentum            & 0.9       & 0.9       & 0.9               &   0.9     & 0.9           & 0.0\\
        $\ltwo$ reg.        & 0.0001    & 0.0001    & 0.0001            & 0.0001    & 0.0001        & 0.001\\
        Data aug.         & \multicolumn{5}{c}{RandomCrop(size=224), RandomHorizontalFlip} & Whitening\\
        \bottomrule
    \end{tabular}
\end{table}

\subsection{Hyperparameters of Experiments in \cref{sec:application}}

The block configuration of ConvNeXt is summarized in \cref{table:convnext_blocks}.
We use a small patch size of $1$ and kernel sizes of $3$ to accommodate the small image sizes of CIFAR100 and TinyImageNet. We also control the number of blocks in each stage to obtain a model size similar to ResNet18 used by previous works.
We remove the biases in LayerNorm layers. We also reduce both the \texttt{layer\_scale} parameters in residual blocks and the elementwise affine parameters in LayerNorm layers to scalars.
The goal is to reduce the number of parameters that are not protected by GP.

\begin{table}
    \caption{Block configuration of ConvNeXt.}\label{table:convnext_blocks}
    \begin{tabular}{lccccc}
        \toprule
        &   Input channel & Hidden channel & Output channel & Number of blocks & Kernel size\\
        \midrule
        Embedding & 3 & N/A & 64 & N/A & 1\\
        Stage 1 & 64 & 256 & 128 & 4 & 3\\
        Stage 2 & 128 & 512& 256 & 3 & 3\\
        Stage 3 & 256 & 1024 & 512 & 3 & 3\\
        Stage 4 & 512 & 2048 & 512 & 4 & 3\\
        \bottomrule
    \end{tabular}
\end{table}

Then we select the hyperparameters of the experiments, which are summarized in \cref{table:hyperparameters}.
The process of determining the hyperparameters is as follows:
(1) selecting the $\acc$-best hyperparameters for each \texttt{<baseline> + $\spectralReg$} method using grid search over SVD/GPM thresholds and scale coefficients, where plasticity is more preferred between hyperparameters with similar $\acc$; 
(2) replacing the spectral regularization with $\ltwo$ regularization and running the experiments for \texttt{<baseline>} methods;
(3) using the same backward SVD threshold as the forward SVD/GPM thresholds for \texttt{<baseline> + $\spectralReg$ + \backNSCL} methods;
(4) increasing the forward or backward SVD/GPM thresholds of \texttt{<baseline>} + $\spectralReg$ + \backNSCL{} methods if too much plasticity is lost.

\begin{table}
    \centering
    \caption{Hyperparameters of experiments for evaluation and comparison.}\label{table:hyperparameters}
    \begin{tabular}{lccc}
        \toprule
        & 10-split CIFAR100 & 20-split CIFAR100 & TinyImageNet-25\\
        \midrule
        Optimizer & \multicolumn{3}{c}{AdamW}\\
        Batch size & \multicolumn{3}{c}{128}\\
        Epoch & \multicolumn{3}{c}{400}\\
        \midrule
        Initialization\\
        \quad \texttt{Linear} & \multicolumn{3}{c}{The default in \texttt{PyTorch}: Kaiming uniform initialization with $a=\sqrt{5}$}\\
        \quad \texttt{Conv} & \multicolumn{3}{c}{The default in \texttt{PyTorch}: Kaiming uniform initialization with $a=\sqrt{5}$}\\
        \quad \texttt{layer\_scale} & \multicolumn{3}{c}{$10^{-6}$}\\
        \midrule
        Learning rate\\
        \quad Classifier & $1\times 10^{-2}$ & $1\times 10^{-2}$ & $1\times 10^{-2}$\\
        \quad Backbone & $3 \times 10^{-3}$ & $3 \times 10^{-3}$ & $5 \times 10^{-4}$\\
        Scheduler & OneCycle & OneCycle & OneCycle\\
        \midrule
        Data aug. & \multicolumn{2}{c}{\begin{tabular}{c}
            RandomHorizontalFlip\\
            AutoAugment(policy=CIFAR10)\\
            RandomCrop(size=32, padding=4)
        \end{tabular}} & \begin{tabular}{c}
            RandomHorizontalFlip\\
            AutoAugment(policy=ImageNet)\\
            RandomCrop(size=64, padding=8)\\
            RandAugment\\
            RandomErasing(p=0.25)\\
        \end{tabular}\\
        \midrule
        $\ltwo$ reg.\optional & \multicolumn{3}{c}{$1.0 \times 10^{-2}$}\\
        Spectral reg.\optional\\
        \quad First task & $3.0 \times 10^{-2}$ & $3.0 \times 10^{-2}$ & $1.0 \times 10^{-1}$\\
        \quad Other tasks & $0$ & $0$ & $0$\\
        \midrule
        AdamNSC\optional\\
        \quad $\epsilon_{\text{SVD}}$ & $0.30$ & $0.20$ & $0.20$\\
        \quad $\epsilon_{\backGP}$\optional & $0.30$ & $0.20$ & $0.20$\\
        \midrule
        GPM\optional\\
        \quad $\epsilon_{\text{GPM}}$ & $0.20$ & $0.10$ & $0.10$\\
        \quad $\epsilon_{\backGP}$\optional & $0.30$ & $0.15$ & $0.10$\\
        \midrule
        SGP\optional\\
        \quad $\epsilon_{\text{GPM}}$ & $0.30$ & $0.20$ & $0.30$\\
        \quad Scale coeff.     & $5$ & $10$ & $5$\\  
        SGP + \backGP\optional\\
        \quad $\epsilon_{\text{GPM}}$ & $0.30$ & $0.40$ & $0.30$\\
        \quad Scale coeff.    & $5$ & $5$ & $5$\\  
        \quad $\epsilon_{\backGP}$\optional & $0.45$ & $0.40$ & $0.75$\\
        \bottomrule
    \end{tabular}
    $\dagger$ means the hyperparameter is used only when the corresponding component is turn-ed on.
\end{table}

\FloatBarrier

\section{More Experiment Results}

\subsection{More Verification of Adversarial Alignment}\label{appendix:verification}

\FloatBarrier

In the experiment reported by \cref{figure:existence}, we repeat 5 trials for each setting but only 1 trial is reported in the main text. The rest 4 CDF diagrams for each setting are reported in \cref{fig:more_existence_cifar100,fig:more_existence_cross_modal,fig:more_existence_randrot_mnist}, where similar conclusions can be drawn.

\begin{figure}
    \newcommand{\ypad}{{\tiny\quad \quad \,\,\,}}
    \newcommand{\xpad}{{\tiny \quad\,\,}}
    \centering
    \setstretch{0.2}
    \begin{subfigure}{\linewidth}
        \centering
        \begin{subfigure}{0.025\linewidth}%
            \rotatebox{90}{\labelsize \ypad Projection CDF}%
        \end{subfigure}%
        \foreach \i in {2,3,4,5}{%
            \begin{subfigure}{0.23\linewidth}%
                \centering
                \includegraphics[width=\linewidth]{pic/appendix/cdf/resnet18_\i.pdf}\\
                \labelsize \xpad Eigenvalue
            \end{subfigure}%
        }
        \caption{ResNet}
    \end{subfigure}
    \begin{subfigure}{\linewidth}
        \centering
        \begin{subfigure}{0.025\linewidth}%
            \rotatebox{90}{\labelsize \ypad Projection CDF}%
        \end{subfigure}%
        \foreach \i in {2,3,4,5}{%
            \begin{subfigure}{0.23\linewidth}%
                \centering
                \includegraphics[width=\linewidth]{pic/appendix/cdf/vit_small_\i.pdf}\\
                \labelsize \xpad Eigenvalue
            \end{subfigure}%
        }
        \caption{VisionTransformer}
    \end{subfigure}
    \begin{subfigure}{\linewidth}
        \centering
        \begin{subfigure}{0.025\linewidth}%
            \rotatebox{90}{\labelsize \ypad Projection CDF}%
        \end{subfigure}%
        \foreach \i in {2,3,4,5}{%
            \begin{subfigure}{0.23\linewidth}%
                \centering
                \includegraphics[width=\linewidth]{pic/appendix/cdf/mixer_\i.pdf}\\
                \labelsize \xpad Eigenvalue
            \end{subfigure}%
        }
        \caption{MLP-Mixer}
    \end{subfigure}
    \caption{10-Split CIFAR100}\label{fig:more_existence_cifar100}
\end{figure}

\begin{figure}
    \newcommand{\ypad}{{\tiny \quad \quad \quad}}
    \newcommand{\xpad}{{\tiny \qquad}}
    \centering
    \setstretch{0.2}
    \foreach \depth in {2,4,6,8,10}{
        \begin{subfigure}{\linewidth}
            \centering
            \begin{subfigure}{0.025\linewidth}%
                \rotatebox{90}{\labelsize \ypad Projection CDF}%
            \end{subfigure}%
            \foreach \i in {2,3,4,5}{%
                \begin{subfigure}{0.23\linewidth}%
                    \centering
                    \includegraphics[width=\linewidth]{pic/appendix/cdf/dln_\depth_\i.pdf}\\
                    \labelsize \xpad Eigenvalue
                \end{subfigure}%
            }%
            \caption{$L=\depth$}
        \end{subfigure}%
    }
    \caption{Synthetic Randomly Rotated Whitened MNIST Tasks}\label{fig:more_existence_randrot_mnist}
\end{figure}

\begin{figure}
    \newcommand{\ypad}{{\tiny \quad \quad \quad}}
    \newcommand{\xpad}{{\tiny \quad\,\,}}
    \setstretch{0.2}
    \centering
    \begin{subfigure}{\linewidth}
        \centering
        \begin{subfigure}{0.025\linewidth}%
            \rotatebox{90}{\labelsize \ypad Projection CDF}%
        \end{subfigure}%
        \foreach \i in {2,3,4,5}{%
            \begin{subfigure}{0.23\linewidth}%
                \centering
                \includegraphics[width=\linewidth]{pic/appendix/cdf/cv_nlp_vit_small_\i.pdf}\\
                \labelsize \xpad Eigenvalue
            \end{subfigure}%
        }
        \caption{VisionTransformer}
    \end{subfigure}
    \begin{subfigure}{\linewidth}
        \centering    
        \begin{subfigure}{0.025\linewidth}%
            \rotatebox{90}{\labelsize \ypad Projection CDF}%
        \end{subfigure}%
        \foreach \i in {2,3,4,5}{%
            \begin{subfigure}{0.23\linewidth}%
                \centering
                \includegraphics[width=\linewidth]{pic/appendix/cdf/cv_nlp_mixer_\i.pdf}\\
                \labelsize \xpad Eigenvalue
            \end{subfigure}%
        }
        \caption{MLP-Mixer}
    \end{subfigure}
    \caption{CIFAR100-SST2 Cross-Modal Tasks}\label{fig:more_existence_cross_modal}
\end{figure}

\FloatBarrier

%% file: proofs/appendix.tex
\input{proofs/implications.tex}
\input{proofs/why.tex}

%% file: proofs/implications.tex
\section{Theoretical Connection between Alignment and Forgetting}

\begin{proposition}\label{proposition:forgetting_decomposition}
    Assume $\OldLoss$ is $2$-times continuously differentiable \wrt{} $\param$.
    Assume $\OldParam$ is a local minimum of the old task loss $\OldLoss$ and $\NewParam$ is the weight of the new-task training. Define $\Delta \param \defeq \NewParam - \OldParam$ to be the new-task update.
    Then we have
    \begin{align}
        \OldLoss(\NewParam) - \OldLoss(\OldParam)
        =&  \frac{1}{2} \cdot\alignment(\oldhessian, \Delta \param) \cdot \norm{\Delta\param}_2^2 \cdot \ex[\vecperturbation \sim \mathcal{N}(\vec{0}, \vec{I} / \dim \param)]{\vecperturbation^\top \oldhessian \vecperturbation} + o(\norm{\Delta \param}_2^2).
    \end{align}
\end{proposition}
\begin{proof}
    By Taylor expansion, we have
    \begin{align}
        \OldLoss(\NewParam) - \OldLoss(\OldParam)
        =& \inner{\nabla_{\param} \OldLoss(\OldParam)}{\Delta \param} + \frac{1}{2} \cdot \Delta \param^\top \oldhessian \Delta \param + o(\norm{\Delta \param}_2^2).
    \end{align}
    Since $\OldParam$ is a local minimum, we have $\nabla_{\param} \OldLoss(\OldParam) = \vec{0}$. As a result, the first term vanishes and all forgetting is due to the second-order and high-order terms.

    Now we decompose the second-order forgetting by
    \begin{align}
        \Delta \param^\top \oldhessian \Delta \param
        =& \dim \param \cdot \frac{\Delta \param^\top \oldhessian \Delta \param}{\trace{\oldhessian} \cdot \norm{\Delta \param}_2^2} \cdot \norm{\Delta \param}_2^2 \cdot \frac{\trace{\oldhessian}}{\dim \param}\\
        =& \alignment(\oldhessian, \Delta \param) \cdot \norm{\Delta \param}_2^2 \cdot \trace{\oldhessian \times \frac{1}{\dim \param} \ex[\vecperturbation \sim \mathcal{N}(\vec{0}, \mat{I})]{\vecperturbation \vecperturbation^\top}}\\
        =& \alignment(\oldhessian, \Delta \param) \cdot \norm{\Delta \param}_2^2 \cdot \trace{\oldhessian \times \ex[\vecperturbation \sim \mathcal{N}(\vec{0}, \mat{I} / \dim \param)]{\vecperturbation \vecperturbation^\top}}\\
        =& \alignment(\oldhessian, \Delta \param) \cdot \norm{\Delta \param}_2^2 \cdot \ex[\vecperturbation \sim \mathcal{N}(\vec{0}, \mat{I} / \dim \param)]{\trace{\oldhessian \vecperturbation \vecperturbation^\top}}\\
        =& \alignment(\oldhessian, \Delta \param) \cdot \norm{\Delta \param}_2^2 \cdot \ex[\vecperturbation \sim \mathcal{N}(\vec{0}, \mat{I} / \dim \param)]{\trace{\vecperturbation^\top \oldhessian \vecperturbation}}\\
        =& \alignment(\oldhessian, \Delta \param) \cdot \norm{\Delta \param}_2^2 \cdot \ex[\vecperturbation \sim \mathcal{N}(\vec{0}, \mat{I} / \dim \param)]{\vecperturbation^\top \oldhessian \vecperturbation},
    \end{align}
    where the penultimate step uses the cyclic property of trace.
    After putting everything together, we finish the proof.
\end{proof}

%% file: proofs/why.tex
\section{Theoretical Results for Adversarial Alignment}\label{appendix:proofs}

In this section, we prove the lower-bound for the alignment between the old and new tasks.
The theoretical results arrive at the alignment between the old and new tasks from the assumed random generation of the new task, which are essentially properties of the old task.
Therefore, the proof proceeds by first reducing the inter-task alignment to the inductive bias of solely the old task.

We define consecutive weight product $\weight_{b: a} = \weight_b \weight_{b-1} \cdots \weight_{a+1} \weight_{a}$ for $b \ge a$.
We slightly abuse the notation of matrix power by forcing $\mat{A}^0 = \mat{I}$ for any real symmetric PSD matrix $\mat{A}$.
In the same spirit, when $b < a$, we let $\weight_{b:a} \defeq \mat{I}$, whose size is the same as the number of columns of $\weight_a$ so that $\weight_a \times \weight_{a-1:a}  = \weight_{a} = \weight_{a:a}$ and $\weight_{a-1:a} \times \weight_{a-1} = \weight_{a-1} = \weight_{a-1:a-1}$, whenever $\weight_a$ and $\weight_{a-1}$ has compatible shapes to multiply together.

\input{proofs/technical.tex}

\subsection{Reducing Inter-Task Alignment to Inductive Bias of the Old Task}

\input{proofs/derivative_structure.tex}

\subsection{Inductive Bias of $\ltwo$-Regularized DLNs in the Old Task}

\input{proofs/inductive_bias.tex}

\subsection{Lowerbound of Alignment between the Old and New Tasks}

\input{proofs/lowerbounding.tex}

\subsection{Extending the Lowerbound to the More General Settings}\label{appendix:relexation}

\input{proofs/extension.tex}

\section{Potential Implicit Power Iteration in the New-Task Training}\label{appendix:power_iteration}

We believe later (but initial) steps of new-task training can be modeled by power iteration: Let $\nabla \emploss_2(\NewParam^{(i-1)})$ be the gradient at the $i$-th new-task step. Then by Taylor expansion of gradients, we have $\nabla \emploss_2(\NewParam^{(2-1)}) \approx \newhessian \times \eta \nabla \emploss_2(\NewParam^{(1-1)}) + \nabla \emploss_2(\NewParam^{(1-1)})$ and similarly $\Delta \param^{(i-1)} \approx \sum_{k=0}^{i-1} \eta^k \newhessian^{k} \nabla \emploss_2(\NewParam^{(1-1)})$ until the new-task update is too far for Taylor approximation. 
Power iteration $\mat{A}^k \vec{v}$ is widely used for computing top eigenvectors, \ie{} aligning $\vec{v}$ to the top eigenvectors of $\mat{A}$.
As a result, latter steps will be more aligned with the \LargeCurvature{} directions of $\newhessian$. Following a similar argument as in \cref{sec:interpretation}, we observe that the new- and old-task \LargeCurvature{} directions will be confined to the same low-dimensional subspaces by the low-rank Jacobians and therefore are similar. Therefore, latter steps will also be more aligned with the old-task \LargeCurvature{} directions, leading to more severe forgetting. We note that a similar power iteration underlies multi-step adversarial samples \citep{implicit_power_iteration}, so that the attacking updates added to the inputs align with losses' \LargeCurvature{} directions \wrt{} inputs and drastically degrade the performance. Therefore, we emphasize by discovering the adversarial nature of the catastrophic forgetting, results in adversarial samples may be potentially transferred to \CL{} research.

%% file: proofs/technical.tex
\subsection{Technical Lemmas}\label{appendix:technical_lemmas}

We will frequently use the following well-known properties of the trace operator:
\begin{itemize}
    \item Cyclic property: $\trace{\mA \mB \mat{C}} = \trace{\mB \mat{C} \mA} = \trace{\mat{C} \mA \mB}$;
    \item Connection with Frobenius norm: $\Fnorm{\mA}^2 = \trace{\mA^\top \mA} = \trace{\mA \mA^\top}$;
\end{itemize}
Additionally, we recall von Neumann's trace inequality:
\begin{lemma}[von Neumann's trace inequality \citep{majorization}]
    Let $\mA, \mB$ be two square matrices of the same size. Then
    \begin{align}
        \trace{\mA \mB} \le \abs{\trace{\mA \mB}} \le \sum_i \sigma_i(\mA) \cdot \sigma_i(\mB) \le \sigma_1(\mA) \cdot \sum_i \sigma_i(\mB),
    \end{align}
    where $\sigma_i(\cdot)$ denotes the $i$-th largest singular value of a matrix.
\end{lemma}

We then prove several technical lemmas regarding the moments of random matrices.

\begin{lemma}\label{lemma:second_matrix_moment}
    Let $\mat{A}$ be a real symmetric matrix and $\mat{R}$ be a random matrix with compatible shape that satisfies the following property: for each column $i$ and another column $j \neq i$, one has $\distribution_{\mat{R}_{\cdot, i}} = \distribution_{\mat{R}_{\cdot, j}} = \distribution_{\vec{r}}$ and $\ex{\mat{R}_{\cdot, j} \mid \mat{R}_{\cdot, i}} = 0$.
    Then we have
    \begin{align}
        \ex[\mat{R}]{\mat{R}^\top \mat{A} \mat{R}}
        =& \trace{\mA \times \variance{\vec{r}}} \cdot \mI,
    \end{align}
    where $\variance{\cdot}$ denotes covariance.
\end{lemma}
\begin{proof}
    For any row index $i$ and column index $j$, we have
    \begin{align}
        \ex[\mat{R}]{\mat{R}^\top \mat{A} \mat{R}}_{i, j}
        =& \ex[\mat{R}]{ \mat{R}_{\cdot, i}^\top \mA \mat{R}_{\cdot, j}}.
    \end{align}
    When $i \neq j$, by applying the conditional centeredness assumption, we have
    \begin{align}
        \ex[\mat{R}]{\mat{R}^\top \mat{A} \mat{R}}_{i, j}
        =&  \ex[R_{\cdot, i}]{\mat{R}_{\cdot, i}^\top \mA \times \ex{\mat{R}_{\cdot, j} \mid \mat{R}_{\cdot, i}}}  = 0.
    \end{align} 
    For $i = j$, we have
    \begin{align}
        \ex[\mat{R}]{\mat{R}^\top \mat{A} \mat{R}}_{i, i}
        =& \ex[\mat{R}_{\cdot, i}]{R_{\cdot, i}^\top \mA \mat{R}_{\cdot, i}} 
        = \ex[\vec{r}]{\trace{\vec{r}^\top \mA \vec{r}}} \\
        =& \ex[\vec{r}]{\trace{\mA \vec{r} \vec{r}^\top}}
        = \trace{\mA \times \ex[\vec{r}]{\vec{r} \vec{r}^\top}}.
    \end{align}
    The conditional centeredness assumption implies that $\ex[\vec{r}]{\vec{r}} = 0$ and thus $\ex[\vec{r}]{\vec{r} \vec{r}^\top} = \variance{\vec{r}}$.
    As a result, the lemma is proved. 
\end{proof}

\begin{corollary}\label{corollary:second_matrix_moment_concrete}
    Some sufficient conditions for \cref{lemma:second_matrix_moment} include:
    \begin{itemize}
        \item Entries in $\mat{R}$ are mutually independent, identically distributed (I.I.D.) and centered. In this case, we have $\ex[\mat{R}]{\mat{R}^\top \mat{A} \mat{R}} = \trace{\mA} \cdot \sigma^2 \mat{I}$, where $\sigma^2$ is the variance of each entry in $\mat{R}$.
        \item $\mat{R}$ is uniformly random orthogonal matrix, \ie sampled from the Haar measure on the orthogonal group $O(d)$. In this case, we have $\ex[\mat{R}]{\mat{R}^\top \mat{A} \mat{R}} = \frac{\trace{\mat{A}}}{d} \mat{I}$.
    \end{itemize}
\end{corollary}
\begin{proof}
    The sufficiency of the first condition is straightforward. In this case, we have $\variance{\vec{r}} = \sigma^2 \cdot \mI$ and $\ex[\mat{R}]{\mat{R}^\top \mat{A} \mat{R}}
        = \trace{\mA \times \variance{\vec{r}}} \cdot \mI = \sigma^2 \trace{\mA \times \mI} \cdot \mI = \trace{\mA} \cdot \sigma^2 \mI$.

    For the second condition, since the Haar measure on the orthogonal group satisfies that for any orthogonal matrix $\mat{U}$ in the group, we have $\distribution_{\mat{R} \times \mat{U}} = \distribution_{\mat{R}}$. 
    By selecting $\mat{U}$ to the permutation matrix that swaps the $i$-th and $j$-th columns, we have $\distribution_{\mat{R}_{\cdot, i}} = \distribution_{\mat{R}_{\cdot, j}}$. 
    By selecting $\mat{U}$ to be the diagonal matrix whose $j$-th diagonal entry is $-1$ and all other entries are $1$, we have $\distribution_{\mat{R}_{\cdot, i}, -\mat{R}_{\cdot, j}} = \distribution_{\mat{R}_{\cdot, i}, \mat{R}_{\cdot, j}}$. This equality between the joint distribution implies that between the conditional distribution: given any $\mat{R}_{\cdot, i}$, we have $\distribution_{-\mat{R}_{\cdot, j} \mid \mat{R}_{\cdot, i}} = \distribution_{\mat{R}_{\cdot, j} \mid \mat{R}_{\cdot, i}}$. With such condition symmetry, the conditional centeredness is satisfied. The second moment follows from the well-known fact that $\ex[\vec{r}]{\vec{r} \vec{r}^\top} = \frac{1}{d} \mat{I}$ for the uniformly distributed unit vector $\vec{r}$.
\end{proof}

\begin{lemma}\label{lemma:untransposed_second_matrix_moment}
    Let $\RandRot$ be a random orthogonal matrix sampled from the Haar measure on the orthogonal group $O(d)$ and let $\mB$ be a real matrix. Then $\ex[\RandRot]{\RandRot \mB \RandRot} = \frac{\mB^\top}{d}$.
\end{lemma}
\begin{proof}
    We compute $\ex[\RandRot]{\RandRot \mB \RandRot}$ entry by entry. Let $\vec{e}_i$ be the $i$-th standard basis vector in $\reals^d$. Then we have
    \begin{align}
        \ex[\RandRot]{\RandRot \mB \RandRot}_{i, j}
        =& \ex{\trace{\ve_i^\top \RandRot \mB \RandRot \ve_j}}
        =  \ex{\trace{\mB \RandRot \ve_j \ve_i^\top \RandRot}}\\
        =& \trace{\mB \times \ex{\RandRot_{\cdot, j} \times \RandRot_{i, \cdot}}}
        =  \trace{\mB \times \begin{bmatrix}\ex{u_{p, j} \cdot u_{i, q}}\end{bmatrix}_{p, q}} 
    \end{align}
    When $p \neq i$, we have $\ex{\RandRot_{p, \cdot} \mid \RandRot_{i, \cdot}} = \vec{0}$, $\ex{u_{p, j} \mid \RandRot_{i, \cdot}} = 0$ and finally $\ex{u_{p, j} \mid u_{i, q}} = 0$. Therefore, we have $\ex{u_{p, j} \cdot u_{i, q}} = 0$ when $p \neq i$. A similar argument shows when $q \neq j$, we have $\ex{u_{p, j} \cdot u_{i, q}} = 0$.
    
    The only non-zero entry in $\begin{bmatrix}\ex{u_{p, j} \cdot u_{i, q}}\end{bmatrix}_{p, q}$ is when $p = i$ and $q = j$, which is $\ex{u_{i, j} \cdot u_{i, j}} = \frac{1}{d}$. Therefore, we have 
    \begin{align}
        \ex[\RandRot]{\RandRot \mB \RandRot}_{i, j}
        =&  \trace{\mB \times \frac{\ve_i \ve_j^\top}{d}}
        =   \frac{1}{d} \inner{\mB}{\ve_j \ve_i^\top}
        =   \frac{B_{j, i}}{d},
    \end{align}
    which implies 
    \begin{align}
        \ex[\RandRot]{\RandRot \mB \RandRot} = \frac{\mB^\top}{d}.
    \end{align}
\end{proof}

We now prove several lemmas regarding partition-then-norm structure that appears frequently in later proofs.

\begin{lemma}\label{lemma:partitioned_norm}
    Assume $\CommonSV$ is a non-zero diagonal matrix with non-negative entries. Let $A < B$ be positive integers and let $f(x) \defeq \Fnorm{\CommonSV^{B-x}}^2 \cdot \Fnorm{\CommonSV^{x-A}}^2$ for $x \in [A, B]$. Then $f$ is convex and symmetric about $x_0 = \frac{A+B}{2}$.
    As a result, $f$ takes minimum at $x_0$ and maximum at $x = A$ or $x = B$, and $f$ is monotonic in $[A, x_0]$ and $[x_0, B]$.
\end{lemma}
\begin{proof}
    The symmetry is straightforward from the definition of $f$.

    Now we prove the convexity of $f$.
    First assume all entries in $\CommonSV$ are non-zero.
    Let $\sigma_i$ be the $i$-th diagonal entry of $\CommonSV$.
    By definition of $f$, we have
    \begin{align}
        f(x) 
        =& \sum_{i} \sigma_i^{2(B-x)} \sum_j \sigma_j^{2(x-A)}\\
        =& \sum_{i} \sigma_i^{2(A-B)} + 2 \sum_{i < j} \left(\frac{\sigma_i^{2 B}}{\sigma_j^{2 A}} \cdot \left(\frac{\sigma_j}{\sigma_i}\right)^{2 x} + \frac{\sigma_j^{2 B}}{\sigma_i^{2 A}} \cdot \left(\frac{\sigma_i}{\sigma_j}\right)^{2 x}\right).
    \end{align}
    Since $\left(\frac{\sigma_j}{\sigma_i}\right)^{2 x}$ and $\left(\frac{\sigma_i}{\sigma_j}\right)^{2 x}$ are both convex functions of $x$, and all other coefficients or constants are non-negative, we have $f(x)$ is convex.
    The rest of claims follows from the symmetry and convexity of $f$.

    \newcommand{\size}[1]{\left|#1\right|}
    \newcommand{\zeroset}{\mathcal{Z}}
    \newcommand{\indicator}[1]{\mathbb{I}\left[#1\right]}
    If some entries in $\CommonSV$ are zero, we have a slightly more complicated expression:
    \begin{align}
        f(x) 
        =& \left(\sum_{i: \sigma_i > 0} \sigma_i^{2(B-x)} + \sum_{i: \sigma_i = 0} \indicator{x = B}\right) \left(\sum_{j: \sigma_j > 0} \sigma_j^{2(x-A)} + \sum_{j: \sigma_j = 0} \indicator{x = A}\right)\\
        =& \left(\sum_{i: \sigma_i > 0} \sigma_i^{2(B-x)} + \size{\zeroset} \cdot [x = B]\right) \left(\sum_{j: \sigma_j > 0} \sigma_j^{2(x-A)} + \size{\zeroset} \cdot \indicator{x = A}\right)\\
        =& \sum_{i: \sigma_i > 0} \sigma_i^{2(A-B)} + 2 \sum_{i < j: \sigma_i > 0, \sigma_j > 0} \left(\frac{\sigma_i^{2 B}}{\sigma_j^{2 A}} \cdot \left(\frac{\sigma_j}{\sigma_i}\right)^{2 x} + \frac{\sigma_j^{2 B}}{\sigma_i^{2 A}} \cdot \left(\frac{\sigma_i}{\sigma_j}\right)^{2 x}\right)\\
            &
            +   \sum_{i: \sigma_i > 0} \sigma_i^{2(B-x)} \cdot \size{\zeroset} \cdot \indicator{x = A}
            +   \sum_{j: \sigma_j > 0} \sigma_j^{2(x-A)} \cdot \size{\zeroset} \cdot \indicator{x = B}\\
            &
            +   \size{\zeroset}^2 \cdot \indicator{x = A} \cdot \indicator{x = B}\\
        =& \sum_{i: \sigma_i > 0} \sigma_i^{2(A-B)} + 2 \sum_{i < j: \sigma_i > 0, \sigma_j > 0} \left(\frac{\sigma_i^{2 B}}{\sigma_j^{2 A}} \cdot \left(\frac{\sigma_j}{\sigma_i}\right)^{2 x} + \frac{\sigma_j^{2 B}}{\sigma_i^{2 A}} \cdot \left(\frac{\sigma_i}{\sigma_j}\right)^{2 x}\right)\\
            &
            +   \sum_{i: \sigma_i > 0} \sigma_i^{2(B-A)} \cdot \size{\zeroset} \cdot \indicator{x = A}
            +   \sum_{j: \sigma_j > 0} \sigma_j^{2(B-A)} \cdot \size{\zeroset} \cdot \indicator{x = B}\\
            &
            +   \size{\zeroset}^2 \cdot \indicator{x = A \land x = B}\\
        =& \sum_{i: \sigma_i > 0} \sigma_i^{2(A-B)} + 2 \sum_{i < j: \sigma_i > 0, \sigma_j > 0} \left(\frac{\sigma_i^{2 B}}{\sigma_j^{2 A}} \cdot \left(\frac{\sigma_j}{\sigma_i}\right)^{2 x} + \frac{\sigma_j^{2 B}}{\sigma_i^{2 A}} \cdot \left(\frac{\sigma_i}{\sigma_j}\right)^{2 x}\right)\\
            &
            +   \sum_{i: \sigma_i > 0} \sigma_i^{2(B-A)} \cdot \size{\zeroset} \cdot \indicator{x = A}
            +   \sum_{j: \sigma_j > 0} \sigma_j^{2(B-A)} \cdot \size{\zeroset} \cdot \indicator{x = B}.
    \end{align}
    where $\zeroset = \set{i: \sigma_i = 0}$ and the last step is because $A \neq B$.
    Since $\indicator{x = A}$ and $\indicator{x = B}$ are both convex functions over $[A, B]$, and their coefficients are non-negative, the new terms are also convex.
    Combined with convexity of the previous terms, we have $f$ is convex.
\end{proof}

\begin{lemma}\label{lemma:imbalance_fraction_to_erank}
    Let $\CommonSV$ be a non-zero diagonal matrix with non-negative entries. Let $A, B \ge 0$. Then we have
    \begin{align}
        \erank{\CommonSV^{2 \max(A, B)}} \le \frac{\Fnorm{\CommonSV^{A}}^2 \cdot \Fnorm{\CommonSV^{B}}^2}{\Fnorm{\CommonSV^{A + B}}^2} \le \erank{\CommonSV^{2 \min(A, B)}}.
    \end{align}
\end{lemma}
\begin{proof}
    To prove the first inequality, it is equivalent to show
    \begin{align}
        \erank{\CommonSV^{2 \max(A, B)}} 
        \defeq& \frac{\trace{\CommonSV^{2\max(A, B)}}^2}{\trace{\left(\CommonSV^{2\max(A, B)}\right)^2}}
        = \frac{\Fnorm{\CommonSV^{\max(A, B)}}^2 \cdot \Fnorm{\CommonSV^{\max(A, B)}}^2}{\Fnorm{\CommonSV^{2\max(A, B)}}^2}\\
        \le& \frac{\Fnorm{\CommonSV^{\min(A, B)}}^2 \cdot \Fnorm{\CommonSV^{\max(A, B)}}^2}{\Fnorm{\CommonSV^{\min(A, B) + \max(A, B)}}^2}.
    \end{align}
    Then it is equivalent to show
    \begin{align}
        \Fnorm{\CommonSV^{\max(A, B)}}^2 \cdot \Fnorm{\CommonSV^{\min(A, B) + \max(A, B)}}^2
        \le& \Fnorm{\CommonSV^{\min(A, B)}}^2 \cdot \Fnorm{\CommonSV^{2\max(A, B)}}^2
    \end{align}
    Note that the exponents of the both sides have the same sum, \ie{} $\max(A, B) + (\min(A, B) + \max(A, B)) = \min(A, B) + 2\max(A, B)$.
    Recall the function $f$ in \cref{lemma:partitioned_norm}. Then the two sides are two values of $f$ in the interval $[0, \min(A, B) + 2\max(A, B)]$. According to $\cref{lemma:partitioned_norm}$, the one closer to the middle $\frac{\min(A, B) + 2\max(A, B)}{2} = \max(A, B) + \frac{\min(A, B)}{2}$ is smaller.
    The left-hand side's distance to the middle is $\frac{\min(A, B)}{2}$ while the right-hand side's is $\max(A, B) - \frac{\min(A, B)}{2}$. Since $\frac{\min(A, B)}{2} \le \max(A, B) - \frac{\min(A, B)}{2}$ by definitions of $\min$ and $\max$, the left-hand side is smaller than the right-hand side, which proves the first inequality.

    The proof to the second inequality is similar and is omitted.
\end{proof}

We then prove lemmas that bound the traces of various matrix products:
\begin{lemma}\label{lemma:trace_bounds}
    Let $\mM, \mA$ be real symmetric PSD matrices such that the null spaces of $\mA$ are superset of $\mM$ null space. Under such condition, we have
    \begin{align}
        \sigma_{\min}(\mM) \cdot \trace{\mA}
        \le \trace{\mA \mM}
        \le \sigma_1(\mM) \cdot \trace{\mA},
    \end{align}
    where $\sigma_{\min}(\cdot)$ denotes the least non-zero singular value and $\sigma_1(\cdot)$ denotes the largest singular value.

    A corollary is that for any real matrices $\mB, \mat{N}$, such that $\mB$'s right nullspace is the superset of $\mat{N}$'s left nullspace, we have
    \begin{align}
        \sigma_{\min}^2 (\mM)\cdot \Fnorm{\mB}^2 \le  \Fnorm{\mB \mM}^2 \le \sigma_1^2(\mM) \cdot \Fnorm{\mB}^2.
    \end{align}
\end{lemma}
\begin{proof}
    \newcommand{\nullstart}{i_{\text{null}}}
    \newcommand{\mLambda}{\mat{\Lambda}}
    Let $\mM = \mV_{\mM} \mLambda_{\mM} \mV_{\mM}^\top$ be the eigenvalue decomposition of $\mM$. Let $\nullstart$ be the start of the nullspace of $\mM$. Therefore, for any $i \ge \nullstart$, we have $\lambda_{\mM, i} = 0$. Moreover, by assumption that $\mA$'s null space is superset of $\mM$'s, we have $\mV_{\mM}^\top \mA \mV_{\mM} = \begin{bmatrix}
        \tilde{\mA} & \mat{0}\\
        \mat{0} & \mat{0}
    \end{bmatrix}$, where $\tilde{\mA} \in \reals^{(\nullstart - 1) \times (\nullstart - 1)}$.
    Therefore, we have
    \begin{align}
        \trace{\mA \mM} 
        =& \trace{\mLambda_{\mM}  \mV_{\mM}^\top \mA \mV_{\mM}}
        = \trace{(\mLambda_{\mM})_{1:\nullstart-1, 1:\nullstart-1} \tilde{\mA}}.
    \end{align}
    Since $\mA$ is symmetric and PSD, so is $\mV_{\mM}^\top \mA \mV_{\mM}$ and $\tilde{\mA}$, whose diagonal entries are non-negative. Therefore, we have
    \begin{align}
        \trace{\mA \mM}  
        \in& [\sigma_{\min}(\mM) \cdot \trace{\tilde{\mA}}, \sigma_{1}(\mM) \cdot \trace{\tilde{\mA}}]\\
        =& [\sigma_{\min}(\mM) \cdot \trace{\mV_{\mM}^\top \mA \mV_{\mM}}, \sigma_{1}(\mM) \cdot \trace{\mV_{\mM}^\top \mA \mV_{\mM}}]\\
        =& [\sigma_{\min}(\mM) \cdot \trace{\mA}, \sigma_{1}(\mM) \cdot \trace{\mA}].
    \end{align}
    The corollary is proved by applying the above result to $\Fnorm{\mB \mat{N}}^2 = \trace{\mB \mat{N} (\mB \mat{N})^\top} = \trace{\mB \mB^\top \mat{N} \mat{N}^\top}$.
\end{proof}

\begin{lemma}\label{lemma:trace_bounds_four_components}
    Let $\CommonSV$ be a diagonal matrix with non-negative entries. Let $\mA, \mB$ be two real matrices. Let $l, r > 0$. Then we have
    \begin{align}
        \trace{\mA \CommonSV^{l} \mB \CommonSV^{r}}
        \le& \sigma_1(\mA) \cdot \sigma_1(\mB) \cdot \trace{\CommonSV^{l + r}}.
    \end{align} 
\end{lemma}
\begin{proof}
    \newcommand{\vu}{\vec{u}}
    \newcommand{\vv}{\vec{v}}
    \newcommand{\mQ}{\mat{Q}}
    By von Neumann's trace inequality, we have
    \begin{align}
        &\trace{\mA \CommonSV^{l} \mB \CommonSV^{r}}\\
        \le& \sum_i \sigma_i(\mA) \cdot \sigma_i(\CommonSV^l \mB \CommonSV^r)
        \le  \sigma_1(\mA) \sum_i \sigma_i(\CommonSV^l \tilde{\mB} \CommonSV^r)\\
        \le& \sigma_1(\mA) \cdot  \nuclearnorm{\CommonSV^l \mB \CommonSV^r}
        ,
    \end{align}
    where $\nuclearnorm{\mat{M}} \defeq \sum_k \sigma_k(\mat{M})$ is the nuclear norm.
    By the dual characterization of nuclear norm
    \begin{align}
        \nuclearnorm{\mat{M}} = \sup_{\mQ: \snorm{\mQ} \le 1} \inner{\mQ}{\mat{M}} = \sup_{\mQ: \snorm{\mQ} \le 1} \trace{\mQ^\top \mat{M}},
    \end{align}
    we have
    \begin{align}
        \trace{\mA \CommonSV^{l} \mB \CommonSV^{r}}
        \le& \sigma_1(\mA) \cdot \sup_{\mQ: \snorm{\mQ}\le 1} \trace{\mQ^\top \CommonSV^l \mB \CommonSV^r}\\
        \le& \sigma_1(\mA) \cdot \sup_{\mQ: \snorm{\mQ}\le 1} \trace{\mB \CommonSV^r \mQ^\top \CommonSV^l}\\
        \le& \sigma_1(\mA) \cdot \sup_{\mQ: \snorm{\mQ}\le 1} \sum_i \sigma_i(\mB) \cdot \sigma_i(\CommonSV^r \mQ^\top \CommonSV^l)\\
        \le& \sigma_1(\mA) \cdot \sigma_1(\mB) \cdot \sup_{\mQ: \snorm{\mQ}\le 1} \sum_i \sigma_i(\CommonSV^r \mQ^\top \CommonSV^l).
    \end{align}
    From the proof by \citet[Page 342]{majorization}, we have that $\sum_{i} \sigma_i(\mat{M} \mat{N}) \le \sum_i \sigma_i(\mat{M}) \cdot \sigma_i(\mat{N})$, which implies
    \begin{align}
        \trace{\mA \CommonSV^{l} \mB \CommonSV^{r}}
        \le& \sigma_1(\mA) \cdot \sigma_1(\mB) \cdot \sup_{\mQ: \snorm{\mQ}\le 1} \sum_i \sigma_i(\CommonSV^r) \cdot \sigma_i (\mQ^\top \CommonSV^l).
    \end{align}
    By the well-known result that $\sigma_i(\mat{M} \mat{N}) \le \sigma_1(\mat{M}) \cdot \sigma_i(\mat{N})$, we have
    \begin{align}
        \trace{\mA \CommonSV^{l} \mB \CommonSV^{r}}
        \le& \sigma_1(\mA) \cdot \sigma_1(\mB) \cdot \sup_{\mQ: \snorm{\mQ}\le 1} \sum_i \sigma_i(\CommonSV^r) \cdot \sigma_1 (\mQ^\top) \cdot \sigma_i(\CommonSV^l)\\
        \le& \sigma_1(\mA) \cdot \sigma_1(\mB) \cdot \sup_{\mQ: \snorm{\mQ}\le 1} \sum_i \sigma_i(\CommonSV^r) \cdot 1 \cdot \sigma_i(\CommonSV^l)\\
        =& \sigma_1(\mA) \cdot \sigma_1(\mB) \cdot \trace{\CommonSV^l \CommonSV^r}.
    \end{align}
\end{proof}

Finally, we complete the proof of \cref{prop:kronecker_erank} claimed in \cref{sec:definition_erank}.

\begin{proof}[Proof of \cref{prop:kronecker_erank}]
    For the effective rank of Kronecker products, we have
    \begin{align}
         \erank{\mA \kronecker \mB} 
        =& \frac{\trace{\mA \kronecker \mB}^2}{\trace{(\mA \kronecker \mB) (\mA \kronecker \mB)}} \\
        &= \frac{\trace{\mA \kronecker \mB}^2}{\trace{\mA^2 \kronecker \mB^2}} & ((\mat{M} \kronecker \mat{N})(\mat{P} \kronecker \mat{Q}) = (\mat{M} \mat{P}) \kronecker (\mat{N} \mat{Q})) \\
        =& \frac{\trace{\mA}^2 \cdot \trace{\mB}^2}{\trace{\mA^2} \cdot \trace{\mB^2}} & (\trace{\mat{M} \kronecker \mat{N}} = \trace{\mat{M}} \cdot \trace{\mat{N}})\\ 
        =& \erank{\mA} \cdot \erank{\mB}.
    \end{align}

    For the effective rank of block-diagonal matrices, we have 
    \begin{align}
        \erank{\Diag(\mA_i)}
        =& \frac{\left(\sum_i \trace{\mA_i}\right)^2}{\sum_i \trace{\mA_i^2}}\\
        =& \frac{\left(\sum_i \sqrt{\erank{\mA_i}} \cdot \sqrt{\trace{\mA_i^2}}\right)^2}{\sum_i \trace{\mA_i^2}}\\
        =& \left(\frac{\inner{\begin{bmatrix}
            \sqrt{\erank{\mA_i}}
        \end{bmatrix}_i}{\begin{bmatrix}
            \sqrt{\trace{\mA_i^2}}
        \end{bmatrix}_i}}{\norm{\begin{bmatrix}
            \sqrt{\trace{\mA_i^2}}
        \end{bmatrix}_i}}\right)^2\\
        \le& \left(\frac{\inner{\begin{bmatrix}
            \sqrt{\erank{\mA_i}}
        \end{bmatrix}_i}{\begin{bmatrix}
            \sqrt{\erank{\mA_i}}
        \end{bmatrix}_i}}{\norm{\begin{bmatrix}
            \sqrt{\erank{\mA_i}}
        \end{bmatrix}_i}}\right)^2\\
        =& \norm{\begin{bmatrix}
            \sqrt{\erank{\mA_i}}
        \end{bmatrix}_i}^2\\
        =& \sum_i \erank{\mA_i},
    \end{align}
    where the inequality is due to Cauchy-Schwarz inequality.
\end{proof}

%% file: proofs/derivative_structure.tex
We now prove the theoretical results stated in \cref{sec:results}.
They are proved under the setting of regression using deep linear networks (DLN) of depth $L$, which is defined by $f_{\param}(\Input) \defeq \weight_{L} \weight_{L-1} \cdots \weight_{1} \Input$, where $\weight_{i} \in \reals^{\dim \Input \times \dim \Input}$.
The model is trained by the mean square error (MSE): $\emploss(\param, (\Input, \Label)) \defeq \Fnorm{f_{\param}(\Input) - \Label}^2$.
The following lemma gives the gradient of the empirical MSE loss \wrt{} the weights, which is straightforward to verify using the chain rule.

\begin{lemma}[Gradient of empirical MSE loss]\label{lemma:gradient_empirical_mse_loss}
    For any task $\task = (\EmpiricalInputs, \EmpiricalLabels)$, the gradient of the empirical mean square error (MSE) loss \wrt the weights of a deep linear network (DLN) is given by
    \begin{align}
        \frac{\partial \emploss(\param, \task)}{\partial \weight_i} = \weight_{L:i+1}^\top (\weight_{L:1} \EmpiricalInputs - \EmpiricalLabels) \EmpiricalInputs^\top \weight_{i-1:1}^\top.
    \end{align} 
\end{lemma}

Then we compute the three components in the definition of $\alignment$: the trace of Hessian, the norm of gradient, and the quadratic form of the Hessian and the gradient.

\subsubsection{Trace of Hessian}

\begin{lemma}\label{lemma:hessian_trace}
    \newcommand{\randvec}{\vec{r}}
    \newcommand{\mR}{\mat{R}}
    The the trace of Hessian $\hessian$ at the task $\task=(\EmpiricalInputs, \EmpiricalLabels)$ is given by
    \begin{align}
        \trace{\hessian} 
        =& \sum_{i=1}^{L} \ex{\Fnorm{\weight_{L:i+1} \mR_i \weight_{i-1:1} \EmpiricalInputs}^2}
        = \sum_{i=1}^L \Fnorm{\weight_{L:i+1}}^2 \cdot \Fnorm{\weight_{i-1:1} \EmpiricalInputs}^2,
    \end{align}
    where $\mR_i \in \set{-1, +1}^{m_i \times n_i}$ is a random matrix with independent and identically distributed entries sampled from $\prob{R_{i, p, q} = -1} = \prob{R_{i, p, q} = +1} = \frac{1}{2}$.
\end{lemma}
\begin{proof}
    \newcommand{\randvec}{\vec{r}}
    \newcommand{\mR}{\mat{R}}
    Let $\randvec \defeq \vectorize{\packed{\mat{R}_i}}$ for convenience.
    By construction of $\left(\mat{R}_i\right)_{i=1}^L$, we have $\ex{\randvec \randvec^\top} = \mat{I}$. As a result, we can compute the Hessian trace by
    \begin{align}
        \trace{\hessian}
        =& \trace{\hessian \times \ex{\randvec \randvec^\top}}
        = \ex{\randvec^\top \hessian \randvec}
        = \ex{\randvec^\top \frac{\partial \left(\frac{\partial \emploss}{\partial \param^\top}\right)}{\partial \param} \randvec}
        = \ex{\randvec^\top \frac{\partial \inner{\frac{\partial \emploss}{\partial \param}}{\randvec}}{\partial \param}}\\
        =&\sum_{j=1}^L \ex{\trace{\mR_j^\top \frac{\partial \inner{\frac{\partial \emploss}{\partial \param}}{\randvec}}{\partial \weight_j}}}.
    \end{align}
    The gradient-random vector inner product can be computed as
    \begin{align}
        & \inner{\frac{\partial \emploss}{\partial \param}}{\randvec}
        =  \sum_{i=1}^L \inner{\frac{\partial \emploss}{\partial \weight_i}}{\mR_i}
        =   \sum_{i=1}^L \trace{\mR_i^\top \partialfrac{\emploss}{\weight_i}}\\
        =&  \sum_{i=1}^L \left(\trace{\mR_i^\top \weight_{L:i+1}^\top \weight_{L:1} \EmpiricalInputs \EmpiricalInputs^\top \weight_{i-1:1}^\top} - \trace{\mR_i^\top \weight_{L:i+1}^\top \EmpiricalLabels \EmpiricalInputs^\top \weight_{i-1:1}^\top }\right)\\
        =& \sum_{i=1}^L \left( \trace{ \weight_{L:1} \EmpiricalInputs \EmpiricalInputs^\top \left( \weight_{L:i+1} \mR_i \weight_{i-1:1} \right)^\top } - \trace{ \mR_i^\top \weight_{L:i+1}^\top \EmpiricalLabels \EmpiricalInputs^\top \weight_{i-1:1}^\top } \right).
    \end{align}
    Now we take differential of the inner product \wrt $\weight_j$:
    \begin{align}
        &\diff \inner{\partialfrac{\emploss}{\param}}{\randvec}
        = \trace{\left(\partialfrac{\inner{\partialfrac{\emploss}{\param}}{\randvec}}{\weight_j}\right)^\top \diff \weight_j} \\
        =& \sum_{i=1}^L \left( \diff \trace{\weight_{L:1} \EmpiricalInputs \EmpiricalInputs^\top (\weight_{L:i+1} \mR_i \weight_{i-1:1})^\top} - \diff \trace{\mR_i^\top \weight_{L:i+1}^\top \EmpiricalLabels \EmpiricalInputs^\top \weight_{i-1:1}^\top} \right) \\
        =& \sum_{i=1}^L \trace{ (\diff \weight_{L:1}) \EmpiricalInputs \EmpiricalInputs^\top (\weight_{L:i+1} \mR_i \weight_{i-1:1})^\top } 
        \\
            &+ \sum_{i=1}^L \left( \trace{ \weight_{L:1} \EmpiricalInputs \EmpiricalInputs^\top \diff (\weight_{L:i+1} \mR_i \weight_{i-1:1})^\top } - \diff \trace{\mR_i^\top \weight_{L:i+1}^\top \EmpiricalLabels \EmpiricalInputs^\top \weight_{i-1:1}^\top } \right) \\
        =& \trace{ \weight_{L:j+1} \diff \weight_j \weight_{j-1:1} \EmpiricalInputs \EmpiricalInputs^\top \left( \sum_{i=1}^L \weight_{L:i+1} \mR_i \weight_{i-1:1} \right)^\top }\\
            &+ \sum_{i=1}^L \trace{(\weight_{L:1} \EmpiricalInputs - \EmpiricalLabels) \EmpiricalInputs^\top \diff (\weight_{L:i+1} \mR_i \weight_{i-1:1})^\top}. \label{eq:reusable_vector_hessian_product}
    \end{align}

    When $j = i$, the term $(\weight_{L:i+1} \mR_i \weight_{i-1:1})^\top$ does not contain $\weight_j = \weight_i$ and taking differential \wrt{} it leads to zero. 
    When $j \neq i$, $\diff (\weight_{L:i+1} \mR_i \weight_{i-1:1})^\top$ contains $\mR_i$ as a factor. Since $\mR_i$ is centered by construction, after taking expectation, the term vanishes.
    In both cases, the term has no contribution to the trace of Hessian after we take expectation. Therefore, we can ignore the second term and focus on $\trace{ \weight_{L:j+1} \diff \weight_j \weight_{j-1:1} \EmpiricalInputs \EmpiricalInputs^\top \left( \sum_{i=1}^L \weight_{L:i+1} \mR_i \weight_{i-1:1} \right)^\top }$. 

    The standard process follows extracting $\diff \weight_j$ in the trace and see the rest as $\left(\partialfrac{\inner{\partialfrac{\emploss}{\param}}{\randvec}}{\weight_j}\right)^\top$. After that, we take inner product of the gradient with $\mR_j$. This process effectively replaces $\diff \weight_j$ with $\mR_j$. Therefore, we take a shortcut where we directly compute the trace of Hessian as
    \begin{align}
        \trace{\hessian}
        =& \sum_{j=1}^L \ex{\trace{\mR_j^\top \frac{\partial \inner{\frac{\partial \emploss}{\partial \param}}{\randvec}}{\partial \weight_j}}}\\
        =&  \sum_{j=1}^L \ex{\trace{ \weight_{L:j+1} \mR_j \weight_{j-1:1} \EmpiricalInputs \EmpiricalInputs^\top \left( \sum_{i=1}^L \weight_{L:i+1} \mR_i \weight_{i-1:1} \right)^\top}}\\
        =&  \sum_{j=1}^L \sum_{i=1}^L \ex{\trace{ \weight_{L:j+1} \mR_j \weight_{j-1:1} \EmpiricalInputs \EmpiricalInputs^\top \left(  \weight_{L:i+1} \mR_i \weight_{i-1:1} \right)^\top}}\\
        =&  \sum_{i=1}^L \ex{\trace{ \weight_{L:i+1} \mR_i \weight_{i-1:1} \EmpiricalInputs \EmpiricalInputs^\top \left( \weight_{L:i+1} \mR_i \weight_{i-1:1} \right)^\top}}\\
        =&  \sum_{i=1}^L \ex{\Fnorm{\weight_{L:i+1} \mR_i \weight_{i-1:1} \EmpiricalInputs}^2},
    \end{align} 
    where the forth equality follows from the fact when $i \neq j$, $\mR_i$ and $\mR_j$ are independent and the fact that $\ex{\mR_i} = \ex{\mR_j} = \mat{0}$.
    To remove $\mR_i$, we use \cref{corollary:second_matrix_moment_concrete} and obtain
    \begin{align}
        \ex{\Fnorm{\weight_{L:i+1} \mR_i \weight_{i-1:1} \EmpiricalInputs}^2}
        =& \trace{\weight_{L:i+1} \ex{\mR_i \weight_{i-1:1} \EmpiricalInputs \EmpiricalInputs^\top \weight_{i-1:1}^\top \mR_i^\top} \weight_{L:i+1}^\top}\\
        =& \trace{\weight_{L:i+1} \left(\trace{\weight_{i-1:1} \EmpiricalInputs \EmpiricalInputs^\top \weight_{i-1:1}^\top} \cdot 1 \cdot \mat{I}\right)  \weight_{L:i+1}^\top}\\
        =& \trace{\weight_{L:i+1} \weight_{L:i+1}^\top} \cdot \trace{\weight_{i-1:1} \EmpiricalInputs \EmpiricalInputs^\top \weight_{i-1:1}^\top}\\
        =& \Fnorm{\weight_{L:i+1}}^2 \cdot \Fnorm{\weight_{i-1:1} \EmpiricalInputs}^2.
    \end{align}
    Plugging this result back, the lemma is proved.
\end{proof}

\subsubsection{Norm of Gradient}

\begin{LocalAssumption}[Symmetric distribution of inputs.]\label{assumption:symmetric_inputs}
    A random $\task = (\EmpiricalInputs, \EmpiricalLabels) \sim \TaskDistribution$ has symmetric inputs if for any supported $\EmpiricalLabels$, we have $\distribution_{-\EmpiricalInputs \mid \EmpiricalLabels} = \distribution_{\EmpiricalInputs \mid \EmpiricalLabels}$.

\end{LocalAssumption}

\begin{lemma}\label{lemma:new_gradient_norm}
    Assume a random task $\task = (\EmpiricalInputs, \EmpiricalLabels) \sim \TaskDistribution$ satisfies \cref{assumption:symmetric_inputs}. 
    Then the expected squared norm of the gradient of the empirical MSE loss \wrt the weights is given by
    \begin{align}
        \ex[\task \sim \TaskDistribution]{\norm{\partialfrac{\emploss(\param, \task)}{\param}}^2} = \sum_{i=1}^L \ex{
                \Fnorm{\weight_{L:i+1}^\top \weight_{L:1} \EmpiricalInputs \EmpiricalInputs^\top \weight_{i-1:1}^\top}^2
                +\Fnorm{\weight_{L:i+1}^\top \EmpiricalLabels \EmpiricalInputs^\top \weight_{i-1:1}^\top}^2 
        }.
    \end{align}
\end{lemma}

\begin{proof}
    By \cref{lemma:gradient_empirical_mse_loss}, we can compute the expected gradient norm as
    \begin{align}
        &\ex[\task \sim \TaskDistribution]{\norm{\partialfrac{\emploss(\param, \task)}{\param}}^2}
        =  \sum_{i=1}^L \ex{\Fnorm{\partialfrac{\emploss(\param, \task)}{\weight_i}}^2}\\
        =& \sum_{i=1}^L \ex{\Fnorm{\weight_{L:i+1}^\top (\weight_{L:1} \EmpiricalInputs - \EmpiricalLabels) \EmpiricalInputs^\top \weight_{i-1:1}^\top}^2}\\
        =& \sum_{i=1}^L \ex{
                \Fnorm{\weight_{L:i+1}^\top \weight_{L:1} \EmpiricalInputs \EmpiricalInputs^\top \weight_{i-1:1}^\top}^2
                +\Fnorm{\weight_{L:i+1}^\top \EmpiricalLabels \EmpiricalInputs^\top \weight_{i-1:1}^\top}^2 
            }\\
            &- 2 \cdot \ex{\trace{\weight_{L:i+1}^\top \weight_{L:1} \EmpiricalInputs \EmpiricalInputs^\top \weight_{i-1:1}^\top \left(\weight_{L:i+1}^\top \EmpiricalLabels \EmpiricalInputs^\top \weight_{i-1:1}^\top\right)^\top}}\\
        =& \sum_{i=1}^L \ex{
                \Fnorm{\weight_{L:i+1}^\top \weight_{L:1} \EmpiricalInputs \EmpiricalInputs^\top \weight_{i-1:1}^\top}^2
                +\Fnorm{\weight_{L:i+1}^\top \EmpiricalLabels \EmpiricalInputs^\top \weight_{i-1:1}^\top}^2 
            }\\
            &- 2 \cdot \trace{\weight_{L:i+1}^\top \weight_{L:1} \ex[\EmpiricalLabels]{\ex[\EmpiricalInputs \mid \EmpiricalLabels]{\EmpiricalInputs \EmpiricalInputs^\top \weight_{i-1:1}^\top \weight_{i-1:1} \EmpiricalInputs} \EmpiricalLabels^\top} \weight_{L:i+1}}\\
    \end{align}
    By \cref{assumption:symmetric_inputs}, the conditional distribution of $\EmpiricalInputs$ given $\EmpiricalLabels$ is symmetric, the third moment $\ex[\EmpiricalInputs \mid \EmpiricalInputs]{\EmpiricalInputs \EmpiricalInputs^\top \weight_{i-1:1}^\top \weight_{i-1:1} \EmpiricalInputs} = 0$. Therefore, we have
    \begin{align}
        &\ex{\norm{\partialfrac{\emploss(\param, \task)}{\param}}^2}
        =& \sum_{i=1}^L \ex{
                \Fnorm{\weight_{L:i+1}^\top \weight_{L:1} \EmpiricalInputs \EmpiricalInputs^\top \weight_{i-1:1}^\top}^2
                +\Fnorm{\weight_{L:i+1}^\top \EmpiricalLabels \EmpiricalInputs^\top \weight_{i-1:1}^\top}^2 
        },
    \end{align}
    and the lemma is proved.
\end{proof}

\subsubsection{The Product between the Gradient and the Hessian}

\begin{lemma}\label{lemma:gradient_hessian_product}
    Assume the old task $\oldtask = (\EmpiricalInputs_1, \EmpiricalLabels_1)$ is fixed.
    For convenience, define $\hessian_1 \defeq \frac{\partial^2 \emploss(\param, \oldtask)}{\partial \param \partial \param^\top}$ as the Hessian of the empirical MSE loss \wrt the weights at the old task $\task_1$.

    Assume the random new task $\newtask = (\EmpiricalInputs_2, \EmpiricalLabels_2) \sim \TaskDistribution[2]$ satisfies \cref{assumption:symmetric_inputs}. 
    For convenience, define $\vg \defeq \partialfrac{\emploss(\param, \newtask)}{\param}$ as the gradient of the empirical MSE loss \wrt the weights at the new task $\newtask$. Let $\mG_i \defeq \partialfrac{\emploss(\param, \newtask)}{\weight_i}$ be the matrix-structured version of $\vg$. Be noted that $\vg$ and $\set{\mG_i}_{i=1}^L$ are all random vectors/matrices while $\oldhessian$ is not.

    Then we have 
    \begin{align}
        \ex[\newtask]{\vg^\top \oldhessian \vg}
        =&
            \ex{\Fnorm{ \sum_{i=1}^L \weight_{L:i+1} \left(\weight_{L:i+1}^\top \weight_{L:1} \NewInputs \NewInputs^\top \weight_{i-1:1}^\top\right) \weight_{i-1:1} \OldInputs}^2}\\
            &+ \ex{\Fnorm{ \sum_{i=1}^L \weight_{L:i+1} \left(\weight_{L:i+1}^\top  \NewLabels \NewInputs^\top \weight_{i-1:1}^\top\right) \weight_{i-1:1} \OldInputs}^2}\\
            &+2 \sum_{i < j} \trace{(\weight_{L:1} \OldInputs - \OldLabels) \OldInputs^\top \times \ex{\weight_{L:1}\left[\frac{\mG_i}{\weight_i}, \frac{\mG_j}{\weight_j}\right]}^\top}.
    \end{align}
    See \cref{eq:error_related_term} for the expression for $\ex{\weight_{L:1}\left[\frac{\mG_i}{\weight_i}, \frac{\mG_j}{\weight_j}\right]}$.
\end{lemma}

\begin{proof}
    We repeat the proof of \cref{lemma:hessian_trace} with $\vec{r}$ replaced by $\vg$ and $\mat{R}_j$ by $\mG_j$ until \cref{eq:reusable_vector_hessian_product} because these steps does not rely on any specific structure of $\vec{r}$. 
    As a result, we have
    \begin{align}
        \trace{\vg^\top \oldhessian \vg}
        =& \sum_{j=1}^L \ex{\trace{\mG_j^\top \partialfrac{\inner{\partialfrac{\emploss}{\param}}{\vg}}{\weight_j}}},
    \end{align}
    and 
    \begin{align}
        \diff \inner{\partialfrac{\emploss}{\param}}{\vg}
        =& \trace{ \weight_{L:j+1} \diff \weight_j \weight_{j-1:1} \OldInputs \OldInputs^\top \left( \sum_{i=1}^L \weight_{L:i+1} \mG_i \weight_{i-1:1} \right)^\top }\\
            &+ \sum_{i=1}^L \trace{(\weight_{L:1} \OldInputs - \OldLabels) \OldInputs^\top \diff (\weight_{L:i+1} \mG_i \weight_{i-1:1})^\top}. 
    \end{align}
    Replacing $\diff \weight_j$ with $\mG_j$ leads to
    \begin{align}
        \trace{\mG_j^\top \partialfrac{\inner{\partialfrac{\emploss}{\param}}{\vg}}{\weight_j}}
        =&  \trace{ \weight_{L:j+1} \mG_j \weight_{j-1:1} \OldInputs \OldInputs^\top \left( \sum_{i=1}^L \weight_{L:i+1} \mG_i \weight_{i-1:1} \right)^\top }\\
            &+ \sum_{i \in [L] \setminus \set{j}} \trace{(\weight_{L:1} \OldInputs - \OldLabels) \OldInputs^\top \left(\weight_{L:1}\left[\frac{\mG_i}{\weight_i}, \frac{\mG_j}{\weight_j}\right]\right)^\top},
    \end{align}
    where $\mat{A}_{L:1}\left[\frac{\mat{B}_i}{\mat{A}_i}, \frac{\mat{B}_j}{\mat{A}_j}\right]$ denotes product after replacement:
    \begin{align}
        \mat{A}_{L:1}\left[\frac{\mat{B}_i}{\mat{A}_i}, \frac{\mat{B}_j}{\mat{A}_j}\right] \defeq \mat{C}_{L:1}, \text{where }
         \mat{C}_k \defeq \begin{cases}
            \mat{A}_k & \text{if }  k \neq i, j\\
            \mat{B}_i & \text{if } k = i\\
            \mat{B}_j & \text{if } k = j
        \end{cases}.
    \end{align}
    Taking summation and expectation leads to
    \begin{align}
        \ex{\vg^\top \oldhessian \vg}
        =& \ex{\sum_{j=1}^L \trace{ \weight_{L:j+1} \mG_j \weight_{j-1:1} \OldInputs \OldInputs^\top \left( \sum_{i=1}^L \weight_{L:i+1} \mG_i \weight_{i-1:1} \right)^\top }}\\
            &+ \ex{\sum_{j=1}^L \sum_{i \in [L] \setminus \set{j}} \trace{(\weight_{L:1} \OldInputs - \OldLabels) \OldInputs^\top \left(\weight_{L:1}\left[\frac{\mG_i}{\weight_i}, \frac{\mG_j}{\weight_j}\right]\right)^\top}}\\
        =& \underbrace{\ex{\Fnorm{ \sum_{i=1}^L \weight_{L:i+1} \mG_i \weight_{i-1:1} \OldInputs}^2}}_{\text{error-unrelated term}} \\
            &+ 2 \underbrace{\sum_{i < j} \trace{(\weight_{L:1} \OldInputs - \OldLabels) \OldInputs^\top \times \ex{\weight_{L:1}\left[\frac{\mG_i}{\weight_i}, \frac{\mG_j}{\weight_j}\right]}^\top}}_{\text{error-related term}}.
    \end{align}

    The error-unrelated and -related terms are computed separately.
    For the error-unrelated term, we have
    \begin{align}
        &\ex{\Fnorm{ \sum_{i=1}^L \weight_{L:i+1} \mG_i \weight_{i-1:1} \OldInputs}^2}\\
        =&  \ex{\Fnorm{ \sum_{i=1}^L \weight_{L:i+1} \left(\weight_{L:i+1}^\top (\weight_{L:1} \NewInputs - \NewLabels) \NewInputs^\top \weight_{i-1:1}^\top\right) \weight_{i-1:1} \OldInputs}^2}\\
        =& 
            \ex{\Fnorm{ \sum_{i=1}^L \weight_{L:i+1} \left(\weight_{L:i+1}^\top \weight_{L:1} \NewInputs \NewInputs^\top \weight_{i-1:1}^\top\right) \weight_{i-1:1} \OldInputs}^2}\\
            &+ \ex{\Fnorm{ \sum_{i=1}^L \weight_{L:i+1} \left(\weight_{L:i+1}^\top  \NewLabels \NewInputs^\top \weight_{i-1:1}^\top\right) \weight_{i-1:1} \OldInputs}^2}\\
            &- 2 \sum_{i=1}^L \sum_{j=1}^L \trop \exx \biggl[\left(\weight_{L:i+1} \left(\weight_{L:i+1}^\top \weight_{L:1} \NewInputs \NewInputs^\top \weight_{i-1:1}^\top\right) \weight_{i-1:1} \OldInputs\right) \\ & \qquad \qquad \qquad \qquad \qquad \qquad \times  \left(\weight_{L:j+1} \left(\weight_{L:j+1}^\top \NewLabels \NewInputs^\top \weight_{j-1:1}^\top\right) \weight_{j-1:1} \OldInputs\right)^\top \biggr].
    \end{align}
    Again, due to \cref{assumption:symmetric_inputs} on $\newtask = (\NewInputs, \NewLabels)$, we have conditional symmetry of $\NewInputs$ and all the third moments vanish.
    The error-unrelated term thus simplifies to
    \begin{align}
        \ex{\Fnorm{ \sum_{i=1}^L \weight_{L:i+1} \mG_i \weight_{i-1:1} \OldInputs}^2}
        =& 
            \ex{\Fnorm{ \sum_{i=1}^L \weight_{L:i+1} \left(\weight_{L:i+1}^\top \weight_{L:1} \NewInputs \NewInputs^\top \weight_{i-1:1}^\top\right) \weight_{i-1:1} \OldInputs}^2}\\
            &+ \ex{\Fnorm{ \sum_{i=1}^L \weight_{L:i+1} \left(\weight_{L:i+1}^\top  \NewLabels \NewInputs^\top \weight_{i-1:1}^\top\right) \weight_{i-1:1} \OldInputs}^2}.
    \end{align}

    In the error-related term, after the definitions of $\mG_i$ and $\mG_j$ are plugged in  and the whole product is expanded, a similar third moment shows up and vanishes due to \cref{assumption:symmetric_inputs}. Therefore, when $i < j$, we have
    \begin{align}
        & \ex{\weight_{L:1}\left[\frac{\mG_i}{\weight_i}, \frac{\mG_j}{\weight_j}\right]}\label{eq:error_related_term}\\
        =&  
            \ex{\weight_{L:j+1} \weight_{L:j+1}^\top \weight_{L:1} \NewInputs \NewInputs^\top \weight_{j-1:1}^\top \weight_{j-1:i+1} \weight_{L:i+1}^\top \weight_{L:1} \NewInputs \NewInputs^\top \weight_{i-1:1}^\top \weight_{i-1:1}}\\
            &+\ex{\weight_{L:j+1} \weight_{L:j+1}^\top \NewLabels \NewInputs^\top \weight_{j-1:1}^\top \weight_{j-1:i+1} \weight_{L:i+1}^\top \NewLabels \NewInputs^\top \weight_{i-1:1}^\top \weight_{i-1:1}}.
    \end{align}

    Putting everything together, the theorem is proved.

\end{proof}

%% file: proofs/inductive_bias.tex
\begin{lemma}[Auto-alignment at the end of the old task]\label{lemma:auto-alignment}
    Assume a DLN with weights $(\weight_i)_{i=1}^{L}$ is sufficiently trained on the old task $\task_1$ using MSE loss under $\ltwo$ regularization so that it lies at a local minimum of the regularized loss $\emploss((\weight_i)_{i=1}^L, \task_1) + \lambda \sum_{i=1}^L \Fnorm{\weight_i}^2$ for some $\lambda > 0$. Then we have the following auto-alignment property: \AutoAlignmentProperty 
\end{lemma}

\cref{lemma:auto-alignment} is an immediate implication of the following \cref{lemma:auto-alignment_abstract}, \ie the assumption of \cref{lemma:auto-alignment} implies the assumption of \cref{lemma:auto-alignment_abstract}. 
To see this, assume for contradiction that the DLN weight does not locally minimize the regularization loss under the constraint that $W_{L:1}$ does not change. Then for every neighborhood of $\allweights$, there exists a better regularized weight with the same output (and the same empirical loss), leading to a better regularized loss. This contradicts the assumption that the DLN weight is a local minimum of the regularized loss. 

\begin{lemma}\label{lemma:auto-alignment_abstract}
    Assume a DLN weight $(\weight_i)_{i=1}^{L}$ is a local minimum of the regularization loss $\sum_{i=1}^L \Fnorm{\weight_i}^2$ under the constraint that the product $\weight_{L:1}$ remains the same.
    Then we have the auto-alignment property: \AutoAlignmentProperty 
\end{lemma}

\begin{proof}
    Under the assumption, we have for every $i \in [L-1]$, $(\weight_i, \weight_{i+1})$ is a local minimum of the two-layer regularization loss $\Fnorm{\weight_i}^2 + \Fnorm{\weight_{i+1}}^2$ under the constraint that the two-layer product $\weight_{i+1:i}$ remains the same.
    Otherwise, we can replace $(\weight_i, \weight_{i+1})$ with the better regularized neighbor to reduce the full regularization loss while keeping the full product $\weight_{L:1}$ unchanged.

    As a result, we have $(\weight_i, \weight_{i+1})$ as a local minimizer of the following optimization problem:
    \begin{align}
        \min & \quad \Fnorm{\weight_i}^2 + \Fnorm{\weight_{i+1}}^2 \\
        \text{s.t.} & \quad \weight_{i+1} \weight_{i} = \mat{C}
    \end{align}
    for some constant matrix $\mat{C}$.

    The method of Lagrange multipliers gives us the necessary condition for the local minima. To this end, let 
    \begin{align}
        \lagrangian(\weight_i, \weight_{i+1}, \mat{\Lambda}) = \Fnorm{\weight_i}^2 + \Fnorm{\weight_{i+1}}^2 + \trace{\mat{\Lambda} \times (\weight_{i+1} \weight_{i} - \mat{C})}
    \end{align}
    be the Lagrangian multiplier, whose gradients are
    \begin{align}
        \frac{\partial \lagrangian}{\partial \weight_i} =& 2 \weight_i + \weight_{i+1}^\top \mat{\Lambda}^\top,\\
        \frac{\partial \lagrangian}{\partial \weight_{i+1}} =& 2 \weight_{i+1} + \mat{\Lambda}^\top \weight_{i}^\top.
    \end{align}
    Forcing them to be zero indicates that there exists a matrix $\mat{\Lambda}$ such that
    \begin{align}
        2 \weight_i + \weight_{i+1}^\top \mat{\Lambda}^\top =& 0,\\
        2 \weight_{i+1} + \mat{\Lambda}^\top \weight_{i}^\top =& 0,
    \end{align}
    which implies
    \begin{align}
        2 \weight_i \weight_i^\top + \weight_{i+1}^\top \mat{\Lambda}^\top \weight_i^\top =& 0,\\
        2 \weight_{i+1}^\top \weight_{i+1} + \weight_{i+1}^\top \mat{\Lambda}^\top \weight_{i}^\top =& 0,
    \end{align}
    and
    \begin{align}
        \weight_{i+1}^\top \weight_{i+1} = - \frac{1}{2} \weight_{i+1}^\top \mat{\Lambda}^\top \weight_{i}^\top =\weight_{i}^\top \weight_{i}^\top.
    \end{align}
\end{proof}

\begin{lemma}[Implication of auto-alignment.]\label{lemma:auto_alignment_implication_singular_values}
    Assume a DLN with weights $(\weight_i)_{i=1}^{L}$ satisfies the auto-alignment property: \AutoAlignmentProperty.

    Then the singular values of every weight matrix are the same: Denoting $\SingularMatrix{\weight_i}$ as the diagonal matrix whose diagonal entries are singular values of $\weight_i$ in decreasing order, we have $\SingularMatrix{\weight_i} = \SingularMatrix{\weight_j}$ for every $i, j \in [L]$.

    Moreover, there exists a series of orthogonal matrices $((\mU_i, \mV_i))_{i=1}^L$ as the singular vector matrices of weights (\ie $\weight_i = \mU_i \SingularMatrix{\weight_i} \mV_i^\top$) such that the adjacent weights share the same "adjacent-side" singular vectors, \ie $\mV_{i+1} = \mU_i$.
\end{lemma}
\begin{proof}
    Since by the uniqueness of singular values, we have $\SingularMatrix{\weight_{i+1}^\top \weight_{i+1}} = \SingularMatrix{\weight_i \weight_i^\top}$. Since $\SingularMatrix{\weight_{i+1}^\top \weight_{i+1}} = \SingularMatrix{\weight_{i+1}}^\top \SingularMatrix{\weight_{i+1}}$ and $\SingularMatrix{\weight_i \weight_i^\top} = \SingularMatrix{\weight_i} \SingularMatrix{\weight_i}^\top$, we have $\SingularMatrix{\weight_{i+1}} = \SingularMatrix{\weight_i}$.

    Now we construct the singular vectors by induction. 
    The inductive hypothesis at step $i$ is that for $j \ge i$, the decomposition of $\weight_j$ and they satisfy $\mV_{j+1} = \mU{j}$ for $j \in [i, L-1]$.
    \begin{itemize}
        \item Base: When $i=L$, pick any left singular vector matrix $\mU_L$ such that there exists a right singular vector matrix $\mV_L$ such that $\weight_L = \mU_L \SingularMatrix{\weight_L} \mV_L^\top$.
        \item Induction: When $i < L$, assume the inductive hypothesis holds for $i+1$. Let $(\mU'_i, \mV'_i)$ be any singular vector matrices such that $\weight_i = \mU'_i \SingularMatrix{\weight_i} \left(\mV'_i\right)^\top$. Since $\weight_{i+1}^\top \weight_{i+1} = \weight_i \weight_i^\top$, we have
        \begin{align}
            \mV_{i+1} \Sigma_{\weight_{i+1}}^\top \Sigma_{\weight_{i+1}} \mV_{i+1}^\top = \mU'_i \Sigma_{\weight_i} \Sigma_{\weight_i}^\top \left(\mU'_i\right)^\top.
        \end{align}
        By the uniqueness of singular value decomposition, when singular values are distinct, we have $\mV_{i+1} = \mU'_i$; when singular values are repeated, we $\mV_{i+1}$ and $\mU'_i$ are unique up to orthogonal transforms within the subspaces spanned by each group of repeated singular values. That is, there exists an orthogonal matrix $\mat{O}$, which is block-diagonal and the diagonal blocks are orthogonal matrices whose sizes are the same as the number of repeated singular values, such that $\mV_{i+1} = \mU'_i \mat{O}$.
        By its block-diagonal structure, $\mat{O}$ is commutative with $\SingularMatrix{\weight_i} \SingularMatrix{\weight_i}^\top$ and $\Sigma_{\weight_i}$. As a result, we have $\weight_i = \mU'_i \mat{O} \mat{O}^\top \SingularMatrix{\weight_i} \left(\mV'_i\right)^\top = \mV_{i+1} \SingularMatrix{\weight_i} \mat{O}^\top \left(\mV'_{i}\right)^\top$.
        Setting $\mU_{i} = \mV_{i+1}$ and $\mV_{i} = \mV'_{i} \mat{O}$ finishes this inductive step.
    \end{itemize}

\end{proof}

%% file: proofs/lowerbounding.tex
Using the lemmas from previous subsections, we can now prove the lowerbound of the alignment between the old and new tasks.
To make the lowerbound more simple and clear, we first assume idealized conditions. They essentially include assumptions that the old input data is whitened, the new task is generated by randomly rotating and reflecting the old input data, and the old task is well-trained so that the DLN interpolates the old task well and reaches a local minimum of \emph{the regularized loss}:

\begin{LocalAssumption}[Whitened old task.]\label{assumption:whitened_old_task}
    Assume the old task has whitened inputs, \ie $\EmpiricalInputs_1 \EmpiricalInputs_1^\top = I$.
\end{LocalAssumption}

\begin{LocalAssumption}[Generation of the new task.]\label{assumption:generate_new_task}
    The new task $(\EmpiricalInputs_2, \EmpiricalLabels_2)$ are generated by randomly rotating the inputs of the olds task, \ie 1) sampling a random orthogonal matrix $\RandomOrthogonalMatrix$ from the Haar measure on the orthogonal group, and 2) computing the new task as $\EmpiricalInputs_2 \defeq \RandomOrthogonalMatrix \EmpiricalInputs_1$ and $\EmpiricalLabels_2 \defeq \EmpiricalLabels_1$. Note that this assumption implies \cref{assumption:symmetric_inputs}.
\end{LocalAssumption}

\begin{LocalAssumption}\label{assumption:empirical_loss_local_minima}
    The DLN reaches a local minimum of the old task's empirical loss.
\end{LocalAssumption}

\newcommand{\Sdiff}{\mat{\Delta}}
\begin{LocalAssumption}[The DLN well interpolates the old task.]\label{assumption:interpolate_old_task}
    Let 
    \begin{align}
        \Sdiff \defeq \weight_{L:1}^\pinv \times \OldLabels \OldInputs^\pinv - \mIL{\oldYX}
    \end{align}
    be the relative (spectral) difference between the solution given by DLN and the old-task ground truth, where $(\cdot)^\dagger$ denotes the Moore-Penrose pseudo inverse, and $\mIL{\mA} \defeq \mA^\dagger \mA$. Based on $\Sdiff$, we assume
    \begin{itemize}
        \item the DLN is not rank-deficient on the old task, \ie{} the left and right nullspaces $\weight_{L:1}$ are the subsets of those of $\OldLabels \OldInputs^\pinv$, respectively;
        \item the DLN does not over-estimate the rank of the old task too much, \ie{} 
        the spectrum of $\weight_{L:1}$ falling into $\oldYX$'s nullspace is relatively smaller than $\tau \ll \frac{1}{3}$: 
            \begin{align}
                \forall k \in \set{1, \dots, 2L},\, 
                    \Fnorm{\weight_{L:1}^{3 - k / L} \times (\mI - \mIL{\oldYX})}^2 \le \tau \cdot \Fnorm{\weight_{L:1}^{3 - k / L}}^2;
            \end{align}
        \item the DLN well interpolates the old task, \ie{} 
            \begin{align}
                \snorm{\Sdiff} \defto \rho \ll \frac{1}{3},
            \end{align}
            where $\snorm{\cdot}$ is the spectral norm, \ie{} the largest singular value of a matrix.
    \end{itemize}
\end{LocalAssumption}

\begin{lemma}\label{lemma:property_of_spectral_diff}
    For any real matrix $\mA$, we have 
    \begin{align}
        \mA \times \mIL{\mA} = \mA,\, \mIL{\mA} \mA^\top = \mA^\top,\\
        \left(\mIL{\mA}\right)^\top = \mIL{\mA},\, \mIL{\mA} \mIL{\mA} = \mIL{\mA},
    \end{align}
    and $\mIL{\mA}$'s nullspace is the same $\mA$'s right nullspace.

    Under \cref{assumption:interpolate_old_task}, we have the following facts:
    \begin{itemize}
        \item $\OldLabels \OldInputs^\top = \weight_{L:1} (\mIL{\oldYX} + \Sdiff)$;
        \item The matrix $\Sdiff$'s right nullspace is the superset of the right nullspace of $\oldYX$ and $\Sdiff \times \mIL{\oldYX} = \Sdiff$;
        \item Since $\snorm{\Sdiff} \defto \rho < 1$, the matrix $\mIL{\oldYX} + \Sdiff = \weight_{L:1}^\pinv \times \OldLabels \OldInputs^\pinv$ has the same right nullspace as $\oldYX$'s right nullspace;
        \item $1 - \rho \le \sigma_{\min}(\mIL{\oldYX} + \Sdiff) \le \sigma_{1}(\mIL{\oldYX} + \Sdiff) \le 1 + \rho$.
        \item For any $k \le 2L$, we have 
            \begin{align}
                \trace{\left(\weight_{L:1}^\top \weight_{L:1}\right)^{3 - k / L} \times \mIL{\oldYX}} \ge (1 - \tau) \cdot \trace{\left(\weight_{L:1}^\top \weight_{L:1}\right)^{3 -  k / L}}.
            \end{align}
    \end{itemize}
\end{lemma}
\begin{proof}
    The properties of $\mIL{\mA}$ is direct from the definition of the Moore-Penrose pseudo inverse.

    Now we assume \cref{assumption:interpolate_old_task}.
    The definition of $\Sdiff$ implies that
    \begin{align}
        \weight_{L:1} \weight_{L:1}^\pinv \times \OldLabels \OldInputs^\pinv = \weight_{L:1} (\mIL{\oldYX} + \Sdiff).
    \end{align}
    Since the left nullspace of $\weight_{L:1}$ is the subset of that of $\OldLabels \OldInputs^\pinv$, we have $\weight_{L:1} \weight_{L:1}^\pinv \times \OldLabels \OldInputs^\pinv = \OldLabels \OldInputs^\top$, leading to $\OldLabels \OldInputs^\top = \weight_{L:1} (\mIL{\oldYX} + \Sdiff)$.

    By construction of $\weight_{L:1}^\pinv \times \oldYX$, its right nullspace is the superset of the right nullspace of $\oldYX$. Since the right nullspace of $\mIL{\oldYX}$ is also the superset of the right nullspace of $\oldYX$, their difference $\Sdiff$ also has a right nullspace that is the superset of the right nullspace of $\oldYX$.

    Finally, since $\snorm{\Sdiff} \defto \rho < 1$, for any non-zero vector $\vec{u}$ that is orthogonal to the right nullspace of $\oldYX$, we have
    \begin{align}
        \norm{\vec{u}^\top (\mIL{\oldYX} + \Sdiff)} \ge \norm{\vec{u}^\top \mIL{\oldYX}} - \norm{\vec{u}^\top \Sdiff} \ge \norm{\vec{u}} - \rho \cdot \norm{\vec{u}} > 0. \label{eq:non_zero_least_singular_value}
    \end{align}
    Therefore, any vector that is orthogonal to the right nullspace of $\oldYX$ is not in the right nullspace of $\mIL{\oldYX} + \Sdiff$, indicating that the right nullspace of $\mIL{\oldYX} + \Sdiff$ is the subset of the right nullspace of $\oldYX$. Combining the above fact that $\mIL{\oldYX} + \Sdiff = \weight_{L:1}^\pinv \times \OldLabels \OldInputs^\pinv$ has a superset nullspace than $\oldYX$, we conclude that $\mIL{\oldYX} + \Sdiff$ has the same right nullspace as $\oldYX$'s right nullspace.
    
    For the range of non-zero singular values, we only need probe within the subspace orthogonal to the right nullspace of $\oldYX$. The lower-bound is already proved in \cref{eq:non_zero_least_singular_value}. For the upper-bound, we have
    \begin{align}
        \sigma_{1}(\mIL{\oldYX} + \Sdiff) \le \snorm{\mIL{\oldYX}} + \snorm{\Sdiff} \le 1 + \rho.
    \end{align}
    The last inequality is direct from \cref{assumption:interpolate_old_task}, the connection between the Frobenius norm and the trace, and the properties of $\mIL{\mA}$.
\end{proof}

\begin{LocalAssumption}[The DLN is well-trained by regularized loss.]\label{assumption:well_trained_old_task}
    Assume the DLN is well-trained on the old task by regularization, \ie it lies at a local minimum of the regularized loss $\emploss((\weight_i)_{i=1}^L, \task_1) + \lambda \sum_{i=1}^L \Fnorm{\weight_i}^2$ for some $\lambda > 0$.
\end{LocalAssumption}

\newcommand{\ProductWithReplacement}{\mat{E}}
With these assumptions, we can simplify the alignment between the old Hessian and the new gradient.
\begin{lemma}\label{lemma:alignment_full_expression}
    Under \cref{assumption:whitened_old_task,assumption:generate_new_task,assumption:empirical_loss_local_minima,assumption:interpolate_old_task,assumption:well_trained_old_task}, we have
    \begin{align}
        &\alignment(\oldhessian, \vg)
        \defeq \dim \param \cdot \frac{\ex{\vg^\top \oldhessian \vg}}{\trace{\oldhessian} \cdot \ex{\norm{\vg}^2}}\\
        =& \dim \param \cdot {\textstyle
            \frac{\substack{
                    \Fnorm{ \sum_{i=1}^L \weight_{L:i+1}\weight_{L:i+1}^\top \weight_{L:1}  \weight_{i-1:1}^\top\weight_{i-1:1}}^2
                +
                    2 \sum_{i<j} \trace{(\weight_{L:1} \OldInputs - \OldLabels) \OldInputs^\top \ProductWithReplacement_{i, j}^\top}\\
                +
                    \frac{1}{\dim \sampleinput} \sum_{i=1}^L \sum_{j=1}^L \Fnorm{\weight_{i-1:1} \weight_{j-1:1}^\top}^2 \cdot \trace{\weight_{L:i+1} \weight_{L:i+1}^\top  \OldLabels \OldInputs^\top \OldInputs \OldLabels^\top \weight_{L:j+1} \weight_{L:j+1}^\top}        
            }}{\substack{
                \sum_{i=1}^L \Fnorm{\weight_{L:i+1}}^2 \cdot \Fnorm{\weight_{i-1:1}}^2\\
                \cdot 
                \left(
                    \sum_{i=1}^L \Fnorm{\weight_{L:i+1}^\top \weight_{L:1} \weight_{i-1:1}^\top}^2 
                    + 
                    \frac{1}{\dim \sampleinput} \sum_{i=1}^L  \Fnorm{\weight_{L:i+1}^\top \OldLabels \OldInputs^\top}^2 \cdot \Fnorm{\weight_{i-1:1}}^2
                \right)
            }}
        }\\
        =& \dim \param \cdot \scalebox{0.75}{\text{${\textstyle
            \frac{\substack{
                    \Fnorm{ \sum_{i=1}^L \weight_{L:i+1}\weight_{L:i+1}^\top \weight_{L:1}  \weight_{i-1:1}^\top\weight_{i-1:1}}^2
                +
                    2 \sum_{i<j} \trace{(-\Sdiff)  \ProductWithReplacement_{i, j}^\top \weight_{L:1}}\\
                +
                    \frac{1}{\dim \sampleinput} \sum_{i=1}^L \sum_{j=1}^L \Fnorm{\weight_{i-1:1} \weight_{j-1:1}^\top}^2 \cdot \trace{\weight_{L:1}^\top \weight_{L:j+1} \weight_{L:j+1}^\top \weight_{L:i+1} \weight_{L:i+1}^\top \weight_{L:1} \left(\mIL{\oldYX} + \Sdiff\right) \left(\mIL{\oldYX} + \Sdiff\right)^\top}        
            }}{\substack{
                \sum_{i=1}^L \Fnorm{\weight_{L:i+1}}^2 \cdot \Fnorm{\weight_{i-1:1}}^2\\
                \cdot 
                \left(
                    \sum_{i=1}^L \Fnorm{\weight_{L:i+1}^\top \weight_{L:1} \weight_{i-1:1}^\top}^2 
                    + 
                    \frac{1}{\dim \sampleinput} \sum_{i=1}^L  \Fnorm{\weight_{L:i+1}^\top \weight_{L:1} \left(\mIL{\oldYX} + \Sdiff\right)}^2 \cdot \Fnorm{\weight_{i-1:1}}^2
                \right)
            }}
        }$}},
        \label{eq:complicated_matrix_product}
    \end{align}
    where the definition of $\ProductWithReplacement_{i, j}$ can be found in \cref{eq:product_with_replacement}.
\end{lemma}
\begin{proof}
    By \cref{lemma:hessian_trace} and the assumption that $\OldInputs \OldInputs^\top = \mat{I}$, we have 
    \begin{align}
        \trace{\oldhessian}
        = \sum_{i=1}^L \Fnorm{\weight_{L:i+1}}^2 \cdot \Fnorm{\weight_{i-1:1} \EmpiricalInputs}^2
        =& \sum_{i=1}^L \Fnorm{\weight_{L:i+1}}^2 \cdot \Fnorm{\weight_{i-1:1}}^2.
    \end{align}

    Since the old inputs are whitened and the new inputs is generated by multiplying a random orthogonal matrix, the new inputs is also whitened, \ie $\NewInputs \NewInputs^\top = \RandRot \OldInputs \OldInputs^\top \RandRot = \mat{I}$.
    As a result, we can simplify the expected norm of the new gradients in \cref{lemma:new_gradient_norm} as
    \begin{align}
        \ex{\norm{\vg}^2}
        =& \sum_{i=1}^L \ex{
                \Fnorm{\weight_{L:i+1}^\top \weight_{L:1} \NewInputs \NewInputs^\top \weight_{i-1:1}^\top}^2
                +\Fnorm{\weight_{L:i+1}^\top \NewLabels \NewInputs^\top \weight_{i-1:1}^\top}^2 
        }\\
        =& \sum_{i=1}^L \Fnorm{\weight_{L:i+1}^\top \weight_{L:1} \weight_{i-1:1}^\top}^2 + \sum_{i=1}^L \ex{\Fnorm{\weight_{L:i+1}^\top \OldLabels \OldInputs^\top \RandRot^\top \weight_{i-1:1}^\top}^2}\\
    \end{align}
    The $\RandRot$-related term can be further simplified by \cref{corollary:second_matrix_moment_concrete} to
    \begin{align}
        &   \ex{\Fnorm{\weight_{L:i+1}^\top \OldLabels \OldInputs^\top \RandRot^\top \weight_{i-1:1}^\top}^2}\\
        =& \trace{\weight_{L:i+1}^\top \OldLabels \OldInputs^\top \ex{\RandRot^\top \weight_{i-1:1}^\top \weight_{i-1:1} \RandRot} \OldInputs \OldLabels^\top \weight_{L:i+1}}\\
        =& \trace{\weight_{L:i+1}^\top \OldLabels \OldInputs^\top \frac{\trace{\weight_{i-1:1}^\top \weight_{i-1:1}}}{\dim \sampleinput} \mat{I} \OldInputs \OldLabels^\top \weight_{L:i+1}}\\
        =& \frac{1}{\dim \sampleinput} \Fnorm{\weight_{L:i+1}^\top \OldLabels \OldInputs^\top}^2 \cdot \Fnorm{\weight_{i-1:1}}^2.
    \end{align}
    The expected new gradient norm thus becomes
    \begin{align}
        \ex{\norm{\vg}^2}
        =& \sum_{i=1}^L \Fnorm{\weight_{L:i+1}^\top \weight_{L:1} \weight_{i-1:1}^\top}^2 + \frac{1}{\dim \sampleinput} \sum_{i=1}^L  \Fnorm{\weight_{L:i+1}^\top \OldLabels \OldInputs^\top}^2 \cdot \Fnorm{\weight_{i-1:1}}^2\\
    \end{align}

    In a similar manner, we simplify the product between the new gradients and the old Hessian derived in \cref{lemma:gradient_hessian_product} term by term.
    The first term is
    \begin{align}
        &\ex{\Fnorm{ \sum_{i=1}^L \weight_{L:i+1} \left(\weight_{L:i+1}^\top \weight_{L:1} \NewInputs \NewInputs^\top \weight_{i-1:1}^\top\right) \weight_{i-1:1} \OldInputs}^2}\\
        =& \Fnorm{ \sum_{i=1}^L \weight_{L:i+1}\weight_{L:i+1}^\top \weight_{L:1}  \weight_{i-1:1}^\top\weight_{i-1:1}}^2.
    \end{align}
    The second term is
    \begin{align}
        &\ex{\Fnorm{ \sum_{i=1}^L \weight_{L:i+1} \left(\weight_{L:i+1}^\top  \NewLabels \NewInputs^\top \weight_{i-1:1}^\top\right) \weight_{i-1:1} \OldInputs}^2}\\
        =& \ex{\sum_{i=1}^L \sum_{j=1}^L \trace{\weight_{L:i+1} \weight_{L:i+1}^\top  \NewLabels \NewInputs^\top \weight_{i-1:1}^\top \weight_{i-1:1} \weight_{j-1:1}^\top \weight_{j-1:1} \NewInputs \NewLabels^\top \weight_{L:j+1} \weight_{L:j+1}^\top}}\\
        =& \sum_{i, j} \trace{\weight_{L:i+1} \weight_{L:i+1}^\top  \OldLabels \OldInputs^\top \ex{\RandRot^\top \weight_{i-1:1}^\top \weight_{i-1:1} \weight_{j-1:1}^\top \weight_{j-1:1} \RandRot} \OldInputs \OldLabels^\top \weight_{L:j+1} \weight_{L:j+1}^\top}\\
        =& \sum_{i, j} \trace{\weight_{L:i+1} \weight_{L:i+1}^\top  \OldLabels \OldInputs^\top \frac{\trace{\weight_{i-1:1}^\top \weight_{i-1:1} \weight_{j-1:1}^\top \weight_{j-1:1}}}{\dim \sampleinput} \mat{I} \OldInputs \OldLabels^\top \weight_{L:j+1} \weight_{L:j+1}^\top}\\
        =& \frac{1}{\dim \sampleinput} \sum_{i=1}^L \sum_{j=1}^L \Fnorm{\weight_{i-1:1} \weight_{j-1:1}^\top}^2 \cdot \trace{\weight_{L:i+1} \weight_{L:i+1}^\top  \OldLabels \OldInputs^\top \OldInputs \OldLabels^\top \weight_{L:j+1} \weight_{L:j+1}^\top}.
    \end{align}
    The essential part of the third term can be simplified by \cref{lemma:untransposed_second_matrix_moment} as
    \begin{align}
        & \ex{\weight_{L:1}\left[\frac{\mG_i}{\weight_i}, \frac{\mG_j}{\weight_j}\right]}\\
        =&  
            \weight_{L:j+1} \weight_{L:j+1}^\top \weight_{L:1} \weight_{j-1:1}^\top \weight_{j-1:i+1} \weight_{L:i+1}^\top \weight _{L:1} \weight_{i-1:1}^\top \weight_{i-1:1}\\
            &+\weight_{L:j+1} \weight_{L:j+1}^\top  \OldLabels \OldInputs^\top \ex{\RandRot^\top \weight_{j-1:1}^\top \weight_{j-1:i+1} \weight_{L:i+1}^\top \OldLabels \OldInputs^\top \RandRot^\top} \weight_{i-1:1}^\top \weight_{i-1:1}\\
        =&  
            \weight_{L:j+1} \weight_{L:j+1}^\top \weight_{L:1} \weight_{j-1:1}^\top \weight_{j-1:i+1} \weight_{L:i+1}^\top \weight _{L:1} \weight_{i-1:1}^\top \weight_{i-1:1}\\
            &+\frac{1}{\dim \sampleinput}\weight_{L:j+1} \weight_{L:j+1}^\top  \OldLabels \OldInputs^\top \left(\weight_{j-1:1}^\top \weight_{j-1:i+1} \weight_{L:i+1}^\top \OldLabels \OldInputs^\top \right)^\top \weight_{i-1:1}^\top \weight_{i-1:1}\\
        =&  
            \weight_{L:j+1} \weight_{L:j+1}^\top \weight_{L:1} \weight_{j-1:1}^\top \weight_{j-1:i+1} \weight_{L:i+1}^\top \weight _{L:1} \weight_{i-1:1}^\top \weight_{i-1:1}\\
            &+\frac{1}{\dim \sampleinput}\weight_{L:j+1} \weight_{L:j+1}^\top  \OldLabels \OldInputs^\top \OldInputs \OldLabels^\top \weight_{L:i+1} \weight_{j-1:i+1}^\top \weight_{j-1:1} \weight_{i-1:1}^\top \weight_{i-1:1}\\
        =&  
            \weight_{L:j+1} \weight_{L:j+1}^\top \weight_{L:1} \weight_{j-1:1}^\top \weight_{j-1:i+1} \weight_{L:i+1}^\top \weight_{L:1} \weight_{i-1:1}^\top \weight_{i-1:1}\\
            &+\frac{1}{\dim \sampleinput}\weight_{L:j+1} \weight_{L:j+1}^\top  \weight_{L:1} (\mIL{\oldYX} + \Sdiff) (\mIL{\oldYX} + \Sdiff)^\top \weight_{L:1}^\top \weight_{L:i+1} \weight_{j-1:i+1}^\top \weight_{j-1:1} \weight_{i-1:1}^\top \weight_{i-1:1}\\
        \defto& \ProductWithReplacement_{i, j}. \label{eq:product_with_replacement}
    \end{align}
    After putting everything together and applying \cref{lemma:property_of_spectral_diff} to replace $\oldYX$, the theorem is proved. 

\end{proof}

Now we have expressed the alignment into a fraction involving complicated weight matrix products.
These products feature consecutive matrix multiplication where adjacent matrices are multiplied together.
Auto-alignment property provides further simplification because it shows that among two adjacent weights, the left singular vectors of the shallower weight equal to the right singular vectors of the deeper weight (\cref{lemma:auto_alignment_implication_singular_values}). As a result, the two ``adjacent''  singular vector matrices cancel each other when multiplied together, simplifying the complicated general matrix product into simple (singular value) diagonal matrix products. 
It leads to the following formal result: 

\begin{lemma}\label{lemma:alignment_lowerbound_ideal}
    Under \cref{assumption:whitened_old_task,assumption:generate_new_task,assumption:empirical_loss_local_minima,assumption:interpolate_old_task,assumption:well_trained_old_task}.
    Let $\CommonSV \defeq \SingularMatrix{\weight_{L:1}}^{\frac{1}{L}}$.
    Then the alignment between the old Hessian and the new gradient has the lower-bound
    \begin{align}
        \alpha(\oldhessian, \vg)
        \ge& \dim \param \cdot {\textstyle
            \frac{\substack{
                    \left(1 - \rho - \frac{\rho (1 + \rho)^2}{\dim \Input}\right) \cdot L^2 \cdot \Fnorm{\CommonSV^{3L-2}}^2
                +
                    \frac{1 - \tau - 2 \rho}{\dim \sampleinput} \sum_{i=1}^L \sum_{j=1}^L \Fnorm{\CommonSV^{i+j-2}}^2 \cdot \Fnorm{\CommonSV^{3L-(i+j)}}^2        
            }}{\substack{
                \sum_{i=1}^L \Fnorm{\CommonSV^{L-i}}^2 \cdot \Fnorm{\CommonSV^{i-1}}^2
                \cdot 
                \left(
                    L \cdot \Fnorm{\CommonSV^{2L-1}}^2 
                    + 
                    \frac{(1 + \rho)^2}{\dim \sampleinput} \sum_{i=1}^L  \Fnorm{\CommonSV^{2L-i}}^2 \cdot \Fnorm{\CommonSV^{i-1}}^2
                \right).
            }}
        }.
    \end{align}

\end{lemma}

\begin{proof}

    Recall that \cref{lemma:auto_alignment_implication_singular_values} shows that when auto-balanced property is satisfied, there exist singular value decompositions of the weights such that $\mV_{i+1} = \mU_i$ and $\SingularMatrix{\weight_i} = \SingularMatrix{\weight_j}$.
    As a result, when $a < b$, consecutive weight product 
    \begin{align}
        \weight_{b:a} 
        =& \mU_b \SingularMatrix{\weight_b} \mV_b^\top \mU_{b-1} \SingularMatrix{\weight_{b-1}} \mV_{b-1}^\top \cdots \mU_{a+1} \SingularMatrix{\weight_{a+1}} \mV_{a+1}^\top \mU_{a} \SingularMatrix{\weight_{a}} \mV_{a}^\top \\
        =& \mU_b \SingularMatrix{\weight_b} \mI{} \SingularMatrix{\weight_{b-1}} \mI{} \cdots \mI \SingularMatrix{\weight_{a+1}} \mI{} \SingularMatrix{\weight_{a}} \mV_{a}^\top \\
        =& \mU_b \Sigma_{\weight_{b:a}} \mV_a^\top.
    \end{align}
    Particularly, we have $\weight_{L:1} = \mU_L \SingularMatrix{\weight_i}^L \mV_1^\top$, indicating $\SingularMatrix{\weight_i}^L = \SingularMatrix{\weight_{L:1}}$ and $\SingularMatrix{\weight_i} \equiv \CommonSV$.
    This simplifies consecutive weight products into essentially simple diagonal matrices $\weight_{b:a} = \mU_b \CommonSV^{b-a+1} \mV_a^\top$.
    As a result, we can simplify the result of \cref{lemma:alignment_full_expression} into 
    \begin{align}
        \alignment(\oldhessian, \vg)
        \ge& \dim \param \cdot {\textstyle
            \frac{\substack{
                    \Fnorm{ \sum_{i=1}^L \CommonSV^{3L-2}}^2
                -
                    2 \sum_{i<j} \abs{\trace{\Sdiff \times  \ProductWithReplacement_{i, j}^\top \weight_{L:1}}}\\
                +
                    \frac{1}{\dim \sampleinput} \sum_{i=1}^L \sum_{j=1}^L \Fnorm{\CommonSV^{i+j-2}}^2 \cdot \trace{\mV_1 \CommonSV^{6L - 2(i+j)} \mV_1^\top \left(\mIL{\oldYX} + \Sdiff\right) \left(\mIL{\oldYX} + \Sdiff\right)^\top}        
            }}{\substack{
                \sum_{i=1}^L \Fnorm{\CommonSV^{L-i}}^2 \cdot \Fnorm{\CommonSV^{i-1}}^2\\
                \cdot 
                \left(
                    \sum_{i=1}^L \Fnorm{\CommonSV^{2L-1}}^2 
                    + 
                    \frac{1}{\dim \sampleinput} \sum_{i=1}^L  \Fnorm{\mV_{i+1} \CommonSV^{2L-i} \mV_1^\top \left(\mIL{\oldYX} + \Sdiff\right)}^2 \cdot \Fnorm{\CommonSV^{i-1}}^2
                \right)
            }}
        },
    \end{align}
    Now we handle $\Sdiff$ terms. 
    The first one is $\trace{\mV_1 \CommonSV^{6L - 2(i+j)} \mV_1^\top \times (\mIL{\oldYX} + \Sdiff)(\mIL{\oldYX} + \Sdiff)^\top}$. We note that both $\mV_1 \CommonSV^{6L - 2(i+j)} \mV_1^\top$ and $(\mIL{\oldYX} + \Sdiff)(\mIL{\oldYX} + \Sdiff)^\top$ are both real, symmetric and PSD matrices. 
    By von Neumann's trace inequality, we have $\abs{\trace{\mM \mA}} \le \sum_i \sigma_i(\mM) \cdot \sigma_i(\mA) \le \sigma_1(\mM) \cdot \trace{\mA}$ for PSD $\mA$. Combining this fact with $\trace{\mA \mB} \ge 0$ for PSD $\mA, \mB$ and \cref{lemma:property_of_spectral_diff}, we have
    \begin{align}
        & \trace{\mV_1 \CommonSV^{6L - 2(i+j)} \mV_1^\top \times (\mIL{\oldYX} + \Sdiff)(\mIL{\oldYX} + \Sdiff)^\top}\\
        \ge& 
                \trace{\mV_1 \CommonSV^{6L - 2(i+j)} \mV_1^\top \times \mIL{\oldYX}}
            +
                2 \trace{\mV_1 \CommonSV^{6L - 2(i+j)} \mV_1^\top \times \Sdiff} \\
            &+
                \trace{\mV_1 \CommonSV^{6L - 2(i+j)} \mV_1^\top \times \Sdiff \Sdiff^\top}\\
        \ge& 
                \trace{\mV_1 \CommonSV^{6L - 2(i+j)} \mV_1^\top \times \mIL{\oldYX}}
            -
                2 \abs{\trace{\mV_1 \CommonSV^{6L - 2(i+j)} \mV_1^\top \times \Sdiff}} \\
        \ge& 
                (1 - \tau) \cdot \trace{\mV_1 \CommonSV^{6L - 2(i+j)} \mV_1^\top}
            -
                2 \sigma_1(\Sdiff) \cdot \trace{\mV_1 \CommonSV^{6L - 2(i+j)} \mV_1^\top} \\
        =& (1 - \tau - 2 \rho) \cdot \trace{\mV_1 \CommonSV^{6L - 2(i+j)} \mV_1^\top}.
    \end{align}
    The second $\Sdiff$-related term is $\Fnorm{\mV_{i+1} \CommonSV^{2L-i} \mV_1^\top (\mIL{\oldYX} + \Sdiff)}^2$. By the well-known fact that $\Fnorm{\mA \mB} \le \sigma_1(\mA) \cdot \Fnorm{\mB}$, we have
    \begin{align}
        \Fnorm{\mV_{i+1} \CommonSV^{2L-i} \mV_1^\top (\mIL{\oldYX} + \Sdiff)}^2
        \le& \sigma_{1}^2(\mIL{\oldYX} + \Sdiff) \cdot \Fnorm{\mV_{i+1} \CommonSV^{2L-i} \mV_1^\top}^2\\
        \le& (1 + \rho)^2 \cdot \Fnorm{\CommonSV^{2L-i}}^2.
    \end{align} 
    The third $\Sdiff$-related term is $\trace{\Sdiff \times  \ProductWithReplacement_{i, j}^\top \weight_{L:1}}$. 
    This term involves $\ProductWithReplacement_{i, j}$, which can be simplified into
    \begin{align}
        \ProductWithReplacement_{i, j}
        =& \mU_L \CommonSV^{5L-4} \mV_1^\top + \frac{\mU_L \CommonSV^{3L - 2 j} \mV_1^\top \left(\mIL{\oldYX} + \Sdiff\right) \left(\mIL{\oldYX} + \Sdiff\right)^\top \mV_1 \CommonSV^{2L+2j-4} \mV_1^\top}{\dim \Input},\\
        \ProductWithReplacement_{i, j}^\top \weight_{L:1}
        =& \underbrace{\mV_1 \CommonSV^{6L-4} \mV_1^\top}_{\text{PSD}} + \frac{\overbrace{\mV_1 \CommonSV^{2L+2j-4} \mV_1^\top}^{\text{PSD}} \overbrace{\left(\mIL{\oldYX} + \Sdiff\right) \left(\mIL{\oldYX} + \Sdiff\right)^\top}^{\sigma_1 \le (1 + \rho)^2} \overbrace{\mV_1 \CommonSV^{4L - 2 j} \mV_1^\top}^{\text{PSD}}}{\dim \Input}.
    \end{align}
    We combine the well-known result that $\abs{\trace{\mA \mM}} \le \sigma_1(\mA) \cdot \trace{\mM}$ for PSD $\mM$ and \cref{lemma:trace_bounds_four_components} to have
    \begin{align}
        \abs{\trace{\Sdiff \times  \ProductWithReplacement_{i, j}^\top \weight_{L:1}}}
        \le& \rho \left(1 + \frac{(1 + \rho)^2}{\dim \Input}\right) \cdot \trace{\CommonSV^{6L-4}}\\
        =& \rho \left(1 + \frac{(1+\rho)^2}{\dim \Input}\right) \cdot \Fnorm{\CommonSV^{3L-2}}^2.
    \end{align}

    Combining the above results, we have the following drastic simplification:
    \begin{align}
        &\alignment(\oldhessian, \vg)\\ 
        \ge& \dim \param \cdot {\textstyle
            \frac{\substack{
                    \Fnorm{ \sum_{i=1}^L \CommonSV^{3L-2}}^2
                -
                    L^2 \rho \left(1 + \frac{(1 + \rho)^2}{\dim \Input}\right) \cdot \Fnorm{\CommonSV^{3L-2}}^2
                +
                    \frac{1 - \tau - 2 \rho}{\dim \sampleinput} \sum_{i=1}^L \sum_{j=1}^L \Fnorm{\CommonSV^{i+j-2}}^2 \cdot \trace{\CommonSV^{6L-2(i+j)}}        
            }}{\substack{
                \sum_{i=1}^L \Fnorm{\CommonSV^{L-i}}^2 \cdot \Fnorm{\CommonSV^{i-1}}^2
                \cdot 
                \left(
                    \sum_{i=1}^L \Fnorm{\CommonSV^{2L-1}}^2 
                    + 
                    \frac{(1 + \rho)^2}{\dim \sampleinput} \sum_{i=1}^L  \Fnorm{\CommonSV^{2L-i}}^2 \cdot \Fnorm{\CommonSV^{i-1}}^2
                \right)
            }}
        }\\
        =& \dim \param \cdot {\textstyle
            \frac{\substack{
                    \left(1 - \rho - \frac{\rho (1 + \rho)^2}{\dim \Input}\right) \cdot L^2 \cdot \Fnorm{\CommonSV^{3L-2}}^2
                +
                    \frac{1 - \tau - 2 \rho}{\dim \sampleinput} \sum_{i=1}^L \sum_{j=1}^L \Fnorm{\CommonSV^{i+j-2}}^2 \cdot \Fnorm{\CommonSV^{3L-(i+j)}}^2        
            }}{\substack{
                \sum_{i=1}^L \Fnorm{\CommonSV^{L-i}}^2 \cdot \Fnorm{\CommonSV^{i-1}}^2
                \cdot 
                \left(
                    L \cdot \Fnorm{\CommonSV^{2L-1}}^2 
                    + 
                    \frac{(1 + \rho)^2}{\dim \sampleinput} \sum_{i=1}^L  \Fnorm{\CommonSV^{2L-i}}^2 \cdot \Fnorm{\CommonSV^{i-1}}^2
                \right).
            }}
        }.
    \end{align}
\end{proof}

With the simplified expression of the \AlignmentPhenomenon{}, we can now derive the lowerbound of the alignment.
The first one is the most interpretable one by extracting an $\erank{\cdot}$ from the fractions.

\begin{theorem}\label{theorem:alignment_lowerbound_interpretable}
    Under \cref{assumption:empirical_loss_local_minima,assumption:generate_new_task,assumption:interpolate_old_task,assumption:well_trained_old_task,assumption:whitened_old_task}, we have
    \begin{align}
        \alignment(\oldhessian, \vg)
        \ge& \underbrace{\frac{1 - \rho - \frac{\rho (1 + \rho)^2}{\dim \Input}}{1 + (1 + \rho)^2}}_{\substack{\text{less alignment}\\ \text{under} \\\text{insufficient}\\\text{interpolation}}} \cdot \frac{\dim \param}{\dim \Input \cdot \erank{\SingularMatrix{\weight_{L:1}}^{2(1 - 1 / L)}}}.
    \end{align}
\end{theorem}
\begin{proof}
    By \cref{lemma:partitioned_norm}, we have
    \begin{align}
        \Fnorm{\CommonSV^{i+j-2}}^2 \cdot \Fnorm{\CommonSV^{3L-(i+j)}}^2
        \ge& \Fnorm{\CommonSV^{\frac{3L-2}{2}}}^2 \cdot \Fnorm{\CommonSV^{\frac{3L-2}{2}}}^2
        = \Fnorm{\CommonSV^{\frac{3L-2}{2}}}^4
        \ge \Fnorm{\CommonSV^{3L-2}}^2,\\
        \Fnorm{\CommonSV^{2L-i}}^2 \cdot \Fnorm{\CommonSV^{i-1}}^2
        \le& \Fnorm{\CommonSV^{2L-1}}^2 \cdot \Fnorm{\CommonSV^{0}}^2 =  \Fnorm{\CommonSV^{2L-1}}^2 \cdot \dim \Input\label{eq:looseness_start},\\
        \Fnorm{\CommonSV^{L-i}}^2 \cdot \Fnorm{\CommonSV^{i-1}}^2
        \le& \Fnorm{\CommonSV^{L-1}}^2 \cdot \Fnorm{\CommonSV^{0}}^2 =  \Fnorm{\CommonSV^{L-1}}^2 \cdot \dim \Input\label{eq:looseness_end}.
    \end{align}
    Plugging these to \cref{lemma:alignment_lowerbound_ideal}, we have
    \begin{align}
        \alpha(\oldhessian, \vg)
        \ge& \dim \param \cdot {\textstyle
            \frac{\substack{
                    \left(1 - \rho - \frac{\rho (1 + \rho)^2}{\dim \Input}\right) \cdot L^2 \cdot \Fnorm{\CommonSV^{3L-2}}^2
                +
                    \frac{1 - \tau - 2 \rho}{\dim \sampleinput} \cdot L^2 \cdot \Fnorm{\CommonSV^{3L-2}}^2        
            }}{\substack{
                L \Fnorm{\CommonSV^{L-1}}^2 \cdot \dim \Input
                \cdot 
                \left(
                    L \cdot \Fnorm{\CommonSV^{2L-1}}^2 
                    + 
                    \frac{(1 + \rho)^2}{\dim \sampleinput} L \cdot  \Fnorm{\CommonSV^{2 L-1}}^2 \cdot \dim \Input 
                \right).
            }}
        }\\
        \ge&  \left(1 - \rho - \frac{\rho (1 + \rho)^2}{\dim \Input}\right) \cdot \frac{\dim \param}{(1 + (1 + \rho)^2) \dim \Input} \cdot {\textstyle
            \frac{\substack{
                    \Fnorm{\CommonSV^{3L-2}}^2
            }}{\substack{
                \Fnorm{\CommonSV^{L-1}}^2 \cdot \Fnorm{\CommonSV^{2L-1}}^2
            }}
        }.
    \end{align}
    Now we extract $\erank{\cdot}$Applying \cref{lemma:imbalance_fraction_to_erank}, we have
    \begin{align}
        \frac{\substack{
                \Fnorm{\CommonSV^{L-1}}^2 \cdot \Fnorm{\CommonSV^{2L-1}}^2
        }}{\substack{
                \Fnorm{\CommonSV^{3L-2}}^2
        }} \le \erank{\CommonSV^{2\min(L-1, 2L-1)}} = \erank{\CommonSV^{2(L-1)}}.
    \end{align}
    As a result, we have
    \begin{align}
        \alpha(\oldhessian, \vg)
        \ge&  \frac{1 - \rho - \frac{\rho (1 + \rho)^2}{\dim \Input}}{1 + (1 + \rho)^2} \cdot \frac{\dim \param}{\dim \Input} \cdot \frac{1}{\erank{\CommonSV^{2(L-1)}}}.
    \end{align}
    Using the implicit bias of sufficient $\ltwo$ regularization, we have $\CommonSV \defeq \SingularMatrix{\weight_{L:1}}^{1/L} \approx \SingularMatrix{\OldLabels\OldInputs^\top}^{1/L}$, which finishes the proof.
\end{proof}

Although concise and interpretable, the above lower-bound is loose. The looseness mainly comes from \cref{eq:looseness_start}-\cref{eq:looseness_end}, where we essentially bound the value of $f$ of \cref{lemma:partitioned_norm} by its maximum at the left and right ends. However, in fact, $f$'s plot features a wide and flat valley in the middle, meaning most of these terms are severely over-estimated by the maximum value.
We need to further analyze the changes in $f$.
It turns out that the width of the valley is determined by the spread of its spectrum: When most of the spectrum concentrate in one singular value, the rest close-to-zero singular values will remain vanishing at the middle but increases fast when one of the exponents approach $0$, and the valley is much smaller than the maximum; When the spectrum is uniform (\eg{} $\CommonSV = \mI$), $f$ is close to constant and its valley is the same as the maximum.  
Therefore, we can use rank of $\CommonSV$ to better bound $f$, which strengthens the connection to low-rank bias and benefits the interpretability of the lowerbound.

\begin{theorem}\label{theorem:alignment_lowerbound_tighter}
    Under \cref{assumption:empirical_loss_local_minima,assumption:generate_new_task,assumption:interpolate_old_task,assumption:well_trained_old_task,assumption:whitened_old_task}, we have
    \begin{align}
        \alignment
        \ge&  {\textstyle
            \frac{\substack{
                \left(
                    1 - \rho - \frac{\rho (1 + \rho)^2}{\dim \Input}
                    +
                    \frac{1 - \tau - 2\rho}{L^2 \cdot \dim \sampleinput} \sum_{i=1}^L \sum_{j=1}^L \erank{\CommonSV^{2\max(i+j-2, 3L-(i+j))}}      
            \right)}}{\substack{
                \frac{\sum_{i=1}^L\erank{\CommonSV^{2\min(i-1, L-i)}}}{L} 
                 \left(
                    1
                    + 
                    \frac{(1 + \rho)^2\sum_{i=1}^L \erank{\CommonSV^{2\min(i-1, 2L-i)}}}{L \cdot \dim \sampleinput} 
                \right)
            }}
        } \cdot \frac{\dim \param}{\erank{\SingularMatrix{\weight_{L:1}}^{2(1 - 1 / L)}}}.
    \end{align}
\end{theorem}
\begin{proof}
    By \cref{lemma:imbalance_fraction_to_erank}, we have
    \begin{align}
        \Fnorm{\CommonSV^{i+j-2}}^2 \cdot \Fnorm{\CommonSV^{3L-(i+j)}}^2
        =& \Fnorm{\CommonSV^{3L-2}}^2 \cdot \frac{\Fnorm{\CommonSV^{i+j-2}}^2 \cdot \Fnorm{\CommonSV^{3L-(i+j)}}^2}{\Fnorm{\CommonSV^{3L-2}}^2} \\
        \ge& \Fnorm{\CommonSV^{3L-2}}^2 \cdot \erank{\CommonSV^{2\max(i+j-2, 3L-(i+j))}},\\
        \Fnorm{\CommonSV^{L-i}}^2 \cdot \Fnorm{\CommonSV^{i-1}}^2
        =& \Fnorm{\CommonSV^{L-1}}^2 \cdot \frac{\Fnorm{\CommonSV^{L-i}}^2 \cdot \Fnorm{\CommonSV^{i-1}}^2}{\Fnorm{\CommonSV^{L-1}}^2} \\
        \le& \Fnorm{\CommonSV^{L-1}}^2 \cdot \erank{\CommonSV^{2\min(i-1, L-i)}},
        \\
        \Fnorm{\CommonSV^{2L-i}}^2 \cdot \Fnorm{\CommonSV^{i-1}}^2
        =& \Fnorm{\CommonSV^{2L-1}}^2 \cdot \frac{\Fnorm{\CommonSV^{2L-i}}^2 \cdot \Fnorm{\CommonSV^{i-1}}^2}{\Fnorm{\CommonSV^{2L-1}}^2} \\
        \le& \Fnorm{\CommonSV^{2L-1}}^2 \cdot \erank{\CommonSV^{2\min(i-1, 2L-i)}}. \\
    \end{align}
    Plugging these back to \cref{lemma:alignment_lowerbound_ideal}, we have
    \begin{align}
        &\alpha(\oldhessian, \vg)\\
        \ge& \dim \param \cdot {\textstyle
            \frac{\substack{
                \Fnorm{\CommonSV^{3L-2}}^2 \left(
                    1 - \rho - \frac{\rho(1 + \rho)^2}{\dim \Input}
                    +
                    \frac{1 - \tau - 2\rho}{L^2 \cdot \dim \sampleinput} \sum_{i=1}^L \sum_{j=1}^L \erank{\CommonSV^{2\max(i+j-2, 3L-(i+j))}}      
            \right)}}{\substack{
                 \Fnorm{\CommonSV^{L-1}}^2 \cdot 
                \Fnorm{\CommonSV^{2L-1}}^2 \cdot \frac{\sum_{i=1}^L\erank{\CommonSV^{2\min(i-1, L-i)}}}{L} 
                 \left(
                    1 
                    + 
                    \frac{(1 + \rho)^2\sum_{i=1}^L \erank{\CommonSV^{2\min(i-1, 2L-i)}}}{L \cdot \dim \sampleinput} 
                \right)
            }}
        }\\
        \ge&  {\textstyle
            \frac{\substack{
                \left(
                    1 - \rho - \frac{\rho (1 + \rho)^2}{\dim \Input}
                    +
                    \frac{1 - \tau - 2\rho}{L^2 \cdot \dim \sampleinput} \sum_{i=1}^L \sum_{j=1}^L \erank{\CommonSV^{2\max(i+j-2, 3L-(i+j))}}      
            \right)}}{\substack{
                \frac{\sum_{i=1}^L\erank{\CommonSV^{2\min(i-1, L-i)}}}{L} 
                 \left(
                    1
                    + 
                    \frac{(1 + \rho)^2\sum_{i=1}^L \erank{\CommonSV^{2\min(i-1, 2L-i)}}}{L \cdot \dim \sampleinput} 
                \right)
            }}
        } \cdot \frac{\dim \param}{\erank{\CommonSV^{2(L-1)}}}.
    \end{align}
    Replacing $\CommonSV$ finishes the proof.
\end{proof}

%% file: proofs/extension.tex
In this section, we extend the lower-bounds to settings where the input data are not whitened.

We first need a technical lemma on the fourth moment of random orthogonal matrix:
\begin{lemma}\label{lemma:random_orthogonal_matrix_fourth_moment}
    Let $\mA, \mB$ be real-symmetric matrices and $\mS$ be an uniformly distributed random orthogonal matrix. Then
    \begin{align}
        \ex{\mS \mA \mS^\top \mB \mS \mA \mS^\top}
        = p_{\mA} \cdot \frac{\Fnorm{\mA}^2}{\dim \mS} \cdot \trace{\mB} \cdot \mI
        + q_{\mA} \cdot \frac{\Fnorm{\mA}^2}{\dim \mS} \cdot \mB.
    \end{align}
  where
  \begin{align}
    p_{\mA} = \frac{\dim \mS - \erank{\mA}}{(\dim \mS - 1)(\dim \mS + 2)},
    q_{\mA} = \frac{\erank{\mA} + 1 + (\erank{\mA} - 1) / (\dim \mS - 1)}{\dim \mS + 2}.
  \end{align}
\end{lemma}
\begin{remark}
    The coefficients $p_{\mA}$ and $q_{\mA}$ reflects how relative full-rank $\mA$ is.
    When $\erank{\mA}$ increases from $1$ to full, $p_{\mA}$ drops from $\frac{1}{\dim \mS + 2}$ to $0$ and $q_{\mA}$ increases to $1$. They satisfy a negative correlation $p_{\mA} \cdot \dim \mS + q_{\mA} = 1$.
\end{remark}

\begin{proof}
    \newcommand{\mLambda}{\mat{\Lambda}}
    Let $\mA = \mV_{\mA} \mLambda_{\mA} \mV_{\mA}^\top$ and $\mB = \mV_{\mB} \mLambda_{\mB} \mV_{\mB}^\top$ be their eigenvalue decompositions. By definition of Haar measure, $\mV_{\mB}^\top \mS \mV_{\mA}$ is identically distributed as $\mS$. Therefore, we have
    \begin{align}
        \ex{\mS \mA \mS^\top \mB \mS \mA \mS^\top}
        &= \mV_{\mB} \ex{\mS \mLambda_{\mA} \mS^\top \mLambda_{\mB} \mS \mLambda_{\mA} \mS^\top} \mV_{\mB}^\top.
    \end{align}

    We claim that $\ex{\mS \mLambda_{\mA} \mS^\top \mLambda_{\mB} \mS \mLambda_{\mA} \mS^\top}$ is diagonal. For $i \neq j$, we have 
    \begin{align}
        & \ex{\mS \mLambda_{\mA} \mS^\top \mLambda_{\mB} \mS \mLambda_{\mA} \mS^\top}_{i, j}\\
        =&  \sum_{k,l,p,q,r,s} \ex{
            s_{i, k} \lambda_{\mA, k, l} s_{p, l} \lambda_{\mB, p, q}  s_{q, r} \lambda_{\mA, r, s} s_{j, s}
        }\\
        =& \sum_{k, p, r} \ex{
            s_{i, k} \lambda_{\mA, k} s_{p, k} \lambda_{\mB, p} s_{p, r} \lambda_{\mA, r} s_{j, r}
        } & (\text{$\mLambda_{\mA}, \mLambda_{\mB}$ are diagonal})\\
        =& \sum_{p} \lambda_{\mB, p} \cdot \ex{\inner{\mS_{i, \cdot}}{\vec{\lambda}_{\mA} \hadamard \mS_{p, \cdot}} \cdot  \inner{\vec{\lambda}_{\mA} \hadamard \mS_{p, \cdot}}{\mS_{j, \cdot}}},
    \end{align}
    where $\mS_{i, \cdot}$ is the \emph{column} vector formed by the $i$-th \emph{row} of $\mS$.
    We discuss whether $p$ collides with $i$ or $j$: (1) When $p \not\in \set{i, j}$, we have $i \not\in \set{p, j}$ and $\ex{\mS_{i, \cdot} \mid \mS_{p, \cdot}, \mS_{j, \cdot}} = \vec{0}$. As a reult, we have $\ex{\inner{\mS_{i, \cdot}}{\vec{\lambda}_{\mA} \hadamard \mS_{p, \cdot}} \cdot  \inner{\vec{\lambda}_{\mA} \hadamard \mS_{p, \cdot}}{\mS_{j, \cdot}}} = \ex{\inner{\ex{\mS_{i, \cdot} \mid \mS_{p, \cdot}, \mS_{j, \cdot}}}{\vec{\lambda}_{\mA} \hadamard \mS_{p, \cdot}} \cdot  \inner{\vec{\lambda}_{\mA} \hadamard \mS_{p, \cdot}}{\mS_{j, \cdot}}} = 0$. (2) When $p = i$, since $p = i \neq j$, we have $\ex{\mS_{i, \cdot}^\top \mLambda_{\mA} \mS_{p, \cdot} \mS_{p, \cdot}^\top \mLambda_{\mA} \mid \mS_{j, \cdot}} = \vec{0}^\top$ and $\ex{\inner{\mS_{i, \cdot}}{\vec{\lambda}_{\mA} \hadamard \mS_{p, \cdot}} \cdot  \inner{\vec{\lambda}_{\mA} \hadamard \mS_{p, \cdot}}{\mS_{j, \cdot}}} = \ex{\ex{\mS_{i, \cdot}^\top \mLambda_{\mA} \mS_{p, \cdot} \mS_{p, \cdot}^\top \mLambda_{\mA} \mid \mS_{j, \cdot}} \mS_{j, \cdot}} = 0$. Therefore, when $i \neq j$, we have $\ex{\mS \mLambda_{\mA} \mS^\top \mLambda_{\mB} \mS \mLambda_{\mA} \mS^\top}_{i, j} = 0$ and we only need the diagonal entries:
    \begin{align}
        \ex{\mS \mLambda_{\mA} \mS^\top \mLambda_{\mB} \mS \mLambda_{\mA} \mS^\top}_{ii}
        &= \trace{\ex{\mS \mLambda_{\mA} \mS^\top \ve_i \ve_i^\top \mS \mLambda_{\mA} \mS^\top} \mLambda_{\mB}} \\
        &= \inner{\diag \ex{\mS \mLambda_{\mA} \mS^\top \ve_i \ve_i^\top \mS \mLambda_{\mA} \mS^\top}}{\vec{\lambda}_{\mB}}
    \end{align}
    where $\ve_i$ is the $i$-th standard basis vector. Then we turn to the diagonal elements of $\ex{\mS \mLambda_{\mA} \mS^\top \ve_i \ve_i^\top \mS \mLambda_{\mA} \mS^\top}$:
    \begin{align}
        \ex{\mS \mLambda_{\mA} \mS^\top \ve_i \ve_i^\top \mS \mLambda_{\mA} \mS^\top}_{jj}
        = \ex{\left(\mS_j^\top \mLambda_{\mA} \mS_i\right)^2},
    \end{align}
    where $\mS_{i}$ denote the $i$-th \emph{column} of $\mS$.
    For all $j_1, j_2$ such that $j_1, j_2 \neq i$, the conditional distributions $\prob{\mS_{j_1} \mid \mS_i} = \prob{\mS_{j_2} \mid \mS_i}$ are equal. Therefore, given $i$, all non-$i$ diagonal entries are equal $\ex{\mS \mLambda_{\mA} \mS^\top \ve_i \ve_i^\top \mS \mLambda_{\mA} \mS^\top}_{j_1, j_1} = \ex{\mS \mLambda_{\mA} \mS^\top \ve_i \ve_i^\top \mS \mLambda_{\mA} \mS^\top}_{j_2, j_2}$. 
    Therefore, we only need to compute the sum of the all diagonal entries, \ie{} the trace, and compute the $i$-th diagonal entry in order to recover the whole diagonal.
    The trace is
    \begin{align*}
        \trace{\ex{\mS \mLambda_{\mA} \mS^\top \ve_i \ve_i^\top \mS \mLambda_{\mA} \mS^\top}}
        &= \ex{\trace{\ve_i^\top \mS \mLambda_{\mA} \mLambda_{\mA} \mS^\top \ve_i}} \\
        &= \ex{\left(\mS^\top \mLambda_{\mA} \mLambda_{\mA} \mS\right)_{ii}} \\
        &= \frac{\trace{\mLambda_{\mA}^2}}{\dim \mS} & \text{(\cref{corollary:second_matrix_moment_concrete})}\\
        &= \frac{\Fnorm{\mA}^2}{\dim \mS}
    \end{align*}
    The $i$-th diagonal element is
    \begin{align}
        \ex{\left(\mS_i^\top \mLambda_{\mA} \mS_i\right)^2}
        = \ex{\Fnorm{\mLambda_{\mA}^{1/2} \mS_i}^4}
    \end{align}
    Let (scalar) random variable $K \sim \mathcal{N}(0, \mat{I}_{\dim \mS})$. Then $\norm{\vK} \cdot \mS_i$ is a Gaussian random vector with covariance $\ex{(\norm{\vK} \cdot \mS_i) (\norm{\vK} \cdot \mS_i)^\top} = \ex{\norm{\vK}^2} \ex{\mS_i \mS_i^\top} = \dim \mS \cdot \frac{\mI}{\dim \mS} = \mI$.
    As a result, we have random vector $\vec{N} \defeq \norm{\vK} \cdot \mLambda_{\mA}^{1/2} \mS_i \sim \mathcal{N}(0, \mLambda_{\mA})$.
    According to the following well-known result on the forth moment of Gaussian random vectors
    \begin{align}
        \ex{\vec{X}^\top \vec{X} \vec{X}^\top \vec{X}} = 2 \mat{\Sigma}_{\vec{X}}^2 + \trace{\mat{\Sigma}_{\vec{X}}} \cdot \mat{\Sigma}_{\vec{X}},\\
        \ex{\norm{\vec{X}}^4} = \trace{\ex{\vec{X}^\top \vec{X} \vec{X}^\top \vec{X}}} = 2 \trace{\mat{\Sigma}_{\vec{X}}^2} + \trace{\mat{\Sigma}_{\vec{X}}}^2,
    \end{align}
    we have
    \begin{align}
        \ex{\norm{\vec{N}}^4} 
        =& 2 \trace{\mLambda_{\mA}^2} + \left(\trace{\mLambda_{\mA}}\right)^2 = 2 \Fnorm{\mA}^2 + \left(\trace{\mA}\right)^2\\
        =& \ex{\norm{ \norm{\vK} \cdot\mLambda_{\mA}^{1/2} \mS_i}^4}
        = \ex{\norm{\vK}^4} \cdot \ex{\norm{\mLambda_{\mA}^{1/2} \mS_i}^4} \\
        =& \left(2 \dim \mS + (\dim \mS)^2\right) \cdot \ex{\norm{\mLambda_{\mA}^{1/2} \mS_i}^4}.
    \end{align}
    As a result, we have
    \begin{align}
        \ex{\norm{\mLambda_{\mA}^{1/2} \mS_i}^4}
        = \frac{2 \Fnorm{\mA}^2 + \left(\trace{\mA}\right)^2}{2 \dim \mS + (\dim \mS)^2}
        = \frac{(2 + \erank{\mA}) \Fnorm{\mA}^2}{(2 + \dim \mS) (\dim \mS)}
    \end{align}
    where $\erank{\mA} = \frac{\left(\trace{\mA}\right)^2}{\Fnorm{\mA}^2}$.

    For $j \neq i$, the diagonal entries are
    \begin{align}
        \frac{1}{\dim \mS - 1} \left(\frac{\Fnorm{\mA}^2}{\dim \mS} - \frac{(2 + \erank{\mA}) \Fnorm{\mA}^2}{(2 + \dim \mS) (\dim \mS)}\right)
        = \frac{(\dim \mS -\erank{\mA}) \Fnorm{\mA}^2}{(\dim \mS - 1) (2 + \dim \mS) (\dim \mS)}
    \end{align}
    Therefore,
    \begin{align}
        \ex{\mS \mLambda_{\mA} \mS^\top \ve_i \ve_i^\top \mS \mLambda_{\mA} \mS^\top}
        = p_{\mA} \cdot \frac{\Fnorm{\mA}^2}{\dim \mS} \cdot \mI + q_{\mA} \cdot \frac{\Fnorm{\mA}^2}{\dim \mS} \cdot \ve_i \ve_i^\top
    \end{align}
    where $p_{\mA} = \frac{\dim \mS - \erank{\mA}}{(\dim \mS - 1)(\dim \mS + 2)}$ and $q_{\mA} = \frac{\erank{\mA} + 1 + (\erank{\mA} - 1) / (\dim \mS - 1)}{\dim \mS + 2}$.

    Finally,
    \begin{align}
        \ex{\mS \mLambda_{\mA} \mS^\top \mLambda_{\mB} \mS \mLambda_{\mA} \mS^\top}
        = p_{\mA} \cdot \frac{\Fnorm{\mA}^2}{\dim \mS} \cdot \trace{\mLambda_{\mB}} \cdot \mI
        + q_{\mA} \cdot \frac{\Fnorm{\mA}^2}{\dim \mS} \cdot \mLambda_{\mB}
    \end{align}
    Transforming back, we obtain
    \begin{align}
        \ex{\mS \mA \mS^\top \mB \mS \mA \mS^\top}
        = p_{\mA} \cdot \frac{\Fnorm{\mA}^2}{\dim \mS} \cdot \trace{\mB} \cdot \mI
        + q_{\mA} \cdot \frac{\Fnorm{\mA}^2}{\dim \mS} \cdot \mB.
    \end{align}
\end{proof}

\begin{theorem}
    Assume \cref{assumption:generate_new_task,assumption:empirical_loss_local_minima,assumption:well_trained_old_task}.
    Finally, assume the DLN perfectly interpolates the old task, \ie{} $\rho = 0$.
    Define condition number $\kappa(\mA) \defeq \frac{\sigma_{\max}(\mA)}{\sigma_{\min}(\mA)}$ to measure the input data's deviation from being whitened, where $\sigma_{\min}(\cdot)$ is the least \emph{non-zero} singular value.
    Then we have the following lower-bound for the alignment between the old Hessian and the new gradient:
    \begin{align}
        \alignment(\oldhessian, \vg)
        \ge& \underbrace{\left( \scriptstyle \frac{1}{\kappa^3(\OldInputs \OldInputs^\top)} \cdot \frac{\sum_{i, j} \erank{\SingularMatrix{\OldLabels \OldInputs^\dagger}^{\max(i+j-2, 3L - i - j) / L}} }{\sum_{i=1}^L \erank{\SingularMatrix{\OldLabels \OldInputs^\dagger}^{\min(i-1, 2L - i) / L}}} \right)}_{\text{anisotropic inputs decrease alignment}} \\&\times \underbrace{\frac{\dim \param}{\erank{\SingularMatrix{\OldLabels \OldInputs^\dagger}^{2(1 - 1/L)}} \sum_{i=1}^L \erank{\SingularMatrix{\OldLabels \OldInputs^\dagger}^{2 \min(i - 1, L - i) / L}} }}_{\text{terms that can be found in whitened bounds}}. 
    \end{align}
\end{theorem}

\begin{remark}
    This theorem shows that when inputs are no longer whitened, alignment may drop.
    Specifically, when the inputs are not whitened, their largest and least singular values may differ a lot and have large condition number $\kappa$.
    Moreover, when non-zero singular values are not uniform, $\CommonSV^{i}$ may differ a lot between large and small $i$. As a result, they will have different effective ranks, making $\frac{\sum_{i, j} \erank{\SingularMatrix{\OldLabels \OldInputs^\dagger}^{\max(i+j-2, 3L - i - j) / L}} }{\sum_{i=1}^L \erank{\SingularMatrix{\OldLabels \OldInputs^\dagger}^{\min(i-1, 2L - i) / L}}}$ smaller than $L$. 
\end{remark}

\begin{proof}
    We extend the previous proofs to the setting where $\OldInputs$ is not full-rank.
    By the assumption that the old task is interpolated and the model weight is a local minimum of the $\ltwo$ regularization, we have $\weight_{L:1} = \OldLabels \OldInputs^\dagger$, where $\dagger$ denotes the Moore-Penrose pseudo-inverse.
    The extension is done to the Hessian trace, the gradient norm and the Hessian-gradient product.
    \subsubsection{Hessian trace}
    The Hessian trace from \cref{lemma:hessian_trace} can be upper-bounded by
    \begin{align}
        \trace{\oldhessian} 
        =&  \sum_{i=1}^L \Fnorm{\weight_{L:i+1}} \cdot \Fnorm{\weight_{i-1:1} \OldInputs}^2\\
        \le& \sum_{i=1}^L \Fnorm{\weight_{L:i+1}} \cdot \Fnorm{\weight_{i-1:1}}^2 \cdot \sigma_{\max}(\OldInputs \OldInputs^\top)\\
        =& \sigma_{\max}(\OldInputs \OldInputs^\top) \sum_{i=1}^L \Fnorm{\CommonSV^{L-i}}^2 \cdot \Fnorm{\CommonSV^{i-1}}^2 \\
        \le& \sigma_{\max}(\OldInputs \OldInputs^\top) \cdot \sum_{i=1}^L \Fnorm{\CommonSV^{L - 1}}^2 \cdot \erank{\CommonSV^{2 \min(i - 1, L - i)}}.
    \end{align}

    \subsubsection{Gradient Norm}

    The gradient norm from \cref{lemma:new_gradient_norm} can be upper-bounded by
    \begin{align}
        &   \ex{\norm{\partialfrac{\emploss_2(\param)}{\param}}^2} \\
        =& \sum_{i=1}^L \ex{
                \Fnorm{\weight_{L:i+1}^\top \weight_{L:1} \NewInputs \NewInputs^\top \weight_{i-1:1}^\top}^2
                +\Fnorm{\weight_{L:i+1}^\top \NewInputs \NewInputs^\top \weight_{i-1:1}^\top}^2 
        }\\
        =& \sum_{i=1}^L \biggl(
                \trace{
                    \weight_{L:i+1}^\top \weight_{L:1} \ex{\NewInputs \NewInputs^\top \weight_{i-1:1}^\top \weight_{i-1:1}  \NewInputs \NewInputs^\top} \weight_{L:1}^\top \weight_{L:i+1}
                }\\
                &\qquad\qquad\qquad + \trace{
                    \weight_{L:i+1}^\top \NewLabels \ex{\NewInputs^\top \weight_{i-1:1}^\top \weight_{i-1:1} \NewInputs} \NewLabels^\top \weight_{L:i+1}
                }
        \biggr)\\
        =& \sum_{i=1}^L \biggl(
                \trace{
                    \weight_{L:i+1}^\top \weight_{L:1} \ex{\RandRot \OldInputs \OldInputs^\top \RandRot^\top \weight_{i-1:1}^\top \weight_{i-1:1}  \RandRot \OldInputs \OldInputs^\top \RandRot^\top} \weight_{L:1}^\top \weight_{L:i+1}
                }\\
                &\qquad\qquad\qquad + \trace{
                    \weight_{L:i+1}^\top \OldLabels \OldInputs^\top \ex{ \RandRot^\top \weight_{i-1:1}^\top \weight_{i-1:1} \RandRot}\OldInputs \OldLabels^\top \weight_{L:i+1}
                }
        \biggr)\\
        =& \sum_{i=1}^L \biggl(
                \trace{
                    \weight_{L:i+1}^\top \weight_{L:1} \ex{\RandRot \OldInputs \OldInputs^\top \RandRot^\top \weight_{i-1:1}^\top \weight_{i-1:1}  \RandRot \OldInputs \OldInputs^\top \RandRot^\top} \weight_{L:1}^\top \weight_{L:i+1}
                }\\
                &+ \trace{
                    \weight_{L:i+1}^\top \OldLabels \OldInputs^\dagger \OldInputs \OldInputs^\top \ex{ \RandRot^\top \weight_{i-1:1}^\top \weight_{i-1:1} \RandRot} \OldInputs \OldInputs^\top \left(\OldInputs^\dagger\right)^\top \OldLabels^\top \weight_{L:i+1}
                }
        \biggr)
    \end{align}
    where $\RandRot$ is the random orthogonal matrix in \cref{assumption:generate_new_task}.
    Using \cref{corollary:second_matrix_moment_concrete,lemma:random_orthogonal_matrix_fourth_moment}, we have
    \begin{align}
        &\trace{
            \weight_{L:i+1}^\top \weight_{L:1} \ex{\RandRot \OldInputs \OldInputs^\top \RandRot^\top \weight_{i-1:1}^\top \weight_{i-1:1}  \RandRot \OldInputs \OldInputs^\top \RandRot^\top} \weight_{L:1}^\top \weight_{L:i+1}
        }\\
        =& \trace{
            \weight_{L:i+1}^\top \weight_{L:1}
                \frac{\Fnorm{\OldInputs \OldInputs^\top}^2}{\dim \Input} \cdot \left(p_{\OldInputs \OldInputs^\top} \cdot \Fnorm{\weight_{i-1:1}}^2 \cdot \mI + q_{\OldInputs \OldInputs^\top} \cdot \weight_{i-1:1}^\top \weight_{i-1:1} \right)
            \weight_{L:1}^\top \weight_{L:i+1}
        }\\
        =& \frac{\Fnorm{\OldInputs \OldInputs^\top}^2}{\dim \Input} \left(
            p_{\OldInputs \OldInputs^\top} \Fnorm{\weight_{L:i+1}^\top \weight_{L:1}} \cdot \Fnorm{\weight_{i-1:1}}^2
            + q_{\OldInputs \OldInputs^\top} \Fnorm{\weight_{L:i+1}^\top \weight_{L:1} \weight_{i-1:1}}^2
        \right),
    \end{align}
    and
    \begin{align}
        &\trace{
            \weight_{L:i+1}^\top \OldLabels \OldInputs^\dagger \OldInputs \OldInputs^\top \ex{ \RandRot^\top \weight_{i-1:1}^\top \weight_{i-1:1} \RandRot} \OldInputs \OldInputs^\top \left(\OldInputs^\dagger\right)^\top \OldLabels^\top \weight_{L:i+1}
        }\\
        =& \trace{
            \weight_{L:i+1}^\top \OldLabels \OldInputs^\dagger \OldInputs \OldInputs^\top \times 
            \frac{\trace{\weight_{i-1:1}^\top \weight_{i-1:1}}}{\dim \Input} \cdot \mI
            \times \OldInputs \OldInputs^\top \left(\OldInputs^\dagger\right)^\top \OldLabels^\top \weight_{L:i+1}
        }\\
        =& \frac{1}{\dim \Input} \Fnorm{\weight_{i-1:1}}^2 \cdot \Fnorm{\weight_{L:i+1}^\top \OldLabels \OldInputs^\dagger \OldInputs \OldInputs^\top}^2\\
        \le& \frac{\sigma_{\max}^2(\OldInputs \OldInputs^\top)}{\dim \Input} \Fnorm{\weight_{i-1:1}}^2 \cdot \Fnorm{\weight_{L:i+1}^\top \OldLabels \OldInputs^\dagger}.
    \end{align}
    As a result, we have
    \begin{align}
        &   \ex{\norm{\partialfrac{\emploss_2(\param)}{\param}}^2} \\
        \le& \sum_{i=1}^L \frac{\Fnorm{\OldInputs \OldInputs^\top}^2}{\dim \Input} \left(
            p_{\OldInputs \OldInputs^\top} \Fnorm{\weight_{L:i+1}^\top \weight_{L:1}} \cdot \Fnorm{\weight_{i-1:1}}^2
            + q_{\OldInputs \OldInputs^\top} \Fnorm{\weight_{L:i+1}^\top \weight_{L:1} \weight_{i-1:1}}^2
        \right) \\
        &\qquad \qquad + \sum_{i=1}^L \frac{\sigma_{\max}^2(\OldInputs \OldInputs^\top)}{\dim \Input} \Fnorm{\weight_{i-1:1}}^2 \cdot \Fnorm{\weight_{L:i+1}^\top \OldLabels \OldInputs^\dagger}^2 \\
        =& \sum_{i=1}^L \frac{\Fnorm{\OldInputs \OldInputs^\top}^2}{\dim \Input} \left(
            p_{\OldInputs \OldInputs^\top} \Fnorm{\CommonSV^{2L - i}} \cdot \Fnorm{\CommonSV^{i-1}}^2
            + q_{\OldInputs \OldInputs^\top} \Fnorm{\CommonSV^{2L - 1}}^2
        \right) \\
        &+ \sum_{i=1}^L \frac{\sigma_{\max}^2(\OldInputs \OldInputs^\top)}{\dim \Input} \Fnorm{\CommonSV^{i-1}}^2 \cdot \Fnorm{\CommonSV^{2L - i}}^2\\
        \le& \sum_{i=1}^L \frac{\Fnorm{\OldInputs \OldInputs^\top}^2}{\dim \Input} \left(
            p_{\OldInputs \OldInputs^\top} \erank{\CommonSV^{\min(2L - i, i-1)}}
            + q_{\OldInputs \OldInputs^\top}
        \right) \cdot \Fnorm{\CommonSV^{2L - 1}}^2 \\
        &+ \sum_{i=1}^L \frac{\sigma_{\max}^2(\OldInputs \OldInputs^\top)}{\dim \Input} \erank{\CommonSV^{\min(2L-i, i-1)}} \cdot \Fnorm{\CommonSV^{2L - 1}}^2.
    \end{align}

    \subsubsection{Hessian-Gradient Product}

    The Hessian-gradient product from \cref{lemma:gradient_hessian_product} can be expanded into
    \begin{align}
        &\ex{\vg^\top \oldhessian \vg}\\
        =&
            \ex{\Fnorm{ \sum_{i=1}^L \weight_{L:i+1} \left(\weight_{L:i+1}^\top \weight_{L:1} \NewInputs \NewInputs^\top \weight_{i-1:1}^\top\right) \weight_{i-1:1} \OldInputs}^2}\\
            &+ \ex{\Fnorm{ \sum_{i=1}^L \weight_{L:i+1} \left(\weight_{L:i+1}^\top  \NewLabels \NewInputs^\top \weight_{i-1:1}^\top\right) \weight_{i-1:1} \OldInputs}^2}\\
        =&
            \trace{\scriptstyle \sum_{i, j} 
                \weight_{L:i+1} \weight_{L:i+1}^\top \weight_{L:1} \ex{\NewInputs \NewInputs^\top \weight_{i-1:1}^\top \weight_{i-1:1} \OldInputs
                \OldInputs^\top \weight_{j-1:1}^\top  \weight_{j-1:1} \NewInputs \NewInputs^\top} \weight_{L:1}^\top \weight_{L:j+1} \weight_{L:j+1}^\top
                }\\
            &+ \trace{ \scriptstyle \sum_{i, j} 
                \weight_{L:i+1} \weight_{L:i+1}^\top  \NewLabels \ex{\NewInputs^\top \weight_{i-1:1}^\top \weight_{i-1:1} \OldInputs
                \OldInputs^\top \weight_{j-1:1}^\top \weight_{j-1:1} \NewInputs}\NewLabels^\top \weight_{L:j+1} \weight_{L:j+1}^\top
            }\\
        =&
            \trace{\scriptstyle \sum_{i, j} 
                \weight_{L:i+1} \weight_{L:i+1}^\top \weight_{L:1} \ex{\RandRot \OldInputs \OldInputs^\top \RandRot^\top \weight_{i-1:1}^\top \weight_{i-1:1} \OldInputs
                \OldInputs^\top \weight_{j-1:1}^\top  \weight_{j-1:1} \RandRot \OldInputs \OldInputs^\top \RandRot^\top} \weight_{L:1}^\top \weight_{L:j+1} \weight_{L:j+1}^\top
                }\\
            &+ \trace{ \scriptstyle \sum_{i, j} 
                \weight_{L:i+1} \weight_{L:i+1}^\top  \OldLabels \OldInputs^\top \ex{\RandRot^\top \weight_{i-1:1}^\top \weight_{i-1:1} \OldInputs
                \OldInputs^\top \weight_{j-1:1}^\top \weight_{j-1:1} \RandRot} \OldInputs \OldLabels^\top \weight_{L:j+1} \weight_{L:j+1}^\top
            }\\
    \end{align}
    Using \cref{corollary:second_matrix_moment_concrete,lemma:random_orthogonal_matrix_fourth_moment}, we have
    \begin{align}
        & \trace{\scriptstyle
            \weight_{L:i+1} \weight_{L:i+1}^\top \weight_{L:1} \ex{\RandRot \OldInputs \OldInputs^\top \RandRot^\top \weight_{i-1:1}^\top \weight_{i-1:1} \OldInputs
            \OldInputs^\top \weight_{j-1:1}^\top  \weight_{j-1:1} \RandRot \OldInputs \OldInputs^\top \RandRot^\top} \weight_{L:1}^\top \weight_{L:j+1} \weight_{L:j+1}^\top
        }\\
        =& \trr \biggl(\scriptstyle
            \weight_{L:i+1} \weight_{L:i+1}^\top \weight_{L:1} 
                \frac{\Fnorm{\OldInputs \OldInputs^\top}^2}{\dim \Input} \biggl(p_{\OldInputs \OldInputs^\top} \cdot \trace{\weight_{i-1:1}^\top \weight_{i-1:1} \OldInputs
            \OldInputs^\top \weight_{j-1:1}^\top  \weight_{j-1:1}} \cdot \mI \\
              & \scriptstyle + q_{\OldInputs \OldInputs^\top} \cdot \weight_{i-1:1}^\top \weight_{i-1:1} \OldInputs
            \OldInputs^\top \weight_{j-1:1}^\top  \weight_{j-1:1} \biggr) \weight_{L:1}^\top \weight_{L:j+1} \weight_{L:j+1}^\top
        \biggr),\\
        =&  \scriptstyle p_{\OldInputs \OldInputs^\top} \cdot \frac{\Fnorm{\OldInputs \OldInputs^\top}^2}{\dim \Input} \cdot \trace{
            \weight_{L:i+1} \weight_{L:i+1}^\top \weight_{L:1} \weight_{L:1}^\top \weight_{L:j+1} \weight_{L:j+1}^\top} \cdot \trace{\weight_{i-1:1}^\top \weight_{i-1:1} \OldInputs \OldInputs^\top \weight_{j-1:1}^\top  \weight_{j-1:1}}\\
              & \scriptstyle q_{\OldInputs \OldInputs^\top} \cdot \frac{\Fnorm{\OldInputs \OldInputs^\top}^2}{\dim \Input} \cdot \trace{
            \weight_{L:i+1} \weight_{L:i+1}^\top \weight_{L:1}  \cdot \weight_{i-1:1}^\top \weight_{i-1:1} \OldInputs
            \OldInputs^\top \weight_{j-1:1}^\top  \weight_{j-1:1} \weight_{L:1}^\top \weight_{L:j+1} \weight_{L:j+1}^\top
              },\\
        =&  \scriptstyle p_{\OldInputs \OldInputs^\top} \cdot \frac{\Fnorm{\OldInputs \OldInputs^\top}^2}{\dim \Input} \cdot \trace{
            \weight_{L:i+1} \weight_{L:i+1}^\top \weight_{L:1} \weight_{L:1}^\top \weight_{L:j+1} \weight_{L:j+1}^\top} \cdot \trace{\weight_{i-1:1}^\top \weight_{i-1:1} \OldInputs \OldInputs^\top \weight_{j-1:1}^\top  \weight_{j-1:1}}\\
              &+ \scriptstyle q_{\OldInputs \OldInputs^\top} \cdot \frac{\Fnorm{\OldInputs \OldInputs^\top}^2}{\dim \Input} \cdot \trace{
            \weight_{L:1}^\top \weight_{L:j+1} \weight_{L:j+1}^\top \weight_{L:i+1} \weight_{L:i+1}^\top \weight_{L:1}  \cdot \weight_{i-1:1}^\top \weight_{i-1:1} \OldInputs
            \OldInputs^\top \weight_{j-1:1}^\top  \weight_{j-1:1}
              },
    \end{align}
    and
    \begin{align}
        &\trace{ \scriptstyle
                \weight_{L:i+1} \weight_{L:i+1}^\top  \OldLabels \OldInputs^\top \ex{\RandRot^\top \weight_{i-1:1}^\top \weight_{i-1:1} \OldInputs
                \OldInputs^\top \weight_{j-1:1}^\top \weight_{j-1:1} \RandRot} \OldInputs \OldLabels^\top \weight_{L:j+1} \weight_{L:j+1}^\top
        }\\
        =&\trace{ \scriptstyle
                \weight_{L:i+1} \weight_{L:i+1}^\top  \OldLabels \OldInputs^\top \frac{\trace{\weight_{i-1:1}^\top \weight_{i-1:1} \OldInputs
                \OldInputs^\top \weight_{j-1:1}^\top \weight_{j-1:1}}}{\dim \Input} \OldInputs \OldLabels^\top \weight_{L:j+1} \weight_{L:j+1}^\top
        }\\
        =&\frac{1}{\dim \Input} \trace{ \scriptstyle \weight_{i-1:1}^\top \weight_{i-1:1} \OldInputs
                \OldInputs^\top \weight_{j-1:1}^\top \weight_{j-1:1}} \\
                &\cdot  \trace{ \scriptstyle
                \left(\OldInputs^\dagger\right)^\top \OldLabels^\top \weight_{L:j+1} \weight_{L:j+1}^\top \weight_{L:i+1} \weight_{L:i+1}^\top  \OldLabels \OldInputs^\dagger \OldInputs \OldInputs^\top \OldInputs \OldInputs^\top 
                }.
    \end{align}
    Now we handle $\OldInputs \OldInputs^\top$ in the traces.
    Note that other matrices multiplied with it are all symmetric and PSD, including $\weight_{i-1:1}^\top \weight_{i-1:1}, \weight_{j-1:1}^\top \weight_{j-1:1}$, $\weight_{L:1}^\top \weight_{L:j+1} \weight_{L:j+1}^\top \weight_{L:i+1} \weight_{L:i+1}^\top \weight_{L:1}  \cdot \weight_{i-1:1}^\top \weight_{i-1:1} = \mV_{1} \CommonSV^{6L - 2j - 2i - 2} \mV_{1}$, and $\left(\OldInputs^\dagger\right)^\top \OldLabels^\top \weight_{L:j+1} \weight_{L:j+1}^\top \weight_{L:i+1} \weight_{L:i+1}^\top  \OldLabels \OldInputs^\dagger = \mV_1 \mat{\Sigma}^{6L - 2i - 2j - 2} \mV_1$. We recall \cref{lemma:trace_bounds} to handle such situation.
    Note that all the above weight-formed PSD matrices satisfy the condition that their null spaces are superset of $\OldInputs \OldInputs^\top$'s null space: Under our assumption that the old-task is interpolated, we have $\weight_{L:1} = \OldLabels \OldInputs^\top$, whose (right) nullspace is the superset of $\OldInputs \OldInputs^\top$'s nullspace. Therefore, $\mV_{1} \Sigma^{k} \mV_{1}$, which shares the same null space as $\mV_{1} \Sigma^{2 L} \mV_{1} = \weight_{L:1}^\top \weight_{L:1}$, has the a superset null space of $\OldInputs \OldInputs^\top$.
    As a result, we can lower-bound the traces as follows:
    \begin{align}
        & \trace{\scriptstyle
            \weight_{L:i+1} \weight_{L:i+1}^\top \weight_{L:1} \ex{\RandRot \OldInputs \OldInputs^\top \RandRot^\top \weight_{i-1:1}^\top \weight_{i-1:1} \OldInputs
            \OldInputs^\top \weight_{j-1:1}^\top  \weight_{j-1:1} \RandRot \OldInputs \OldInputs^\top \RandRot^\top} \weight_{L:1}^\top \weight_{L:j+1} \weight_{L:j+1}^\top
        }\\
        \ge& \scriptstyle  \sigma_{\min}(\OldInputs \OldInputs^\top) \cdot   p_{\OldInputs \OldInputs^\top} \cdot \frac{\Fnorm{\OldInputs \OldInputs^\top}^2}{\dim \Input} \cdot \trace{
            \weight_{L:i+1} \weight_{L:i+1}^\top \weight_{L:1} \weight_{L:1}^\top \weight_{L:j+1} \weight_{L:j+1}^\top} \cdot \trace{\weight_{i-1:1}^\top \weight_{i-1:1} \weight_{j-1:1}^\top  \weight_{j-1:1}}\\
              &+ \scriptstyle \sigma_{\min}(\OldInputs \OldInputs^\top) \cdot  q_{\OldInputs \OldInputs^\top} \cdot \frac{\Fnorm{\OldInputs \OldInputs^\top}^2}{\dim \Input} \cdot \trace{
            \weight_{L:1}^\top \weight_{L:j+1} \weight_{L:j+1}^\top \weight_{L:i+1} \weight_{L:i+1}^\top \weight_{L:1}  \cdot \weight_{i-1:1}^\top \weight_{i-1:1} \weight_{j-1:1}^\top  \weight_{j-1:1}
              },\\
        \ge& \sigma_{\min}(\OldInputs \OldInputs^\top) \cdot \frac{\Fnorm{\OldInputs \OldInputs^\top}^2}{\dim \Input} \left(   
                p_{\OldInputs \OldInputs^\top} \cdot \Fnorm{\CommonSV^{3L - i - j}}^2 \cdot \Fnorm{\CommonSV^{i+j-2}}^2 + q_{\OldInputs \OldInputs^\top} \cdot \Fnorm{\CommonSV^{3L - 2}}
        \right)\\
        \ge& \sigma_{\min}(\OldInputs \OldInputs^\top) \cdot \frac{\Fnorm{\OldInputs \OldInputs^\top}^2}{\dim \Input} \cdot \Fnorm{\CommonSV^{3L - 2}} \left(   
                p_{\OldInputs \OldInputs^\top} \cdot \erank{\CommonSV^{\max(i+j-2, 3L - i - j)}} + q_{\OldInputs \OldInputs^\top}
        \right),
    \end{align}
    and
    \begin{align}
        &\trace{ \scriptstyle
                \weight_{L:i+1} \weight_{L:i+1}^\top  \OldLabels \OldInputs^\top \ex{\RandRot^\top \weight_{i-1:1}^\top \weight_{i-1:1} \OldInputs
                \OldInputs^\top \weight_{j-1:1}^\top \weight_{j-1:1} \RandRot} \OldInputs \OldLabels^\top \weight_{L:j+1} \weight_{L:j+1}^\top
        }\\
        \ge & \sigma_{\min}(\OldInputs \OldInputs^\top) \cdot \frac{1}{\dim \Input} \cdot \Fnorm{\CommonSV^{i+j-2}}^2 \cdot  \Fnorm{\CommonSV^{3L - i -j}}^2 \cdot \sigma_{\min}^2(\OldInputs\OldInputs^\top)\\
        \ge& \sigma_{\min}^3(\OldInputs \OldInputs^\top) \cdot \frac{1}{\dim \Input} \cdot \Fnorm{\CommonSV^{3L - 2}}^2 \cdot \erank{\CommonSV^{\max(i+j-2, 3L - i - j)}}.
    \end{align}
    
    As a result, we have the following lower-bound for the Hessian-gradient product:
    \begin{align}
        \ex{\vg^\top \oldhessian \vg}
        \ge& \frac{\sigma_{\min}(\OldInputs \OldInputs^\top)}{\dim \Input}\Biggl(
            \sum_{i, j} \erank{\CommonSV^{\max(i+j-2, 3L - i - j)}} \cdot \left(p_{\OldInputs \OldInputs^\top} \cdot \Fnorm{\OldInputs \OldInputs^\top}^2 + \sigma_{\min}^2(\OldInputs \OldInputs^\top) \right)\\
            & + q_{\OldInputs \OldInputs^\top} \cdot \Fnorm{\OldInputs \OldInputs^\top}^2 \cdot L^2
        \Biggr)  \Fnorm{\CommonSV^{3L - 2}}^2
    \end{align}

    \subsubsection{Alignment}

    Combining the above results, we have the following lower-bound for the alignment:
    \begin{align}
        &\frac{1}{\dim \param}\alignment(\oldhessian, \vg)
        \defeq \cdot \frac{\ex{\vg^\top \oldhessian \vg}}{\trace{\oldhessian} \cdot \ex{\norm{\partialfrac{\emploss_2(\param)}{\param}}^2}}\\
        =&\frac{\scriptstyle
            \frac{\sigma_{\min}(\OldInputs \OldInputs^\top)}{\dim \Input}\Biggl(
            \sum_{i, j} \erank{\CommonSV^{\max(i+j-2, 3L - i - j)}} \cdot \left(p_{\OldInputs \OldInputs^\top} \cdot \Fnorm{\OldInputs \OldInputs^\top}^2 + \sigma_{\min}^2(\OldInputs \OldInputs^\top)\right)
            + q_{\OldInputs \OldInputs^\top} \cdot \Fnorm{\OldInputs \OldInputs^\top}^2 \cdot L^2
        \Biggr)  \Fnorm{\CommonSV^{3L - 2}}^2
        }{\scriptstyle
            \substack{
                \sigma_{\max}(\OldInputs \OldInputs^\top) \cdot \sum_{i=1}^L \Fnorm{\CommonSV^{L - 1}}^2 \cdot \erank{\CommonSV^{2 \min(i - 1, L - i)}}\\
                \cdot \sum_{i=1}^L \left( \frac{\Fnorm{\OldInputs \OldInputs^\top}^2}{\dim \Input} \left(
            p_{\OldInputs \OldInputs^\top} \cdot \erank{\CommonSV^{\min(i-1, 2L - i)}}
            + q_{\OldInputs \OldInputs^\top}
        \right) + \frac{\sigma_{\max}^2(\OldInputs \OldInputs^\top)}{\dim \Input} \erank{\CommonSV^{\min(i-1, 2L - i)}} 
            \right) \cdot \Fnorm{\CommonSV^{2L - 1}}^2  }
        }\\
        =& \frac{1}{\scriptstyle \kappa(\OldInputs \OldInputs^\top) \cdot \erank{\CommonSV^{2(L-1)}}} \\
            & \cdot \frac{\scriptstyle
            \sum_{i, j} \erank{\CommonSV^{\max(i+j-2, 3L - i - j)}} \cdot \left(p_{\OldInputs \OldInputs^\top} \cdot \Fnorm{\OldInputs \OldInputs^\top}^2 + \sigma_{\min}^2(\OldInputs \OldInputs^\top)\right)
            + q_{\OldInputs \OldInputs^\top} \cdot \Fnorm{\OldInputs \OldInputs^\top}^2 \cdot L^2
        }{\scriptstyle
            \substack{
                \sum_{i=1}^L \erank{\CommonSV^{2 \min(i - 1, L - i)}}\\
                \cdot \left(\sum_{i=1}^L \Fnorm{\OldInputs \OldInputs^\top}^2 \left(
            p_{\OldInputs \OldInputs^\top} \cdot \erank{\CommonSV^{\min(i-1, 2L - i)}}
            + q_{\OldInputs \OldInputs^\top}
        \right) + \sum_{i=1}^L \sigma_{\max}^2(\OldInputs \OldInputs^\top)  \erank{\CommonSV^{\min(i-1, 2L - i)}} 
            \right)}
        }\\
        =& \frac{1}{\scriptstyle \kappa(\OldInputs \OldInputs^\top) \cdot \erank{\CommonSV^{2(L-1)}} \sum_{i=1}^L \erank{\CommonSV^{2 \min(i - 1, L - i)}}} \\
        &\cdot \frac{\scriptstyle
            P_{\OldInputs \OldInputs^\top} \sum_{i, j} \erank{\CommonSV^{\max(i+j-2, 3L - i - j)}} \cdot \frac{P^{\min}_{\OldInputs \OldInputs^\top}}{P_{\OldInputs \OldInputs^\top}}
            + Q_{\OldInputs \OldInputs^\top} \cdot L^2
        }{\scriptstyle
            P_{\OldInputs \OldInputs^\top} \cdot \sum_{i=1}^L \erank{\CommonSV^{\min(i-1, 2L - i)}}
            + Q_{\OldInputs \OldInputs^\top} \cdot L
        }, 
    \end{align}
    where $P_{\OldInputs \OldInputs^\top} \defeq p_{\OldInputs \OldInputs^\top} \cdot \Fnorm{\OldInputs \OldInputs^\top}^2 + \sigma_{\max}^2(\OldInputs \OldInputs^\top), P^{\min}_{\OldInputs \OldInputs^\top} \defeq p_{\OldInputs \OldInputs^\top} \cdot \Fnorm{\OldInputs \OldInputs^\top}^2 + \sigma_{\min}^2(\OldInputs \OldInputs^\top)$ and $Q_{\OldInputs \OldInputs^\top} \defeq q_{\OldInputs \OldInputs^\top} \cdot \Fnorm{\OldInputs \OldInputs^\top}^2$.

    It is easy to show given $a, b, c, d, p, q > 0$, one has $\frac{p \cdot a + q \cdot b}{p \cdot c + q \cdot d} \ge \min\left(\frac{a}{c}, \frac{b}{d}\right)$. Therefore, we have
    \begin{align}
        & \frac{\alignment(\oldhessian, \vg)}{\dim \param}\\
        \ge& \frac{1}{\kappa(\OldInputs \OldInputs^\top)} \cdot \frac{1}{\sum_{i=1}^L \erank{\CommonSV^{2 \min(i - 1, L - i)}}} \cdot \min\left( \scriptstyle
            \frac{\sum_{i, j} \erank{\CommonSV^{\max(i+j-2, 3L - i - j)}} }{\sum_{i=1}^L \erank{\CommonSV^{\min(i-1, 2L - i)}}} \cdot \frac{P^{\min}_{\OldInputs \OldInputs^\top}}{P_{\OldInputs \OldInputs^\top}}, \displaystyle
            \frac{L^2}{L}
        \right)\\
        =& \frac{1}{\kappa(\OldInputs \OldInputs^\top)}\cdot \frac{1}{\sum_{i=1}^L \erank{\CommonSV^{2 \min(i - 1, L - i)}}} \cdot \frac{\sum_{i, j} \erank{\CommonSV^{\max(i+j-2, 3L - i - j)}} }{\sum_{i=1}^L \erank{\CommonSV^{\min(i-1, 2L - i)}}} \cdot \frac{P^{\min}_{\OldInputs \OldInputs^\top}}{P_{\OldInputs \OldInputs^\top}}. 
    \end{align}
    To bound $\frac{P^{\min}_{\OldInputs \OldInputs^\top}}{P_{\OldInputs \OldInputs^\top}}$, we have
    \begin{align}
        \frac{P^{\min}_{\OldInputs \OldInputs^\top}}{P_{\OldInputs \OldInputs^\top}} 
        =& \frac{p_{\OldInputs \OldInputs^\top} \cdot \Fnorm{\OldInputs \OldInputs^\top}^2 + \sigma_{\min}^2(\OldInputs \OldInputs^\top)}{p_{\OldInputs \OldInputs^\top} \cdot \Fnorm{\OldInputs \OldInputs^\top}^2 + \sigma_{\max}^2(\OldInputs \OldInputs^\top)} \\
        \ge& \min\left(1, \frac{1}{\kappa^2(\OldInputs \OldInputs^\top)}\right).
    \end{align}
    As a result, we have
    \begin{align}
        & \frac{\alignment(\oldhessian, \vg)}{\dim \param}\\
        =& \frac{1}{\kappa^3(\OldInputs \OldInputs^\top)}\cdot \frac{1}{\sum_{i=1}^L \erank{\CommonSV^{2 \min(i - 1, L - i)}}} \cdot \frac{\sum_{i, j} \erank{\CommonSV^{\max(i+j-2, 3L - i - j)}} }{\sum_{i=1}^L \erank{\CommonSV^{\min(i-1, 2L - i)}}}. 
    \end{align}
\end{proof}